\documentclass[journal]{IEEEtran}

\usepackage{url}
\usepackage{amssymb}
\usepackage[cmex10]{amsmath}
\usepackage{amsthm}

\usepackage{subfigure}
\usepackage{epsfig}
\usepackage[ruled]{algorithm2e}
\usepackage{algorithmic}
\usepackage{multicol}

\usepackage{mathtools}
\DeclarePairedDelimiter{\ceil}{\lceil}{\rceil}
\usepackage{wrapfig}
\usepackage{cite}

\usepackage{accents}

\usepackage{color}

\newcommand{\rev}[1]{{\color{black}{#1}}}

\renewcommand{\max}{\text{max}}
\renewcommand{\min}{\text{min}}

\newtheorem{lemma}{Lemma}

\newtheorem{theorem}{Theorem}
\newtheorem{problem}{Problem}

\begin{document}

% \title{An Approximation Algorithm for Risk-averse Submodular Optimization}
\title{Risk-Aware Submodular Optimization for Multi-Robot Coordination}

\author{Lifeng~Zhou,~\IEEEmembership{Member,~IEEE,} and~Pratap~Tokekar,~\IEEEmembership{Member,~IEEE}
    % <-this % stops a space
\thanks{L. Zhou was with the Department of Electrical and Computer Engineering, Virginia Tech, Blacksburg, VA, USA when part of the work was completed. He is currently with the GRASP Laboratory, University of Pennsylvania, Philadelphia, PA, USA. (email: \texttt{\small lfzhou@seas.upenn.edu}).}
  \thanks{P. Tokekar is with the Department of Computer Science, University of Maryland, College Park, USA. (email: \texttt{\small tokekar@umd.edu}).}
 \thanks{This material is based upon work supported by the NSF under grant number IIS-1637915 and ONR under grant number N00014-18-1-2829.} 
  % <-this % stops a space
% <-this % stops a space
% \thanks{Manuscript received April 19, 2005; revised August 26, 2015.}
}
% % The paper headers
% \markboth{Journal of \LaTeX\ Class Files,~Vol.~14, No.~8, August~2015}%
% {Shell \MakeLowercase{\textit{et al.}}: Bare Demo of IEEEtran.cls for IEEE Journals}

% make the title area
\maketitle

% As a general rule, do not put math, special symbols or citations
% in the abstract or keywords.
\begin{abstract}
We study the problem of incorporating risk while making combinatorial decisions under uncertainty. We formulate a discrete submodular maximization problem for selecting a set using Conditional-Value-at-Risk (CVaR), a risk metric commonly used in financial analysis. While CVaR has recently been used in optimization of linear cost functions in robotics, we take the first step towards extending this to discrete submodular optimization and provide several positive results. Specifically, we propose the Sequential Greedy Algorithm that provides an approximation guarantee on finding the maxima of the CVaR cost function under a matroidal constraint. The approximation guarantee shows that the solution produced by our algorithm is within a constant factor of the optimal and an additive term that depends on the optimal. Our analysis uses the curvature of the submodular set function, and proves that the algorithm runs in polynomial time.  This formulates a number of combinatorial optimization problems that appear in robotics. We use two such problems, vehicle assignment under uncertainty for mobility-on-demand and sensor selection with failures for environmental monitoring, as case studies to demonstrate the efficacy of our formulation. We also study the problem of adaptive risk-aware submodular  maximization. We design a heuristic solution that triggers the replanning only when certain conditions are satisfied, to eliminate unnecessary planning. In particular, for the online mobility-on-demand study, we propose an adaptive triggering assignment algorithm that triggers a new assignment only when it can potentially reduce the waiting time at demand locations. We verify the performance of the Sequential Greedy Algorithm and the adaptive triggering assignment algorithm through simulations.
\end{abstract}

% Note that keywords are not normally used for peerreview papers.
\begin{IEEEkeywords}
risk-aware decision making, conditional value at risk, submodular maximization, sequential greedy algorithm.
\end{IEEEkeywords}

\section{Introduction}\label{sec:intro}
Combinatorial optimization problems find a variety of applications in robotics. Typical examples include:
\begin{itemize}
\item \emph{Sensor placement:} Where to place sensors to maximally cover the environment~\cite{o1987art} or reduce the uncertainty in the environment~\cite{krause2008near}?   
\item \emph{Task allocation:} How to allocate tasks to robots to maximize the overall utility gained by the robots~\cite{gerkey2004formal}?
\item \emph{Combinatorial auction:} How to choose a combination of items for each player to maximize the total rewards~\cite{vondrak2008optimal}?
\end{itemize}
Algorithms for solving such problems find use in sensor placement for environment monitoring~\cite{o1987art,krause2008near}, robot-target assignment and tracking~\cite{spletzer2003dynamic,tekdas2010sensor,tokekar2014multi,zhou2018resilient,zhou2019sensor,zhou2020distributed}, and informative path planning~\cite{singh2009efficient,shi2020robust}. The underlying optimization problem in most cases can be written as:
\begin{equation}
\underset{{\mathcal{S} \in \mathcal{I}, \mathcal{S}\subseteq \mathcal{X}}}{\text{max}} f(\mathcal{S}),
\label{eqn:basic}
\end{equation}
where $\mathcal{X}$ denotes a ground set from which a subset of elements $\mathcal{S}$ must be chosen. Typically, $f$ is a monotone submodular utility function~\cite{nemhauser1978analysis,fisher1978analysis}. Submodularity is the property of diminishing returns. Many information theoretic measures, such as mutual information~\cite{krause2008near}, and geometric measures such as the visible area~\cite{ding2017multi}, are known to be submodular. $\mathcal{I}$ denotes a matroidal constraint~\cite{nemhauser1978analysis,fisher1978analysis}. Matroids are a powerful combinatorial tool that can represent constraints on the solution set, e.g., cardinality constraints (``place no more than $k$ sensors'') and connectivity constraints (``the communication graph of the robots must be connected'')~\cite{williams2017decentralized}. The objective of this problem is to find a set $\mathcal{S}$ satisfying a matroidal constraint $\mathcal{I}$ and maximizing the utility $f(\mathcal{S})$. The general form of this problem is NP-complete. However, a greedy algorithm yields a constant factor approximation guarantee~\cite{nemhauser1978analysis,fisher1978analysis}.

In practice, sensors can fail or get compromised~\cite{wood2002denial} or robots may not know the exact positions of the targets~\cite{dames2017detecting}. Hence, the utility $f(\mathcal{S})$ is not necessarily deterministic but can have uncertainty. Our main contribution is to extend the traditional formulation given in Equation~\ref{eqn:basic} to also account for the uncertainty in the actual cost function. We model the uncertainty by assuming that the utility function is of the form $f(\mathcal{S},y)$ where $\mathcal{S}\in \mathcal{X}$ is the decision variable and $y\in \mathcal{Y}$ represents a random variable which is independent of $\mathcal{S}$. We focus on the case where $f(\mathcal{S},y)$ is monotone submodular in $\mathcal{S}\in \mathcal{X}$ and integrable in $y$.

The traditional way of stochastic optimization is to use the expected utility as the objective function: 
\begin{equation}
    \underset{\mathcal{S} \in \mathcal{I}, \mathcal{S}\in \mathcal{X}}{\max} {\mathbb{E}_y} [f(\mathcal{S},y)].
    \label{eqn:expectation}
\end{equation}
Since the sum of the monotone submodular functions is  monotone submodular, $\mathbb{E}_y [f(\mathcal{S},y)]$ is still monotone submodular in $\mathcal{S}$. Thus, the greedy algorithm still retains its constant-factor performance guarantee~\cite{nemhauser1978analysis,fisher1978analysis}. Examples of this approach include  influence maximization~\cite{kempe2003maximizing}, moving target detection and tracking~\cite{dames2017detecting}, and robot assignment with travel-time uncertainty~\cite{prorok2019redundant}. 

\begin{figure}
  \centering
  \includegraphics[width=0.8\columnwidth]{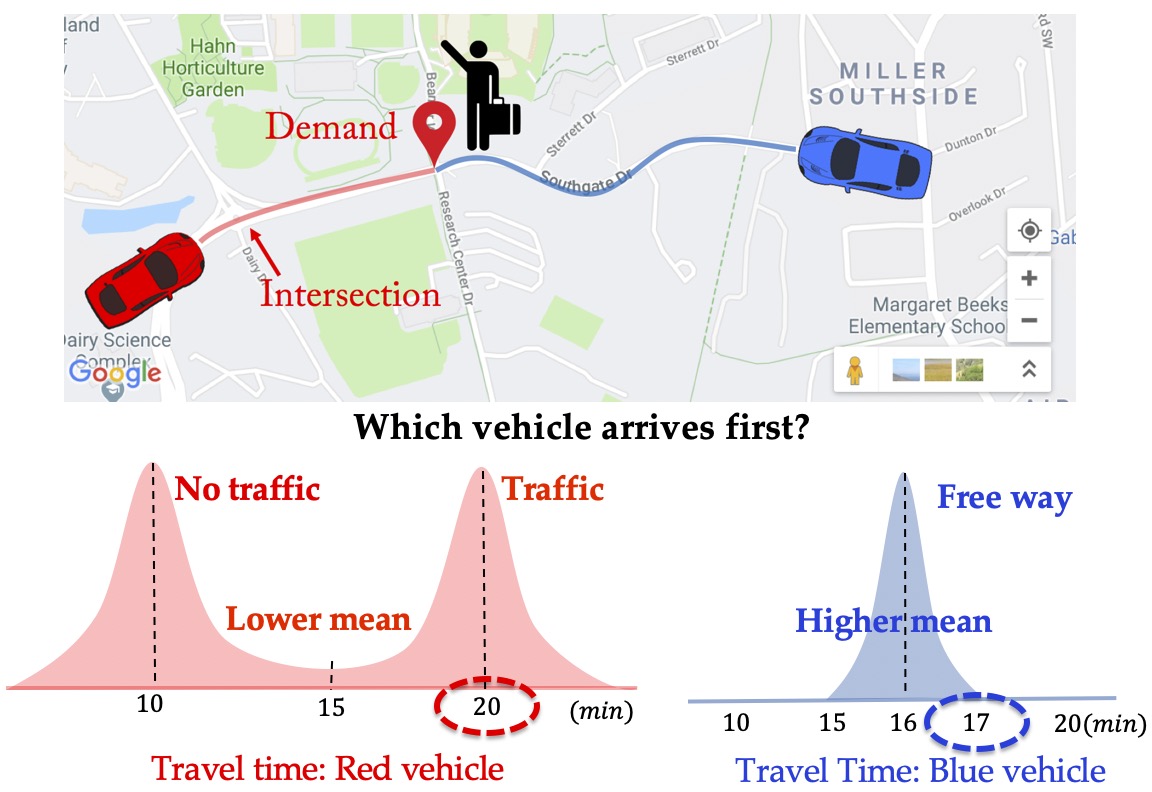}
  \caption{Mobility on demand with travel time uncertainty of self-driving vehicles.}
  \label{fig:simple_mod}
\end{figure}

While optimizing the expected utility has its uses, it also has its pitfalls too. Consider the example of mobility-on-demand where two  vehicles, the red vehicle and the blue vehicle, are available to pick up the passengers at a demand location (Fig.~\ref{fig:simple_mod}).  The red vehicle is closer to the demand location, but it needs to cross an intersection where it may need to stop and wait. The blue vehicle is further from the demand location but there is no intersection along the path. The travel time for the red vehicle follows a bimodal distribution (with and without traffic stop) whereas that for the blue vehicle follows a unimodal distribution with a higher mean but lower uncertainty. Clearly, if the passenger uses the expected travel time as the objective, they would choose the red vehicle. However, they will risk waiting a much longer time, i.e., $17 \sim 20~min$ about half of the times. A more risk-aware passenger would choose the blue vehicle which has higher expected waiting time $16~min$ but a lesser risk of waiting longer.

Thus, in these scenarios, it is natural to go beyond expectation  and focus on a risk-aware measure. One popular coherent risk measure is \emph{Conditional-Value-at-Risk} (CVaR)~\cite{pflug2000some,rockafellar2000optimization}. CVaR is parameterized by a risk level $\alpha$. Informally, maximizing CVaR is equivalent to maximizing the expected utility in the worst $\alpha$-tail scenarios.\footnote{We formally review CVaR and other related concepts in Section~\ref{subsec:risk}.} %This risk-aware decision is rational especially when the failures can lead to unrecoverable consequences, such as a sensor failure.

%, decision-makers often prefer the risk-aware strategy: they would like to reduce the chance of having a very low utility instead of purely considering the average, like the expectation.  Also, in many safety issues, reducing the risk of fault is paramount. For example, if a camera can detect
%illegal poaching of wild animals in $90\%$ cases, but fails entirely in $10\%$ cases, it is inevitable that the wild animals will be exposed to an unacceptable
%safety risk. Thus, a camera can always detect the poaching, even at the expense of higher energy consumption on average, is preferred. 

% \PRT{The transition to related work is abrupt. You've just introduced CVAR. Talk about the work related to CVAR before the chance-constrained one.}

\textbf{Related work}. 
Several works have focused on optimizing CVaR. In their seminal work~\cite{rockafellar2000optimization}, Rockafellar and Uryasev presented an algorithm for CVaR minimization for reducing the risk in financial portfolio optimization with a large number of instruments. Note that, in {portfolio optimization}, we select a distribution over available decision variables, instead of selecting a single one. Later, they showed the advantage of optimizing CVaR for general loss distributions in finance~\cite{rockafellar2002conditional}. 

Another popular risk measure in finance is the Value-at-Risk (VaR)~\cite{morgan1996riskmetrics}, which is generally used for formulating the chance-constrained optimization problems. Yang and Chakraborty studied a chance-constrained combinatorial optimization problem that takes into account the risk in multi-robot assignment~\cite{yang2017algorithm}. They later extended this to knapsack problems~\cite{yang2018algorithm}. They solved the problem by transforming it to a risk-aware problem with mean-variance measure~\cite{markowitz1952portfolio}.

Compared to VaR, Rockafellar and Uryasev stated that CVaR has more desirable mathematical characteristics such as subadditivity, convexity, and coherence, and is easier to optimize when it is calculated from scenarios~\cite{rockafellar2000optimization}. Also, Majumdar and Pavone argued that CVaR has better properties for quantifying risk than VaR or mean-variance based on six proposed axioms in the context of robotics~\cite{majumdar2020should}. 
% For example, in the emergency response scenarios, some low probability delay may cause human deaths.

When the utility is a discrete submodular set function, i.e., $f(\mathcal{S},y)$, 
% of the utility function, that is maximize $\text{CVaR}_{\alpha} (f(\mathcal{S},y))$ with $\mathcal{S} \in \mathcal{I}, \mathcal{S}\in \mathcal{X}, y\in \mathcal{Y}$ Indeed, 
Maehara presented a negative result for maximizing $\text{CVaR}$~\cite{maehara2015risk}---  there is no polynomial time multiplicative approximation algorithm for this problem under some reasonable assumptions in computational complexity. To avoid this difficulty, Ohsaka and Yoshida in~\cite{ohsaka2017portfolio} used the same idea from {portfolio optimization} and proposed a method of selecting a distribution over available sets rather than selecting a single set, and gave a provable guarantee. Following this line, Wilder considered a CVaR maximization of a {continuous submodular} function instead of the submodular set functions~\cite{wilder2018risk}. They gave a $(1 - 1/e)$--approximation algorithm for {continuous submodular} functions and also evaluated the algorithm for discrete submodular  functions using {portfolio optimization}~\cite{ohsaka2017portfolio}.

\textbf{Contributions}. In this paper,
we first focus on the problem of selecting a single set, similar to~\cite{maehara2015risk}, to optimize CVaR of a stochastic submodular set function. We propose an approximation algorithm with provable theoretical guarantees. Then we study the problem of adaptive risk-aware submodular maximization. We design a triggering replanning strategy that solves the CVaR optimization problem only at specific time steps to avoid unnecessary planning costs.

% Then, we present an online variant of this algorithm, specifically for the mobility-on-demand problem. We study how to trigger replanning by solving the CVaR optimization problem to avoid unnecessary vehicle assignments.

% \PRT{This conflicts with the online case where reassignments are done. Also, fundamentally why is portfolio not one-shot? I don't think you need to mention portfolio here. Instead, just rephrase the whole thing and say that we study two versions of the problem and explain them.}
% This is because we are motivated by applications where a one-shot decision (placing sensors and assigning vehicles) must be taken.

Our contributions are as follows: 
\begin{itemize}
\item We propose the Sequential Greedy Algorithm (SGA) which uses the deterministic greedy algorithm~\cite{nemhauser1978analysis,fisher1978analysis} as a subroutine to find the maximum value of CVaR (Algorithm~\ref{alg:sga}).  
\item We prove that the solution found by SGA is within a constant factor of the optimal performance along with an additive term which depends on the optimal value and a sampling error. We also prove that SGA runs in polynomial time (Theorem~\ref{thm:appro_bound_compu}) and the performance improves as the running time increases.

\item We formulate an adaptive risk-aware submodular maximization problem for the online version  (Problem~\ref{pro:replan}). We devise an event-triggered replanning strategy that replans by SGA only when certain conditions are reached. Particularly, we design the Adaptive Triggering Assignment (ATA) for the online mobility-on-demand application (Algorithm~\ref{alg:online_tri_assign_gen}) which triggers reassignment only at some specific time steps to avoid unnecessary assignments.

\item  We demonstrate the utility of the proposed CVaR maximization problem through two case studies (Section~\ref{subsec:case_study}). We evaluate the performance of SGA and ATA through simulations (Section~\ref{sec:simulation}). 
\end{itemize}

 \textbf{Organization of rest of the paper}. We give the necessary background knowledge for the rest of the paper in Section~\ref{sec:background}. We formulate the CVaR submodular maximization problem with two case studies in Section~\ref{sec:problem_case}. We present SGA along with the analysis of its computational complexity and approximation ratio in Section~\ref{sec:alg_ana}. We present an adaptive risk-aware submodular maximization problem and design a triggering replanning strategy in Section~\ref{sec:adaptive_submodular}. We illustrate the performance of SGA to the two case studies and evaluate the performance of ATA with street networks in Section~\ref{sec:simulation}. We conclude the paper in Section~\ref{sec:conclue}. 

A preliminary version of this paper was first presented
in~\cite{zhou2018approximation} without the analysis of the approximation error induced by sampling method (see Section~\ref{sec:alg_ana}), the formulation of the adaptive risk-aware submodular maximization problem and a triggering replanning strategy (see Section~\ref{sec:adaptive_submodular}), and the numerical evaluations of ATA on street networks (see Section~\ref{subsubsec:tri_assign_trig}).

\section{Background and Preliminaries}\label{sec:background}

\begin{figure}
  \centering
  \includegraphics[width=0.8\columnwidth]{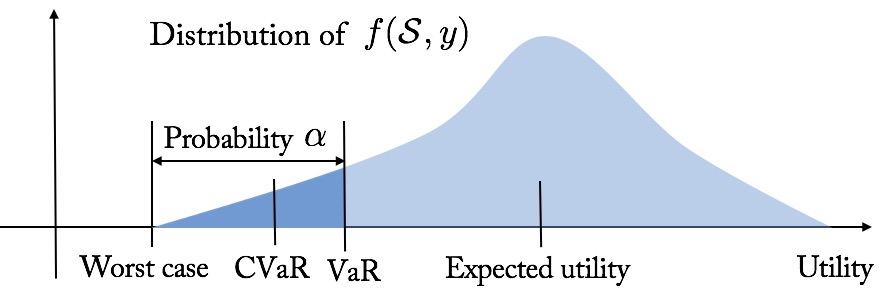}
  \caption{An illustration of risk measures: VaR and CVaR.}
  \label{fig:var_cvar}
\end{figure}

We start by defining the conventions used in the paper. 

Calligraphic font denotes a set (e.g., $\mathcal{A}$).  Given a set $\mathcal{A}$,  $2^{\mathcal{A}}$ denotes its power set. $|\mathcal{A}|$ denotes the cardinality of $\mathcal{A}$. Given a set $\mathcal{B}$, $\mathcal{A}\setminus\mathcal{B}$ denotes the set of elements in $\mathcal{A}$ that are not in~$\mathcal{B}$. $\text{Pr}[\cdot]$ denotes the probability of an event and $\mathbb{E}[\cdot]$ denotes the expectation of a random variable.  $\ceil{x} = \min\{n\in\mathbb{Z}|x\leq n\}$ where  $\mathbb{Z}$ denotes the set of integers. 

Next, we give the background on set functions, the greedy algorithm, and risk measures. 

\subsection{Background on set functions: Monotonicity, Submodularity, Matroid, and Curvature}

We begin by reviewing useful properties of a set function $f(\mathcal{S})$ defined for a finite ground set $\mathcal{X}$ and matroid constraints.

\vspace{3pt}\noindent\textbf{Monotonicity~{\cite{nemhauser1978analysis}}:}  A set function $f: 2^{\mathcal{X}} \mapsto \mathbb{R}$ is monotone (non-decreasing) if and only if for any sets $\mathcal{S}\subseteq \mathcal{S}'\subseteq \mathcal{X}$, we have $f(\mathcal{S})\leq f(\mathcal{S}')$.

\vspace{3pt}\noindent\textbf{Normalized Function~\cite{fisher1978analysis}:}
 A set function $f: 2^{\mathcal{X}} \mapsto \mathbb{R}$ is called normalized if and only if  $f(\emptyset) = 0$.

\vspace{3pt}\noindent\textbf{Submodularity~{\cite[Proposition 2.1]{nemhauser1978analysis}}:} A set function $f: 2^{\mathcal{X}} \mapsto \mathbb{R}$ is submodular if and only if for any sets $\mathcal{S}\subseteq \mathcal{S}'\subseteq \mathcal{X}$, and any element $s\in \mathcal{X}$ and $s \notin \mathcal{S}' $, we have: $f(\mathcal{S}\cup \{s\})- f(\mathcal{S}) \geq f(\mathcal{S}'\cup \{s\})- f(\mathcal{S}')$. Therefore the marginal gain $f(\mathcal{S}\cup \{s\})- f(\mathcal{S})$ is non-increasing. 

\vspace{3pt}\noindent\textbf{Matroid~{\cite[Section 39.1]{schrijver2003combinatorial}}:}  Denote a non-empty collection of subsets of $\mathcal{X}$ as $\mathcal{I}$. The pair $(\mathcal{X}, \mathcal{I})$ is called a matroid if and only if the following conditions are satisfied:\begin{itemize}
\item for any set $\mathcal{S}\subseteq \mathcal{X}$ it must hold that $\mathcal{S}\in\mathcal{I}$, and for any set $\mathcal{P}\subseteq \mathcal{S}$ it must hold that $\mathcal{P}\in\mathcal{I}$. 
\item for any sets $\mathcal{S}, \mathcal{P} \in \mathcal{I}$ and $|\mathcal{P}| \leq |\mathcal{S}|$, it must hold that there exists an element $s\in \mathcal{S} \backslash\mathcal{P}$ such that $\mathcal{P} \cup \{s\} \in \mathcal{I}$. 
\end{itemize} 
We will use two specific forms of matroids that are reviewed next. 

\vspace{3pt}\noindent\textbf{Uniform Matroid:} A \emph{uniform matroid} is a matroid $(\mathcal{X}, \mathcal{I})$ such that  for a positive integer $\kappa$, $\{\mathcal{S}: \mathcal{S}\subseteq \mathcal{X}, |\mathcal{S}|\leq \kappa\}$. Thus, the uniform matroid only constrains the cardinality of the feasible sets in $\mathcal{I}$. 

\vspace{3pt}\noindent\textbf{Partition Matroid:} A \emph{partition matroid} is a matroid $(\mathcal{X}, \mathcal{I})$ such that for a positive integer $n$, disjoint sets $\mathcal{X}_1, ..., \mathcal{X}_n$ and positive integers $\kappa_1,..., \kappa_n$, $\mathcal{X}\equiv \mathcal{X}_1 \cup \cdots \mathcal{X}_n$ and $\mathcal{I} = \{\mathcal{S}: \mathcal{S} \subseteq \mathcal{X}, |\mathcal{S}\cap\mathcal{X}_i|\leq \kappa_i$ for all $i=1,...,n$\}.   

\vspace{3pt}\noindent\textbf{Curvature~\cite{conforti1984submodular}:} 
Consider a matroid $\mathcal{I}$ for $\mathcal{X}$, and a non-decreasing submodular set function $f:2^{\mathcal{X}}\mapsto\mathbb{R}$ such that (without loss of generality) for any element $s \in \mathcal{X}$, $f(s)\neq 0$.  The curvature measures how far $f$ is from submodularity or linearity.  Define \emph{curvature} of $f$ over the matroid $\mathcal{I}$ as: \begin{equation}\label{eqn:curvature}
c_f \triangleq 1-\underset{s\in\mathcal{S}, \mathcal{S}\in \mathcal{I}}{\min} \frac{f(\mathcal{S})-f(\mathcal{S}\setminus \{ s\})}{f(s)}.
\end{equation}
Note that the definition of curvature $c_f$ (Eq.~\ref{eqn:curvature}) implies that  $0 \leq c_f\leq 1$. Specifically, if $c_f = 0$, it means for all the feasible sets $\mathcal{S} \in \mathcal{X}$, $f(\mathcal{S}) = \sum_{s\in \mathcal{S}} f(s)$. In this case, $f$ is a modular function. In contrast, if $c_f = 1$, then there exist a feasible $\mathcal{S}\in\mathcal{I}$ and an element $s \in \mathcal{X}$ such that $f(\mathcal{S}) = f(\mathcal{S} \setminus \{s\})$. In this case, the element $s$ is redundant for the contribution of the value of $f$ given the set $\mathcal{S}\setminus \{s\}$. 

\subsection{Greedy Approximation Algorithm}
In order to maximize a set function $f$, the greedy algorithm selects each element $s$ of $\mathcal{X}$ based on the maximum marginal gain at each round, that is,  
$$s = \underset{s \in \mathcal{X}}{\text{argmax}}~f(\mathcal{P} \cup \{s\}) - f(\mathcal{P}) $$
where $\mathcal{P} \subseteq \mathcal{S}$. 

We consider maximizing a normalized monotone submodular set function $f$. For any matroid, the greedy algorithm gives a $1/2$ approximation~\cite{fisher1978analysis}. 
% that is, 
% $$\frac{f(\mathcal{S}^{G})}{f^{\star}} \geq 1/2. $$ 
In particular, the greedy algorithm can give a $(1-1/e)$--approximation of the optimal solution under the uniform matroid~\cite{nemhauser1978analysis}. %That is, 
% $$\frac{f(\mathcal{S}^{G})}{f^{\star}} \geq 1-1/e $$
If we know the curvature of the set function $f$, we have a $1/(1+c_f)$ approximation for any matroid constraint~\cite[Theorem 2.3]{conforti1984submodular}. That is, 
$$\frac{f(\mathcal{S}^{G})}{f^{\star}} \geq \frac{1}{1+c_f}. $$
where $\mathcal{S}^{G} \in \mathcal{I}$ is the set selected by the greedy algorithm, $\mathcal{I}$ is the uniform matroid and $f^{\star}$ is the function value with optimal solution. 
Note that, if $c_f = 0$, which means $f$ is modular, then the greedy algorithm reaches the optimal. If $c_f = 1$, then we have the $1/2$--approximation.

\subsection{Risk measures}\label{subsec:risk}
Let $f(\mathcal{S}, y)$ be a utility function with decision set $\mathcal{S}$ and the random variable $y$. For each $\mathcal{S}$, the utility $f(\mathcal{S}, y)$ is also a random variable with a distribution induced by that of $y$. First, we define the Value-at-Risk at risk level $\alpha \in (0, 1]$.
% \begin{figure}
% \centering
% \includegraphics[width=0.5\columnwidth]{figs/var_cvar.png}
% \caption{An illustration of risk measures: VaR and CVaR.\label{fig:var_cvar}}
% \end{figure}

\noindent\textbf{Value at Risk:}
\begin{equation}
\text{VaR}_{\alpha}(\mathcal{S}) = \text{inf} \{\tau\in\mathbb{R}, ~\text{Pr} [f(\mathcal{S},y)\leq \tau] \geq \alpha\}.
\label{eqn:VaR}
\end{equation}
Thus, $\text{VaR}_{\alpha}(\mathcal{S})$ denotes the left endpoint of the $\alpha$-quantile(s) of the random variable $f(\mathcal{S},y)$. The Conditional-Value-at-Risk is the expectation of this set of $\alpha$-worst cases of $f(\mathcal{S}, y)$, defined as:

\noindent\textbf{Conditional Value at Risk:}
\begin{equation}
\text{CVaR}_{\alpha}(\mathcal{S}) = \underset{y}{\mathbb{E}}[f(\mathcal{S},y)|f(\mathcal{S},y)\leq \text{VaR}_{\alpha}(\mathcal{S})].
\label{eqn:CVaR}
\end{equation} 
Figure~\ref{fig:var_cvar} shows an illustration of $\text{VaR}_{\alpha}(\mathcal{S})$ and $\text{CVaR}_{\alpha}(\mathcal{S})$. $\text{CVaR}_{\alpha}(\mathcal{S})$ is more popular than $\text{VaR}_{\alpha}(\mathcal{S})$ since it has better properties~\cite{rockafellar2000optimization}, such as  \emph{coherence}~\cite{artzner1999coherent}. 

When optimizing $\text{CVaR}_{\alpha}(\mathcal{S})$, we usually resort to an auxiliary function:
 $$H(\mathcal{S}, \tau) = \tau - \frac{1}{\alpha}\mathbb{E}[(\tau-f(\mathcal{S},y))_{+}],$$
where $(z)_+ = \max(z,0)$. We know that optimizing $\text{CVaR}_{\alpha}(\mathcal{S})$ over $\mathcal{S}$ is equivalent to optimizing the auxiliary function $H(\mathcal{S}, \tau)$ over $\mathcal{S}$ and $\tau$~\cite{rockafellar2000optimization}. The following lemmas give useful properties of the auxiliary function $H(\mathcal{S}, \tau)$.

\begin{lemma}
If $f(\mathcal{S},y)$ is normalized, monotone increasing and submodular in set $\mathcal{S}$ for any realization of $y$, the auxiliary function $H(\mathcal{S},\tau)$ is monotone increasing and submodular, but not necessarily normalized in set $\mathcal{S}$ for any given $\tau$.
\label{lem:auxiliary_function}
\end{lemma}
We provide the proofs for all the lemmas and the theorem in the  appendix.
\begin{lemma}
The auxiliary function $H(\mathcal{S},\tau)$ is concave in $\tau$ for any given set $\mathcal{S}$. 
\label{lem:auxi_concave}
\end{lemma}
\begin{lemma}
For any given set $\mathcal{S}$, the gradient of the  auxiliary function $H(\mathcal{S},\tau)$ with respect to $\tau$ fulfills: $-(\frac{1}{\alpha}-1) \leq \frac{\partial H(\mathcal{S},\tau)}{\partial \tau} \leq 1$. 
\label{lem: gradient_auxiliary_function}
\end{lemma}

\section{Problem Formulation and Case Studies} \label{sec:problem_case}
We first formulate the CVaR submodular maximization problem and then present two applications that we use as case studies. 
\subsection{Problem Formulation}\label{subsec:problem}
We consider the problem of maximizing $\text{CVaR}_{\alpha}(\mathcal{S})$ over a decision set $\mathcal{S}\subseteq \mathcal{X}$ under a matroid constraint $\mathcal{S}\in \mathcal{I}$.  We know that maximizing $\text{CVaR}_{\alpha}(\mathcal{S})$ over $\mathcal{S}$ is equivalent to maximizing the auxiliary function $H(\mathcal{S}, \tau)$ over $\mathcal{S}$ and $\tau$~\cite{rockafellar2000optimization}. Thus, we propose the maximization problem as:

\begin{problem}[CVaR Submodular Maximization]
\begin{align}
\begin{split}
&\emph{\max} ~~\tau - \frac{1}{\alpha}\mathbb{E}[(\tau-f(\mathcal{S},y))_{+}] \\
&  s.t.~~\mathcal{S} \in \mathcal{I}, \tau\in[0, +\infty).
\label{eqn:cvar_max}
\end{split}
\end{align}
\label{pro:cvar_max}
\end{problem}
% where $\Gamma$ is the upper bound of the parameter $\tau$.
Problem~\ref{pro:cvar_max} gives a risk-aware version of maximizing submodular set functions. 

\subsection{Case Studies}\label{subsec:case_study}
The risk-aware submodular maximization (Problem~\ref{pro:cvar_max}) has many applications in robotics. In the following, we describe two specific applications which we will use in the simulations. 

\subsubsection{Resilient Mobility-on-Demand}\label{subsubsec:mod}

\begin{figure}
  \centering
  \includegraphics[width=0.8\columnwidth]{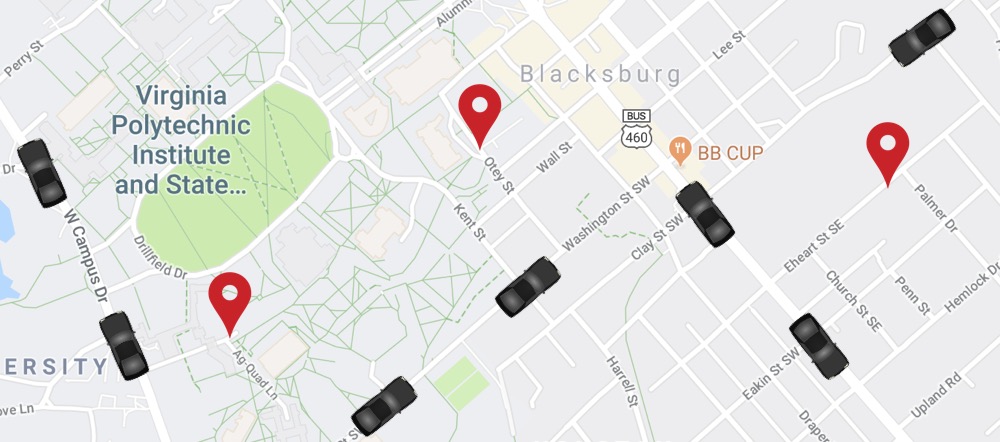}
  \caption{Mobility-on-demand with multiple demands and multiple self-driving vehicles.}
  \label{fig:MoD}
\end{figure}

Consider a mobility-on-demand problem where we assign $R$ vehicles to $N$ demand locations under arrival-time uncertainty. An example is shown in Figure~\ref{fig:MoD} where seven self-driving vehicles must be assigned to three demand locations to pick up passengers. We follow the same constraint setting in~\cite{prorok2019redundant}--- each vehicle can be assigned to at most one demand but multiple vehicles can be assigned to the same demand. Only the vehicle that arrives first is chosen for picking up the passengers. Note that the advantage of the redundant assignment to each demand is that it counters the effect of uncertainty and reduces the waiting time at demand locations~\cite{prorok2019redundant}.  This may be too conservative for consumer mobility-on-demand services but can be crucial for urgent and time-critical tasks such as delivering medical supplies~\cite{ackerman2018medical}.

Assume the arrival time for the vehicle to arrive at a demand location is a random variable. Its randomness can depend on the mean-arrival time. For example, it is possible to have a shorter path that passes through many intersections, which leads to uncertainty on arrival time. While a longer road (possibly a highway) has a lower arrival time uncertainty. Note that for each demand location, there is a set of vehicles assigned to it. The vehicle selected at the demand location is the one that arrives first. 
% We would like to minimize the total arrival time at the demand locations, $\sum_{i=1}^{N}\text{min}_{j\in \mathcal{S}_i} t_{ij}$. 
Then, this problem becomes a minimization one since we would like to minimize the  arrival time at all demand locations. We convert it into a maximization one by taking the reciprocal of the arrival time. Specifically, we use the \rev{\textit{arrival utility}} which is the reciprocal of arrival time. Instead of selecting the vehicle at the demand location with minimum arrival time, we select the vehicle with maximum arrival \rev{utility}. The arrival \rev{utility} is also a random variable, and its randomness depends on mean-arrival \rev{utility}. Denote the arrival \rev{utility} for vehicle $j\in\{1,...,R\}$ arriving at demand location $i\in\{1,...,N\}$ as $e_{ij}$. Denote the subset of robots assigned to demand location $i$  as $\mathcal{S}_i$. \rev{Then, the arrival utility for a demand location $i$ is computed as $\max_{j\in \mathcal{S}_i}e_{ij}$.} Further, we denote the assignment utility as the sum of arrival \rev{utilities} at all locations, that is, 
\begin{equation}
f(\mathcal{S}, y)  = \sum_{i\in N} \max_{j\in \mathcal{S}_i}e_{ij}
\label{eqn:fsy_assign}
\end{equation} with $\bigcup_{i=1}^{N} \mathcal{S}_i = \mathcal{S}$ and $\mathcal{S}_i \cap \mathcal{S}_k =  \emptyset, ~i, k \in \{1, \cdots, N\}$. $\mathcal{S}_i \cap \mathcal{S}_k =  \emptyset$ indicates the selected set $\mathcal{S}$ satisfies a partition matroid constraint, $\mathcal{S} \in \mathcal{I}$, which represents that each vehicle can be assigned to at most one demand.  The assignment utility $f(\mathcal{S}, y)$ is monotone submodular in $\mathcal{S}$ due to the ``max'' function. $f(\mathcal{S}, y)$ is normalized since $f(\emptyset, y)=0$. Here, we regard uncertainty as a risk. Our risk-aware assignment problem is a trade-off between assignment \rev{utility} and uncertainty. Our goal is to maximize the total \rev{utilities} at the demand locations while considering the risk from uncertainty. 
\subsubsection{Robust Environment Monitoring} \label{subsubsec:environment_monitoring}
\begin{figure}[t]
\centering{
\subfigure[Part of Virginia Tech campus from Google Earth.]
{\includegraphics[width=0.49\columnwidth]{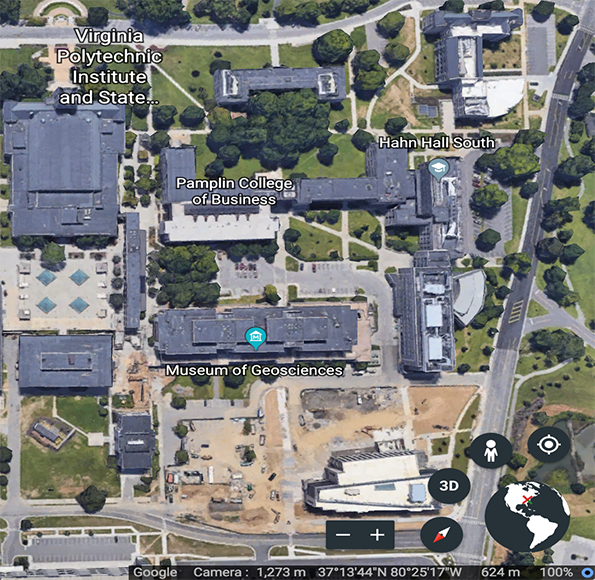}}~
\subfigure[Top view of part of a campus and ground sensor's visibility region.]
{\includegraphics[width=0.49\columnwidth]{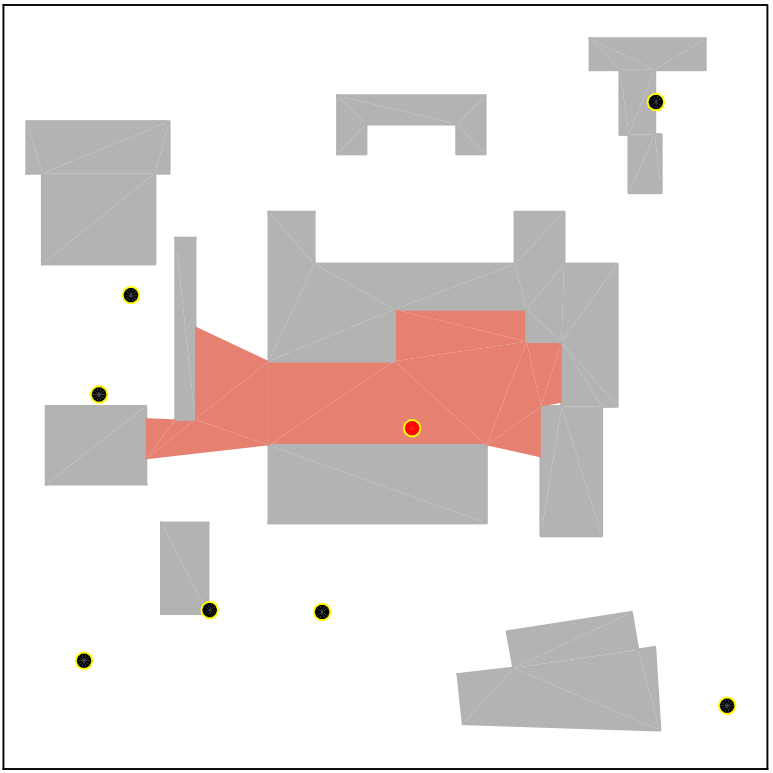}}
\caption{Campus monitoring by using a set of sensors with visibility regions.
\label{fig:environment_monitor}}
}
\end{figure}
Consider an environment monitoring problem where we monitor part of a campus with a group of ground sensors (Fig.~\ref{fig:environment_monitor}). Given a set of $N$ candidate positions $\mathcal{X}$, we would like to choose a subset of $M$ positions $ \mathcal{S}\subseteq \mathcal{X}, M\leq N$, to place visibility-based sensors to maximally cover the environment. The visibility regions of the ground sensors are obstructed by the buildings in the environment (Fig.~\ref{fig:environment_monitor}-(b)). Consider a scenario where the probability of failure of a sensor depends on the area it can cover. That is, a sensor covering a larger area has a larger risk of failure associated with it. This may be due to the fact that the same number of pixels are used to cover a larger area and therefore, each pixel covers proportionally a smaller footprint. As a result, the sensor risks missing out on detecting small objects. 

Denote the probability of success and the visibility region for each sensor $i, i \in\{1,...,M\}$ as $p_i$ and $v_i$, respectively. Thus, the polygon each sensor $i$  monitors is also a random variable. Denote this random polygon as $A_i$ and denote the selection utility as the joint coverage area of a set of sensors, $\mathcal{S}$, that is,
\begin{equation}
f(\mathcal{S}, y)  = \text{area}(\bigcup_{i=1:M} A_i), ~i\in\mathcal{S}, \mathcal{S} \subseteq \mathcal{I}. 
\label{eqn:fsy_covering}
\end{equation}
The selection utility $f(\mathcal{S}, y)$ is monotone submodular in $\mathcal{S}$ due to the overlapping area. $f(\mathcal{S}, y)$ is normalized since $f(\emptyset, y)=0$. Here, we regard the sensor failure as a risk. Our robust environment monitoring problem is a trade-off between area coverage and sensor failure. Our goal is to maximize the joint-area covered while considering the risk from sensor failure.  

\section{Algorithm and Analysis}\label{sec:alg_ana}

We present the Sequential Greedy Algorithm (SGA) for solving Problem~\ref{pro:cvar_max} by leveraging the useful properties of the auxiliary function $H(\mathcal{S}, \tau)$. The pseudo-code  is given in Algorithm~\ref{alg:sga}. SGA mainly consists of searching for the appropriate value of $\tau$ by solving a subproblem for a fixed $\tau$ under a matroid constraint. Even for a fixed $\tau$, the subproblem of optimizing the auxiliary function is NP-complete. Nevertheless, we can employ the greedy algorithm for the subproblem, and sequentially apply it for searching over all $\tau$. We explain each stage in detail next.
\subsection{Sequential Greedy Algorithm}
These are four stages in SGA: 
\paragraph{Initialization (Alg.~\ref{alg:sga}, line~\ref{line:initiliaze})} Algorithm~\ref{alg:sga} defines a storage set $\mathcal{M}$ and initializes it to be the empty set. Note that, for each specific $\tau$, we can use the greedy algorithm to obtain a near-optimal solution $\mathcal{S}^{G}$ based on the monotonicity and submodularity of the auxiliary function $H(\mathcal{S}, \tau)$. $\mathcal{M}$ stores all the $(\mathcal{S}^{G}, \tau)$ pairs when searching all the possible values of $\tau$. 

\paragraph{Searching for $\tau$ (\textbf{for} loop in Alg.~\ref{alg:sga},  lines~\ref{line:search_tau_forstart}--\ref{line:search_tau_forend})} We use a user-defined separation $\Delta$ (Alg.~\ref{alg:sga}, line~\ref{line:search_tau_separation}) to sequentially search for all possible values of $\tau$ within $[0, \Gamma]$. We use $\Gamma$ as an searching upper bound on $\tau$. That is because, when $\tau$ is greater than the maximal value of $f(\mathcal{S},y))$, the auxiliary function $H(\mathcal{S}, \tau) = (1-\frac{1}{\alpha})\tau + \frac{1}{\alpha} f(\mathcal{S}, y)$, which is non-increasing in $\tau$ given the risk level $\alpha \in (0,1]$. Thus, in order to maximize  $H(\mathcal{S}, \tau)$, we do not need to search for the cases when $\tau$ is larger than the maximum value of $f(\mathcal{S},y)$. Therefore we set the value of $\Gamma$ as the maximum value of $f(\mathcal{S},y)$. Even though $f(\mathcal{S},y)$ is a random variable, its maximum value can be set by the user based on the specific problems at hand. For example, in the environment monitoring scenario (Section~\ref{subsec:case_study}), $\Gamma$ can be computed as the total area of the environment because this is the largest area that robots can jointly cover. In the mobility-on-demand scenario (Section~\ref{subsec:case_study}), $\Gamma$ can be the highest total \rev{utilities} for reaching all demand locations (e.g., when there is no traffic congestion and vehicles travel with maximum legal speeds). We show how to find $\Gamma$ for these specific cases in Section~\ref{sec:simulation}. 

\paragraph{Greedy algorithm (Alg.~\ref{alg:sga}, lines~\ref{line:gre_empty}--\ref{line:gre_while_end})} For a specific $\tau$, say $\tau_i$, we use the greedy approach to choose  set $\mathcal{S}_{i}^{G}$. We first initialize set $\mathcal{S}_{i}^{G}$ to be the empty set (Alg.~\ref{alg:sga}, line~\ref{line:gre_empty}). Under a matroid constraint, $\mathcal{S}_{i}^{G} \in \mathcal{I}$ (Alg.~\ref{alg:sga}, line~\ref{line:gre_while_start}), we add a new element $s$ which gives the maximum marginal gain of $\hat{H}(\mathcal{S}_{i}^{G}, \tau_i)$ (Alg.~\ref{alg:sga}, line~\ref{line:gre_while_margin}) into set $\mathcal{S}_{i}^{G}$ (Alg.~\ref{alg:sga}, line~\ref{line:gre_while_pick}) in each round. 

\paragraph{Find the best pair (Alg.~\ref{alg:sga}, line~\ref{line:find_best_pair})} Based on the collection of pairs $\mathcal{M}$ (Alg.~\ref{alg:sga}, line~\ref{line:pair_collection}), we pick the pair $(\mathcal{S}_{i}^{G}, \tau_i) \in \mathcal{M}$ that maximizes $\hat{H}(\mathcal{S}_{i}^{G}, \tau_i)$ as the final solution ${S}_{i}^{G}$. We denote this value of $\tau$ by $\tau^G$. Notably, $\tau^G \in (0, \Gamma]$. 

\textit{Designing an Oracle:} Note that an oracle $\mathcal{O}$ is used to calculate the value of $H(\mathcal{S}, \tau)$. We can use a sampling-based method, as an oracle, to approximate $H(\mathcal{S}, \tau)$. Specifically, we sample $n_s$ realizations $\tilde{y} (s)$ from the distribution of $y$ and approximate  $H(\mathcal{S}, \tau)$ as $\hat{H}(\mathcal{S}, \tau) = \tau - \frac{1}{n_s\alpha}\sum_{\tilde{y}} [(\tau-f(\mathcal{S},\tilde{y}))_{+}]$. According to \cite[Lemmas 4.1-4.5]{ohsaka2017portfolio}, if the number of samples is $n_s = O(\frac{\Gamma^2}{\epsilon^2}\log \frac{1}{\delta}), ~\delta, \epsilon \in (0,1)$, the CVaR approximation error is less than $\epsilon$ with the probability at least $1-\delta$. We provide a brief proof for the number of samples in the appendix.

\textit{Observability of $y$:} SGA uses an oracle $\mathcal{O}$ to sample from $y$ for the approximation of $H(\mathcal{S}, \tau)$. Notably, $y$ is partially known  and its realizations (samples) can be taken from some historical data. For example, in the vehicle assignment problem (Section~\ref{subsec:case_study}), the samples of vehicles' travel times can be drawn from some historical traffic data set. Even though the travel time is not exactly known beforehand, when the vehicle traverses a street (or path), its travel time can be observed and updated, as shown in the online version of the mobility-on-demand (Section~\ref{sec:adaptive_submodular}).

\begin{algorithm}[t]
\caption{Sequential Greedy Algorithm (SGA)} 
\begin{algorithmic}[1]
\REQUIRE 
\begin{itemize}
\item Ground set $\mathcal{X}$ and matroid  $\mathcal{I}$
%\item Size of selected set $\kappa$. 
\item User-defined risk level $\alpha \in (0, 1]$
\item Range of the parameter $\tau \in [0, \Gamma]$ and discretization stage $\Delta \in (0, \Gamma]$
\item An oracle $\mathcal{O}$ that approximates $H(\mathcal{S}, \tau)$ as $\hat{H}(\mathcal{S}, \tau)$
\end{itemize}
\ENSURE 
\begin{itemize}
\item Selected set $\mathcal{S}^{G}$ and corresponding parameter $\tau^{G}$
\end{itemize}

\STATE $\mathcal{M}\leftarrow\emptyset$ \label{line:initiliaze}
\STATE \textbf{for} $ ~i=\{0,1,\cdots, \ceil{\frac{\Gamma}{\Delta}}\}$ \textbf{do} \label{line:search_tau_forstart}
\STATE ~~~~$\tau_i = i\Delta$\label{line:search_tau_separation}
\STATE ~~~~$\mathcal{S}_{i}^{G}\leftarrow\emptyset$ \label{line:gre_empty}
\STATE~~~~\textbf{while} \label{line:gre_while_start}
$\mathcal{S}_{i}^{G} \in \mathcal{I}$ \textbf{do}
\STATE~~~~~~~~$s = \underset{s\in \mathcal{X}\setminus \mathcal{S}_{i}^{G}, \mathcal{S}_{i}^{G}\cup \{s\} \in \mathcal{I}}{\text{argmax}}~\hat{H}((\mathcal{S}_{i}^{G}\cup \{s\}), \tau_i) - \hat{H}(\mathcal{S}_{i}^{G}, \tau_i)$ \label{line:gre_while_margin}
\STATE~~~~~~~~$\mathcal{S}_{i}^{G}\leftarrow \mathcal{S}_{i}^{G} \cup \{s\}$ \label{line:gre_while_pick}
\STATE~~~~\textbf{end while}\label{line:gre_while_end}
\STATE~~~~ $\mathcal{M} = \mathcal{M} \cup \{(\mathcal{S}_{i}^{G}, \tau_i)\}$\label{line:pair_collection}
\STATE \textbf{end for}\label{line:search_tau_forend}
\STATE $(\mathcal{S}^{G}, \tau^{G}) = \underset{(\mathcal{S}_{i}^{G}, \tau_i) \in \mathcal{M}}{\text{argmax}}~{\hat{H}(\mathcal{S}_{i}^{G}, \tau_i)}$ \label{line:find_best_pair}

\end{algorithmic}
\label{alg:sga}
\end{algorithm}

% \begin{algorithm}[t]
% \caption{Greedy Algorithm for the Subproblem of Problem~\ref{pro:cvar_max}.}
% \begin{algorithmic}[1]
% \REQUIRE 
% \begin{itemize}
% \item $\{(\mathcal{S}_{i}^{G}, \tau_i)\}$ pairs, $\mathcal{M}$. 
% \item An oracle $\mathcal{O}$ that evaluates $H(\mathcal{S}, \tau)$
% \end{itemize}
% \ENSURE Selected set $\mathcal{S}^{G}$
% ~
% \STATE \textbf{if} $\mathcal{S}^{G} \in \mathcal{I}$\label{line:matroid_constraint}\\
% \STATE~~~~$\mathcal{S}^{G} = \underset{(\mathcal{S}_{i}^{G}, \tau_i) \in \mathcal{M}}{\text{argmax}}~{H(\mathcal{S}_{i}^{G}, \tau_i) - H(\emptyset, \tau_i)/2}$ \label{line:matroid_evaluate_2}
% \STATE \textbf{else} \label{line:no_matroid}
% \STATE~~~~$\mathcal{S}^{G} = \underset{(\mathcal{S}_{i}^{G}, \tau_i) \in \mathcal{M}}{\text{argmax}}~{H(\mathcal{S}_{i}^{G}, \tau_i) - H(\emptyset, \tau_i)/e}$\label{line:nomatroid_evaluate_e}
% \STATE \textbf{end}
% \end{algorithmic}
% \label{alg:subroutine}
% \end{algorithm}

\subsection{Performance Analysis of SGA}\label{subsec:analysis_alg}
Let $\mathcal{S}^{G} \in \mathcal{I}$ and $\tau^{G} \in (0, \Gamma]$ be the set and the scalar chosen by \text{SGA}, and let the $\mathcal{S}^{\star}$ and $\tau^{\star}$ be the set and the scalar chosen by the optimal solution. Our analysis uses the curvature, $c_f \in [0, 1]$ of function $H(\mathcal{S}, \tau)$  that is submodular in set $\mathcal{S}$. Then, we get our our main result.
\begin{theorem}
Algorithm 1 yields a bounded approximation for Problem 1. Specifically, 
% \PRT{Algorithm 1 yields a bounded approximation for Problem 1. Specifically, }
\begin{align}
{H}(\mathcal{S}^{G},{\tau}^{G}) \geq & ~\frac{1}{1+c_f}(H(\mathcal{S}^{\star},\tau^{\star}) - \Delta) \nonumber\\
& ~-\frac{c_f}{1+c_f}\tau^G(\frac{1}{\alpha} -1) - \epsilon,
\label{eqn:appro_bound_theorem}
\end{align}
with probability at
least $1-\delta$, when the number of samples, $n_s = O(\frac{\Gamma^2}{\epsilon^2}\log \frac{1}{\delta}),~ \delta, \epsilon \in (0,1)$. 
The computational time is $O(\ceil{\frac{\Gamma}{\Delta}} |\mathcal{X}|^{2} n_s)$ where $\Gamma$ and $\Delta$ are the searching upper bound on $\tau$ and searching separation parameter, $|\mathcal{X}|$ is the cardinality of the ground set $\mathcal{X}$.
\label{thm:appro_bound_compu}
\end{theorem}
% \PRT{Why is the previous sentence inside the theorem? Instead, before the theorem, just add a line or two.. our analysis uses the notion of curvature, kf which measures how modular the function is. Please see the appendix..}

SGA gives $1/(1+c_f)$ approximation of the optimal with three additional approximation errors. One approximation error comes from the searching separation $\Delta$. We can make this error arbitrarily small 
% \PRT{What is ``very''? Do you mean arbitrarily small or is it a specific number? Make it precise.} 
by setting $\Delta$ to be close to zero with the cost of increasing the number of search operations $\ceil{\frac{\Gamma}{\Delta}} $. The second one is the sampling error $\epsilon$ from the oracle. Evidently, $\epsilon$ can be reduced by increasing the number of samples $n_s$. The third one comes from the additive term, 
\begin{equation}
H^{\text{add}} = \frac{c_f}{1+c_f}\tau^{G}(\frac{1}{\alpha} -1),
\label{eqn:add_error}
\end{equation}
which depends on the curvature $c_f$ and the risk level $\alpha$. When the risk level $\alpha$ is very small, this error is very large which means  SGA may not give a good performance guarantee of the optimal. However, if the function $H(\mathcal{S}, \tau)$ is close to modular in $\mathcal{S}$ ($c_f \to 0$), this error is close to zero.  Notably, when $c_f \to 0$ and $\Delta, \epsilon \to 0$, SGA gives a near-optimal solution, i.e., ${H}(\mathcal{S}^{G},{\tau}^{G}) \to H(\mathcal{S}^{\star},\tau^{\star})$.

Next, we briefly describe the proof of Theorem~\ref{thm:appro_bound_compu}. 
We start with the proof of approximation ratio, then go to the analysis of the computational time. We need to prove on some structural lemmas beforehand. Let $\mathcal{S}_{i}^{\star}$ be the optimal set for a specific $\tau_i$ that maximizes $H(\mathcal{S}, \tau=\tau_i)$ and $H(\mathcal{S}^{\star}, \tau^{\star})$ be the maximum value of $H(\mathcal{S}, \tau)$. We sequentially search for $\tau \in [0, \Gamma]$ with a searching separation $\Delta$.  We first describe the relationship between the maximum $H(\mathcal{S}, \tau)$ founded by searching $\tau$ with a searching separation $\Delta$ and $H(\mathcal{S}^{\star}, \tau^{\star})$. 
\begin{lemma}
\begin{equation}
\underset{i\in \{0,1,\cdots, \ceil{\frac{\Gamma}{\Delta}}\}}{\emph{\max}}H(\mathcal{S}_{i}^{\star}, \tau_i) \geq H(\mathcal{S}^{\star}, \tau^{\star}) -\Delta. 
\label{eqn:tau_sstar_delta}
\end{equation}
\label{lem:tau_sstar}
\end{lemma}

Next, we describe the relationship between $H(\mathcal{S}_{i}^{G}, \tau)$ with set $\mathcal{S}_{i}^{G}$ selected by our greedy approach and $H(\mathcal{S}_{i}^{\star}, \tau)$ for each $\tau_i$. 
\begin{lemma}
% \begin{eqnarray}
% H(\mathcal{S}_{i}^{G},{\tau}_i) &\geq& \max \{\frac{1}{1+c_f}, 1-\frac{1}{e}\}H(\mathcal{S}_{i}^{\star},\tau_i)\nonumber\\
% &-& \text{min}\{\frac{c_f}{1+c_f}, \frac{1}{e}\} \tau_i(\frac{1}{\alpha} -1).
% \label{eqn:appro_bound}
% \end{eqnarray}
\begin{eqnarray}
H(\mathcal{S}_{i}^{G},{\tau}_i) \geq  \frac{1}{1+c_f} H(\mathcal{S}_{i}^{\star},\tau_i) - \frac{c_f}{1+c_f} \tau_i(\frac{1}{\alpha} -1).
\label{eqn:appro_bound}
\end{eqnarray}
where $c_f$ is the curvature of the function $H(\mathcal{S},\tau)$ in $\mathcal{S}$ with a matroid constraint $\mathcal{I}$. 
\label{lem:rela_gre_opt_tau}
\end{lemma}

Finally, using Lemma~\ref{lem:tau_sstar} and Lemma~\ref{lem:rela_gre_opt_tau} and considering the sampling error, we can find relationship between $H(\mathcal{S}^{G},{\tau}^{G})$ (our greedy value) and $H(\mathcal{S}^{\star}, \tau^{\star})$ (the optimal value). We summarize this relationship in the Theorem~\ref{thm:appro_bound_compu} with a detailed proof in the appendix.

% \PRT{This section is too short. You need to give some intuition before the lemmas and how they are combined.}
\section{Adaptive Risk-Aware Submodular Maximization}  \label{sec:adaptive_submodular}
In this section, we study an adaptive version of the risk-aware submdoular maximization described in Problem~\ref{pro:cvar_max}. This is motivated by the real-time (or online) applications where either the external environment or the internal robot state is dynamic and changing. For example, the stochastic travel times of the vehicles are changing due to real-time traffic conditions in the mobility-on-demand and the sensors can fail at some point for environment monitoring. In these dynamic scenarios, if the planning is only scheduled at the initial time step  and kept fixed (i.e., \textit{one-step} planning), the robots (e.g., vehicles or sensors) may not be able to fully utilize the real-time information for achieving a better team performance.

Instead, the robots may need to replan to adapt to real-time situations. For instance, in the mobility-on-demand scenario, as the vehicles move they can use real-time traffic conditions to update the assignment for achieving shorter arrival time (i.e., higher \rev{utility}) at demand locations as the actual travel time information is revealed.  

One intuitive online strategy is to replan at all time steps. Clearly, the \textit{all-step} planning can achieve a better performance than the \textit{one-step} planning, since it fully utilizes the real-time information at all time steps. However, planning at all time steps can be costly (e.g., in terms of the computing resources or energy). Crucially, planning at all time steps may not be necessary. 

Thus, instead of either \textit{one-step} planning or \textit{all-step} planning, we seek replanning strategies that can make risk-aware decisions with as fewer replanning efforts.  
% We measure the performance of a replanning strategy by comparing it to an \textit{offline greedy} strategy $\mathcal{S}^{o}$ which is assumed to know all the information (e.g., travel time or sensor failure) exactly in the planning process and uses the standard greedy algorithm~\cite{nemhauser1978analysis,fisher1978analysis} to compute the planning strategy. 
Formally, our bicriteria optimization problem is stated as follows. 

\begin{problem}[Adaptive Risk-Aware Submodular Maximization]
Given a risk level $\alpha \in (0,1]$, find a risk-aware sequential replanning strategy that decides when to replan $\{k^{\texttt{rp}}\}$ and how to replan $\{\mathcal{S}(k^{\texttt{rp}})\}$ to
%  \begin{align}
%  \begin{split}
%  & \emph{\max} ~f(\{\mathcal{S}(k^{\texttt{rp}})\})  \label{eqn:max_rp} \\
% %   & \emph{\max} ~\frac{f(\{\mathcal{S}(k^{\texttt{rp}})\})}{f(\mathcal{S}^{o})}  \label{eqn:max_rp} \\
%  & s.t. ~\mathcal{S}(k^{\texttt{rp}}) \in \mathcal{I}, k^{\texttt{rp}} \in [1, T),
%  \end{split}
%   \end{align}
%   and 
% \begin{align}
% \begin{split}
%  & \emph{\min} ~|\{k^{\texttt{rp}}\}| \label{eqn:min_rp} \\
%  & s.t. ~|\{k^{\texttt{rp}}\}| \geq 1, k^{\texttt{rp}} \in [1, T).\\
%  \end{split}
%  \end{align}
\rev{
% The bicriteria optimization problem can be rewritten as
 \begin{align}
 \begin{split}
 & \emph{\max} ~(f(\{\mathcal{S}(k^{\texttt{rp}})\}), -|\{k^{\texttt{rp}}\}|) \\
 & s.t. ~\mathcal{S}(k^{\texttt{rp}}) \in \mathcal{I},\\
 & ~~~~~|\{k^{\texttt{rp}}\}| \geq 1, \\
 & ~~~~~k^{\texttt{rp}} \in [1, T),
 \label{eqn:max_min_rp}
 \end{split}
 \end{align}
\noindent where $[1, T)$ denotes the timespan of the planning process; $k^{\texttt{rp}}$ denotes a replanning time step and $\mathcal{S}(k^{\texttt{rp}})$ denotes the replanning strategy (e.g., replacement or reassignment) at step $k^{\texttt{rp}}$; $\{\mathcal{S}(k^{\texttt{rp}})\}$ denotes a sequence of replanning strategies corresponding to a sequence of replanning time steps $\{k^{\texttt{rp}}\}$ during $[1, T)$;
% $\mathcal{S}(k)$ is the planning strategy at time step $k$; 
% $f(\mathcal{S}(k), y)$ and $f(\mathcal{S}^o)$ denote the stochastic utility of the planning strategy $\mathcal{S}(k)$ and the deterministic utility of the offline greedy strategy $\mathcal{S}^o$, respectively; 
$f(\{\mathcal{S}(k^{\texttt{rp}})\})$ denotes the utility obtained by executing the sequence of replanning strategies $\{\mathcal{S}(k^{\texttt{rp}})\}$; $|\{k^{\texttt{rp}}\}|$ denotes the number of times for replanning, and $|\{k^{\texttt{rp}}\}| \geq 1$ indicates there must exist an initial planning to start the process at time step $1$. 
}
\label{pro:replan}
\end{problem}

% Notably, the process utility function $f(\mathcal{S}(k))$ is deterministic without the random variable $y$. That is because, $f(\mathcal{S}(k))$ captures the planning performance when the process finishes at the final step $k^{\texttt{f}}$. Clearly, at the final step, all information is revealed and there is no uncertainty. For example, in the mobility-on-demand case, when all demand locations are reached by vehicles, we know exactly the arrival \rev{utility} (or arrival time) at all demand locations. Similarly, when all sensors fail (final step), the area covered at all previous steps are known in the environment monitoring case.  

% However, before the final step, the robots can only make decisions based on their real-time beliefs (e.g., stochastic travel time or sensor working status). Thus, when $t \in [k^{\texttt{i}}, k^{\texttt{f}})$, the robots need to optimize the stochastic utility function $f(\mathcal{S}(k),y)$ for planning by using their beliefs at step $k$, which is exactly an instance of Problem~\ref{pro:cvar_max}. This is why we are given a risk level $\alpha$ for Problem~\ref{pro:replan}.  

\rev{In Problem~\ref{pro:replan}, we aim to simultaneously maximize the process utility $f(\{\mathcal{S}(k^{\texttt{rp}})\})$ and minimize the number of replanning times $|\{k^{\texttt{rp}}\}|$. We seek a sequential replanning strategy $(\{k^{\texttt{rp}}\}^\star, \{\mathcal{S}(k^{\texttt{rp}})\}^\star)$ that is nondominated or Pareto optimal for Problem~\ref{pro:replan}. Here, the Pareto optimality depicts a situation where no solution other than  $(\{k^{\texttt{rp}}\}^\star, \{\mathcal{S}(k^{\texttt{rp}})\}^\star)$ could improve the process utility or reduce the number of replanning times without reducing the former or raising  the latter, or there is no scope for either improving the process utility or reducing the number of replanning times. 
% For any feasible solution $(\{k^{\texttt{rp}}\}, \{\mathcal{S}(k^{\texttt{rp}})\})$, we seek its Pareto Pareto-dominated solution $(\{k^{\texttt{rp}}'\}, \{\mathcal{S}(k^{\texttt{rp}})\}')$ (if there exists one) such that
% \begin{align*}
%   & f(\{\mathcal{S}(k^{\texttt{rp}})\}) \leq f(\{\mathcal{S}(k^{\texttt{rp}})\}), |\{k^{\texttt{rp}}\}| \leq |\{k^{\texttt{rp}}\}|, ~\text{and}\\
%   & f(\{\mathcal{S}(k^{\texttt{rp}})\}) \leq f(\{\mathcal{S}(k^{\texttt{rp}})\}), \text{or} |\{k^{\texttt{rp}}\}| \leq |\{k^{\texttt{rp}}\}|, ~\text{and}
% \end{align*}
}

Particularly, the process utility function $f(\{\mathcal{S}(k^{\texttt{rp}})\})$ captures the performance of the sequence of the replanning strategies $\{\mathcal{S}(k^{\texttt{rp}})\}$ during $[1, T)$. For example, in the mobility-on-demand, it can be the total arrival \rev{utilities} of all demand locations when these demand locations are reached by vehicles. Similarly, for environmental monitoring, it can be the joint area covered at the end of the execution horizon. The process utility is \textit{deterministic} since it measures the performance when the process is done (i.e., at the final step $T$). Essentially the optimal process utility would be for a clairvoyant strategy that will know exactly how the scenario will unfold a priori. However, to compute the replanning strategy $\mathcal{S}(k)$ at some intermediate step $k \in [1, T)$, the robots need to optimize a stochastic utility $f(\mathcal{S}(k), y)$, as the robots can only use the information  (e.g., stochastic travel time or sensor working status) revealed so far and thus have uncertain knowledge of the future.

% and thus only their current beliefs (e.g., stochastic travel time or sensor working status) are available.     
% The process utility is stochastic because the information is revealed in real time, and thus the robots can only use their current beliefs (e.g., stochastic travel time or sensor working status) for computing the utility. 
% This also means, at each specific step $k$, $f(\mathcal{S}(k), y)$ turns out to be $f(\mathcal{S}(k), y)$ with randomness induced by the uncertain knowledge of the future (i.e., from step $k$ to final step $k^{\texttt{f}}$). Therefore, at each step $k$, the maximization subproblem (Eq.~\ref{eqn:max_rp}) in Problem~\ref{pro:replan} is just an instance of Problem~\ref{pro:cvar_max}. 

The optimal solution to Problem~\ref{pro:replan} can be found if we assume the robots know all the future information exactly, e.g., what the travel times will be or which sensors will fail. However, this is not the realistic scenario where the robots have no knowledge of what will exactly happen in the future. Thus, finding the optimal replanning strategy for Problem~\ref{pro:replan} is extremely difficult or even impossible. Notably, even with \textit{all-step} planning, there is no guarantee on obtaining the optimal value of the process  utility. That is because, using the information unveiled so far, the planning is myopic and could mislead the planning in the future steps. Also, minimizing the number of replanning times simultaneously makes Problem~\ref{pro:replan} even more challenging. 

Instead, we relax the hardness of Problem~\ref{pro:replan} by fixing the strategy to decide \textit{how to replan} $\mathcal{S}(k^{\texttt{rp}})$ at every replanning step $k^{\texttt{rp}}$ and build on it to present a heuristic solution that is two-fold. Specifically, we leverage  SGA (Algorithm~\ref{alg:sga}) to compute $\mathcal{S}(k^{\texttt{rp}})$. \rev{That is because, at each replanning step $k^{\texttt{rp}} \in [1, T)$, we aim to design a risk-aware strategy to optimize the stochastic utility $f(\mathcal{S}(k^{\texttt{rp}}), y)$ with $y$ representing the uncertainty of the environment in the future steps (i.e., from the current step $k^{\texttt{rp}}$ to the end step $T$). That is, 
    \begin{align*}
      \begin{split}
        & \max ~~\tau - \frac{1}{\alpha}\mathbb{E}[(\tau-f(\mathcal{S}(k^{\texttt{rp}}),y))_{+}] \\
        &  s.t.~~\mathcal{S}(k^{\texttt{rp}}) \in \mathcal{I}, \tau\in[0, +\infty),
      \end{split}
    \end{align*}
which} is exactly an instance of Problem~\ref{pro:cvar_max} and SGA gives a bounded approximation (Theorem~\ref{thm:appro_bound_compu}). 
% That is, at each replanning step $k^{\texttt{rp}} \in [k^{\texttt{i}}, k^{\texttt{f}})$, the robots use SGA to compute a replanning strategy $\mathcal{S}(k^{\texttt{rp}})$ based on their current beliefs. 

Then the question left is to decide \textit{when to replan} to minimize $|\{k^{\texttt{rp}}\}|$. To this end, we exploit an event-driven mechanism where the planning is updated or recomputed only when certain conditions are met. For environmental monitoring, there is an obvious solution that the (well-working) sensors update the replacement by SGA only when some sensor fails. In this way, they can maintain a good monitoring performance without replacing themselves at all time steps. Such a strategy is not obvious for the mobility-on-demand case. Therefore, we explicitly describe the online version of the mobility-on-demand and design a triggering reassignment strategy to determine when to replan. 

\vspace{1mm}

\paragraph*{Online Mobility-on-Demand and Triggering Vehicle Reassignment}
% ~\label{subsec:tri_assign}
We study an online version of the resilient mobility-on-demand described in Section~\ref{subsec:case_study}. Similarly, we assign $R$ vehicles to serve $N$ demand locations. The assignment process starts at time step $1$ when vehicles start traveling toward their assigned demand locations by the initial vehicle-demand assignment. The process ends at time step $T$ when all the demand locations are reached by the vehicles. We denote the stochastic arrival time for each vehicle $j$ to each demand location $i$ at step $k,~k\in[1, T)$ as $t_{ij}(k)$ with the mean $\bar{t}_{ij}(k)$ and variance $\texttt{var}(t_{ij})(k)$. Then the corresponding arrival \rev{utility} at step $k$ is computed as $e_{ij}(k) = 1/t_{ij}(k)$. The arrival time $t_{ij}(k)$ and arrival \rev{utility} $e_{ij}(k)$ can be updated based on the real-time vehicles' positions and traffic conditions (e.g., real-time waiting times at intersections). Similarly, for each demand location, we assign a set of vehicles to it and only the vehicle that arrives first at the demand location is chosen. The objective is to develop a replanning strategy that maximizes the total arrival \rev{utilities} (or equivalently, minimizes the total arrival times) of all demand locations when all these demand locations are reached by vehicles while having the minimum number of times for replanning, as captured by Problem~\ref{pro:replan}.      

% Then, based on Equation~\ref{eqn:fsy_assign}, the stochastic utility function of the entire assignment process, $f(\mathcal{S},y)$ in Problem~\ref{pro:replan}, can be written as, 
% \begin{equation}
% f(\mathcal{S},y)  = \sum_{i\in N} \text{max}_{j\in \mathcal{S}_i(k^{\texttt{f}})}e_{ij}(k^{\texttt{f}}),
% \label{eqn:online_assignment}
% \end{equation}
% where $\mathcal{S}_i(k^{\texttt{f}})$ and $e_{i}(k^\texttt{f})$ denote the set of vehicles assigned to demand location $i$ and its arrival \rev{utility} at final step $k^\texttt{f}$. This process utility captures the total stochastic arrival \rev{utility} over all demand locations at final step $k^{\texttt{f}}$. 

% With the assignment process utility described in Equation~\ref{eqn:online_assignment}, 

Toward this end, we design an adaptive triggering assignment strategy (ATA) in Algorithm~\ref{alg:online_tri_assign_gen}. Specifically, for deciding \textit{how to reassign} at each reassignment step $k^{\texttt{rp}}$, we use SGA to update the vehicle-demand assignment by using the vehicles' real-time arrival \rev{utilities} $e_{ij}(k)$. This is based on the fact that, at each intermediate step $k\in[1, T)$, the problem turns out to be maximizing the CVaR of the stochastic utility function,
\begin{equation}
f(\mathcal{S}(k), y)  = \sum_{i\in N} \text{max}_{j\in \mathcal{S}_i(k)}e_{ij}(k) \nonumber
\label{eqn:fsy_assign_k}
\end{equation}
with $\mathcal{S}(k)$ and $\mathcal{S}_i(k)$ denoting the vehicle-demand assignment and the set of vehicles assigned to demand location $i$ at step $k$. We know optimizing the CVaR of $f(\mathcal{S}(k), y)$ is an instance of Problem~\ref{pro:cvar_max} and SGA yields a solution with a bounded approximation (Theorem~\ref{thm:appro_bound_compu}).  Before that, we need to decide \textit{when to reassign} (i.e., the reassignment step $k^{\texttt{rp}}$) which is tackled by ATA (Alg.~\ref{alg:online_tri_assign_gen}) where an event-driven mechanism is 
adopted to update the assignment only when certain conditions are met. 

\begin{algorithm}[t]
\caption{Adaptive Triggering Assignment (ATA)}
\begin{algorithmic}[1]
\STATE  $\mathcal{S} = \text{SGA}(\{e_{ij}(1), \alpha)$
\label{line:initial_assign_gen}

    \WHILE{not all $N$ demand locations are reached}
    \label{line:while_start_gen}
        \STATE Each vehicle $j$ travels toward assigned demand $i$
        \label{line: travel_to_goal_gen}
        
        \STATE Each vehicle $j$ updates $t_{ij}(k)$, $e_{ij}(k)$
        \label{line: update_t_e_gen}
        
       \STATE Check for all demand locations $d_i, i\in\{1,\cdots, N\}$
        \label{line:check_demand_gen}
        \IF{$\exists~ \bar{t}_{ij}(k) \leq \gamma \bar{t}_{ij'}(k)$ and $\texttt{var}(t_{ij}(k)) \leq \texttt{var}(t_{ij'})(k)$, $j, j'\in \mathcal{S}_i(k)$}
        \label{line:tri_cond_gen}
             \STATE $\mathcal{S} = \text{SGA}(\{e_{ij}(k)\}), \alpha)$
             \label{line:tri_assign_gen}
        \ENDIF
        \label{line:tri_if_end}
        
    \ENDWHILE
    \label{while_end_gen}
\end{algorithmic}
\label{alg:online_tri_assign_gen}
\end{algorithm}

A general version of ATA is described in Algorithm~\ref{alg:online_tri_assign_gen}. Please find a specific implementation of ATA for the mobility-on-demand in street networks in the appendix (Alg.~\ref{alg:online_tri_assign}).\footnote{For clarity, we drop the time indices $k$ of notations for online mobility-on-demand in Alg.~\ref{alg:online_tri_assign} and the corresponding simulations in Section~\ref{subsubsec:tri_assign_trig}.} ATA mainly contains two assignment steps---initial assignment (Alg.~\ref{alg:online_tri_assign_gen}, line~\ref{line:initial_assign_gen}) and triggering assignment (Alg.~\ref{alg:online_tri_assign_gen}, line~\ref{line:tri_assign_gen}). Once the initial assignment (Alg.~\ref{alg:online_tri_assign_gen}, line~\ref{line:initial_assign_gen}) is done, each vehicle $i$ travels toward its (firstly) assigned demand location $j$ (Alg.~\ref{alg:online_tri_assign_gen}, line~\ref{line: travel_to_goal_gen}) and updates its arrival time $t_{ij}(k)$ and corresponding arrival \rev{utility} $e_{ij}(k)$ at each time step $k$ (Alg.~\ref{alg:online_tri_assign_gen}, line~\ref{line: update_t_e_gen}). 

The reassignment is triggered by checking the arrival times of vehicles assigned to each demand location $d_i$ (Alg.~\ref{alg:online_tri_assign_gen}, line~\ref{line:check_demand_gen}). 
% \PRT{it would be good to give the equation from the algorithm here.} 
Specifically, among the set of vehicles $\mathcal{S}_i(k)$ assigned to demand location $i$, if the arrival time of one vehicle $j\in \mathcal{S}_i(k)$ is less than that of one other vehicle $j'\in\mathcal{S}_i(k)$ on average and has smaller uncertainty (Alg.~\ref{alg:online_tri_assign_gen}, line~\ref{line:tri_cond_gen}), a new assignment is triggered (Alg.~\ref{alg:online_tri_assign_gen}, line~\ref{line:tri_assign_gen}). Because, in this case, there is no need to continue assigning vehicle $j'$ to demand location $i$. Instead, it would be helpful to trigger a new assignment to assign vehicle $j'$ to other demand locations. Here, $\gamma \in (0,1)$ is the triggering ratio. It  regulates the triggering frequency, e.g., a larger $\gamma$ triggers more assignments. ATA terminates when all demand locations are reached by vehicles (Alg.~\ref{alg:online_tri_assign_gen}, line~\ref{line:while_start_gen}). 
% In the next section, we evaluate the effect of $\gamma$ in terms of the arrival time when all demand locations are reached and the number of assignments.  

It is intuitive that ATA can (probably) achieve a shorter arrival time than $\textit{one-step}$ assignment since the real-time information (e.g., vehicles' positions and traffic conditions) can be utilized for updating assignments. One simple example could be that if a demand location is reached by some vehicle, the other vehicles assigned to it and yet have not arrived should be reassigned to other demand locations. However, scheduling reassignment at every step, as in \textit{all-step} reassignment, may not be necessary or even detrimental in some cases. That is because, the reassignment based on the information revealed so far (before the final step) is myopic and could be misleading for the future assignments. Clearly, \textit{all-step} reassignment with the most number of reassignments increases the chance of misleading. In the next section, we illustrate the comparisons of these three assignment algorithms and the effect of the triggering ratio $\gamma$.

\section{Simulations}\label{sec:simulation}
We perform numerical simulations to verify the performance of SGA in risk-aware mobility-on-demand and robust environment monitoring, and the performance of adaptive triggering assignment (ATA) in online risk-aware mobility-on-demand. Our code is available online.\footnote{\url{https://github.com/raaslab/risk_averse_submodular_selection}} 
% \rev{We omit the simulations for adaptive triggering placement (ATP) since it is a natural, straightforward strategy and its performance can be easily interpreted by comparing to the \textit{all-step} and \textit{one-step} placement (Section~\ref{subsec:tri_place}).  Also, the author has done a similar study in~\cite{ramachandran2019resilient}. Please refer to~\cite{ramachandran2019resilient} for the details.}

\subsection{Resilient Mobility-on-Demand under Arrival Time Uncertainty}\label{subsec:mod_demand} 
% Because $\text{max}_{j\in \mathcal{S}_i}~\tilde{e}_{ij}$ is submodular in $\mathcal{S}_i$ due to the submodularity of the ``max" function. And the summation of the submodular functions is still submodular. Note that, if no robot is assigned to the location, the arriving \rev{utility} is zero, i.e.,  $f(\emptyset, y))= 0$, which means the assignment utility function is normalized.
We consider assigning $R = 6$ supply vehicles to $N=4$ demand locations in a 2D environment. The positions of the demand locations and the supply vehicles are randomly generated within a square environment of 10-unit side length. Denote the Euclidean distance between demand location $i\in \{1,..., N\}$ and vehicle position $j\in \{1,...,R\}$ as $d_{ij}$. Based on the distribution discussion of the arrival \rev{utility} distribution in Section~\ref{subsec:case_study}, we assume each arrival \rev{utility} $e_{ij}$ has a uniform distribution with its mean proportional to the reciprocal of the distance between demand $i$ and vehicle $j$. Furthermore, the uncertainty is higher if the mean \rev{utility} is higher. Note that, the algorithm can handle other, more complex, distributions for arrival times. We use a uniform distribution for ease of exposition.
%(Fig.~\ref{fig:MoD}-(b)). 
Specifically, denote the mean of $e_{ij}$  as $\bar{e}_{ij}$ and set $\bar{e}_{ij} = 10/d_{ij}$. We model the arrival \rev{utility} distribution to be a uniform distribution as follows: 
$$e_{ij} =[\bar{e}_{ij} - \bar{e}_{ij}^{2.5}/\max\{\bar{e}_{ij}\}, \bar{e}_{ij} + \bar{e}_{ij}^{2.5}/\max\{\bar{e}_{ij}\}], $$ 
where $\max\{\bar{e}_{ij}\} = \max_{i,j}e_{ij}, i\in \{1,..., N\}, j\in \{1,...,R\}$. 

From the assignment utility function (Eq.~\ref{eqn:fsy_assign}),  for any realization of $y$, say $\tilde{y}$, $$f(\mathcal{S}, \tilde{y}) : = \sum_{i\in N} \text{max}_{j\in \mathcal{S}_i}\tilde{e}_{ij}$$ where $\tilde{e}_{ij}$ indicates one realization of $e_{ij}$. If all vehicle-demand pairs are independent from each other, $y$ models a multi-independent uniform distribution. We sample $n_s$ times from underlying multi-independent uniform distribution of $y$ and approximate the auxiliary function $H(\mathcal{S}, \tau)$ 
% $$H(\mathcal{S}, \tau) = \tau - \frac{1}{\alpha}\mathbb{E}[(\tau-f(\mathcal{S},y))_{+}] = \tau - \frac{1}{\alpha}\mathbb{E}[(\tau-\sum_{i\in N} \text{max}_{j\in \mathcal{S}_i}e_{ij})_{+}],$$
as $$\hat{H}(\mathcal{S}, \tau) = \tau - \frac{1}{n_s\alpha}\sum_{\tilde{y}} [(\tau-\sum_{i\in N} \text{max}_{j\in \mathcal{S}_i}\tilde{e}_{ij})_{+}].$$ We set the upper bound of the parameter $\tau$ as $\Gamma = N \max (\tilde{e}_{ij}), i = \{1,...N\}, j= \{1,...,R\}$, to make sure $\Gamma - f(\mathcal{S},y) \geq 0$. We set the searching separation for $\tau$ as $\Delta =1$.

\begin{figure}[t]
\centering{
\subfigure[The value of $H(\mathcal{S}^{G}, \tau^{G})$ with respect to several risk levels $\alpha$.]
{\includegraphics[width=0.65\columnwidth]{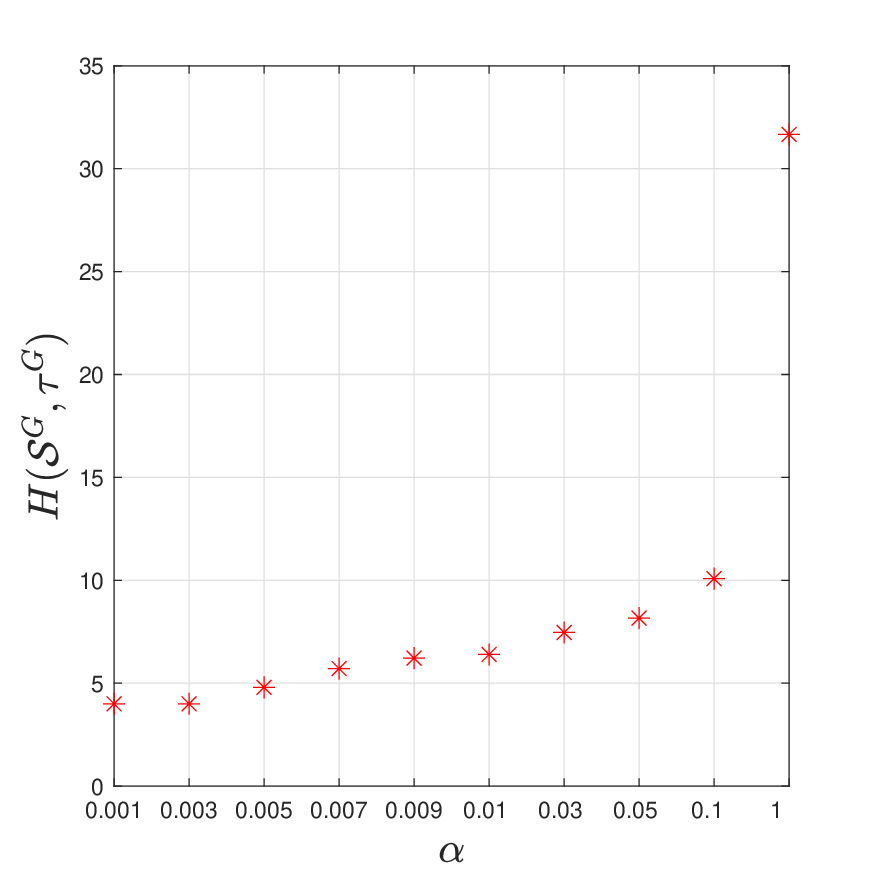}}
\subfigure[Function $H(\mathcal{S}^{G}, \tau)$ with respect to several risk levels $\alpha$.]
{\includegraphics[width=0.65\columnwidth]{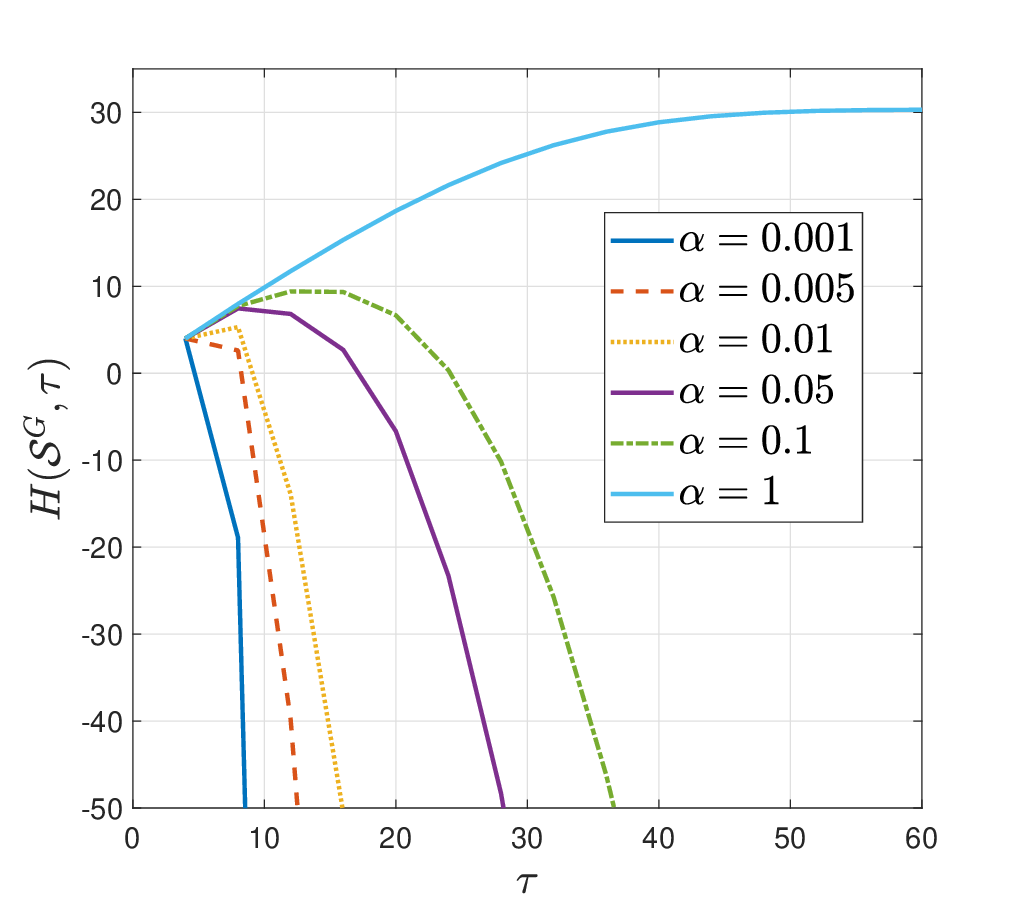}}
\caption{The value of $H(\mathcal{S}, \tau)$ by SGA with respect to different risk levels. }
\label{fig:hsg_alpha_tau}
}
\end{figure}

After receiving the pair $(\mathcal{S}^{G}, \tau^{G})$ from  SGA,  we plot the value of $H(\mathcal{S}^{G}, \tau^{G})$  and  $H(\mathcal{S}^{G}, \tau)$ with respect to different risk levels $\alpha$ in Figure~\ref{fig:hsg_alpha_tau}. Figure~\ref{fig:hsg_alpha_tau}-(a) shows that $H(\mathcal{S}^{G}, \tau^{G})$ increases when $\alpha$ increases. This suggests that SGA correctly maximizes $H(\mathcal{S}, \tau)$. Figure~\ref{fig:hsg_alpha_tau}-(b) shows that $H(\mathcal{S}^{G}, \tau)$ is concave or piecewise concave, which is consistent with the property of $H(\mathcal{S}, \tau)$.

% \begin{figure}[t]
% \centering
% \begin{minipage}[b]{.49\textwidth}
%   \centering
%   \includegraphics[width=.49\linewidth]{figs/pdf_fs_drassign_2.eps}
%   \caption{Distribution of the assignment utility $f(\mathcal{S}^{G}, y)$ by SGA.}
%   \label{fig:dis_assign_utility}
% \end{minipage} 
% \hfill
% \begin{minipage}[b]{.49\textwidth}
%   \centering
%   \includegraphics[width=.49\linewidth]{figs/H_add_assign.eps}
%   \caption{Additive term in the approximation ratio with respect to risk level $\alpha$.}
%   \label{fig:add_err_assign}
% \end{minipage}
% \end{figure}

% \begin{figure}
%   \centering
%   \includegraphics[width=0.65\columnwidth]{figs/pdf_fs_drassign_2.eps}
%   \caption{Distribution of the assignment utility $f(\mathcal{S}^{G}, y)$ by SGA.}
%   \label{fig:dis_assign_utility}
% \end{figure}

% \begin{figure}
%   \centering
%   \includegraphics[width=0.65\columnwidth]{figs/H_add_assign_1.eps}
%   \caption{Additive term (Eq.~\ref{eqn:add_error}) in the approximation ratio with respect to risk level $\alpha$.}
%   \label{fig:add_err_assign}
% \end{figure}

% \begin{figure}
%   \centering
%   \includegraphics[width=0.68\columnwidth]{figs/gre_bf_comp_assign.eps}
%   \caption{Comparison of the values of $H(\mathcal{S}, \tau)$ obtained by SGA (i.e., $H(\mathcal{S}^{G}, \tau^{G})$) and obtained by the sequential brute-force algorithm with respect to small risk levels $\alpha \in [0.001, 0.1]$ in 30 trials.}
%   \label{fig:gre_bf_compare_assign}
% \end{figure}

\begin{figure*}[t]
\minipage{0.316\textwidth}
\centering
  \includegraphics[width=\linewidth]{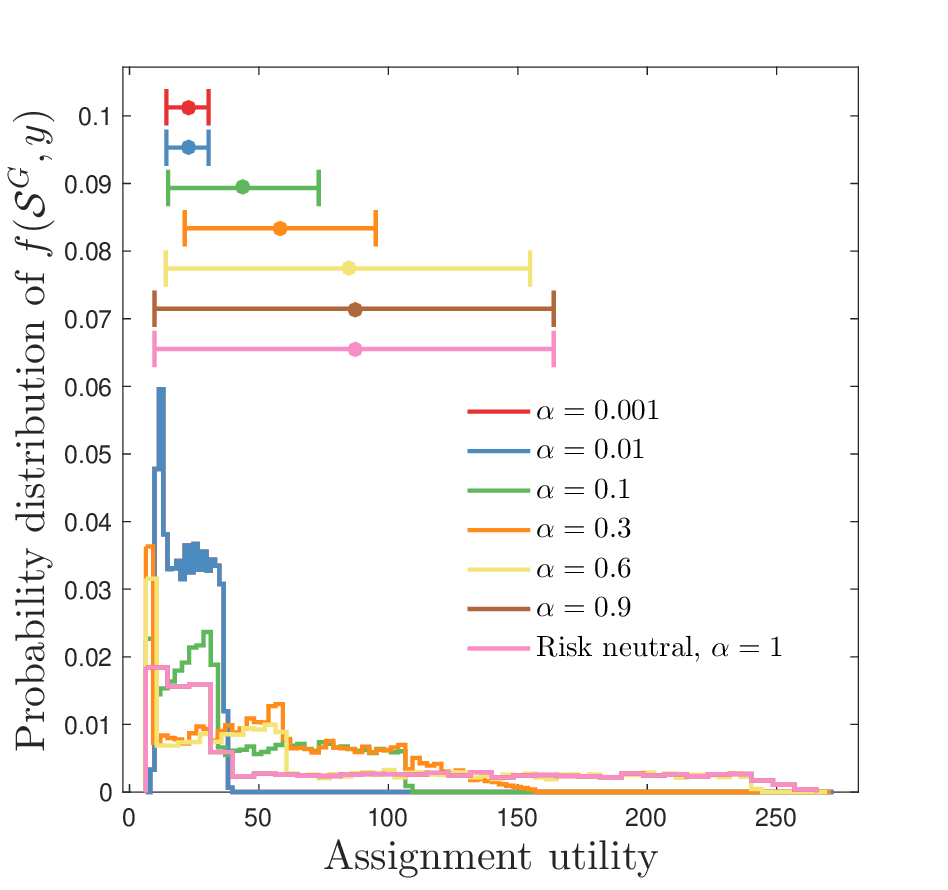}
    \caption{Distribution of the assignment utility $f(\mathcal{S}^{G}, y)$ by SGA.}
  \label{fig:dis_assign_utility}
\endminipage ~~~
\minipage{0.285\textwidth}
\centering
  \includegraphics[width=\linewidth]{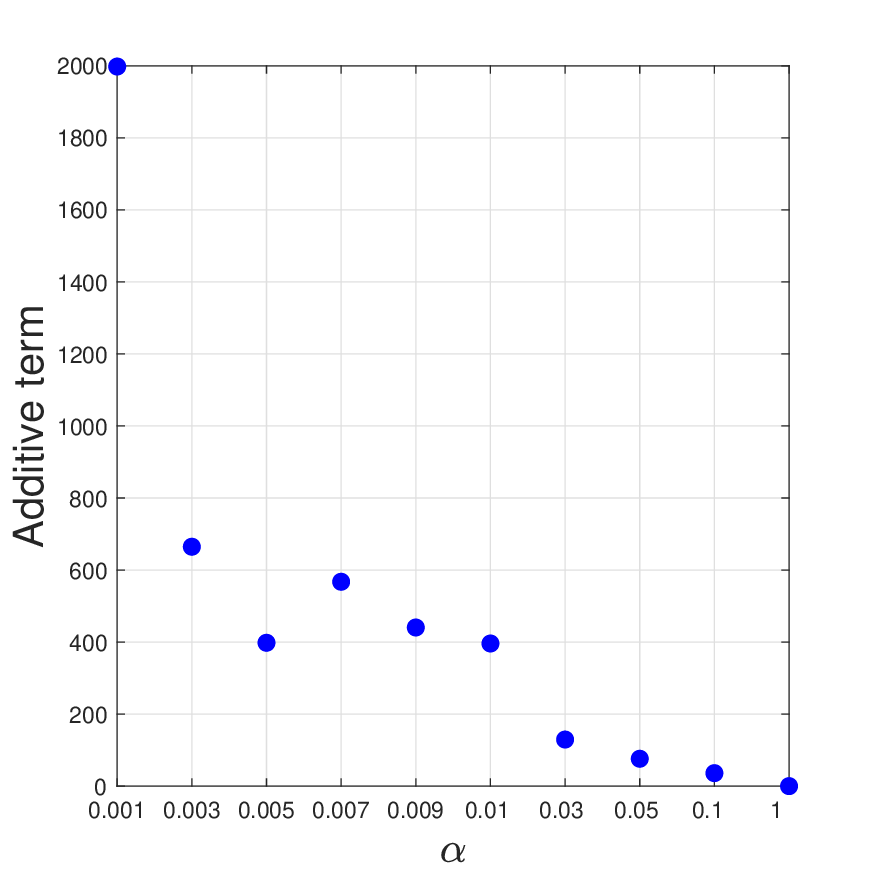}
  \caption{Additive term (Eq.~\ref{eqn:add_error}) in the approximation ratio with respect to risk level $\alpha$.}
  \label{fig:add_err_assign}
\endminipage ~~~
\minipage{0.28\textwidth}%
\centering
  \includegraphics[width=\linewidth]{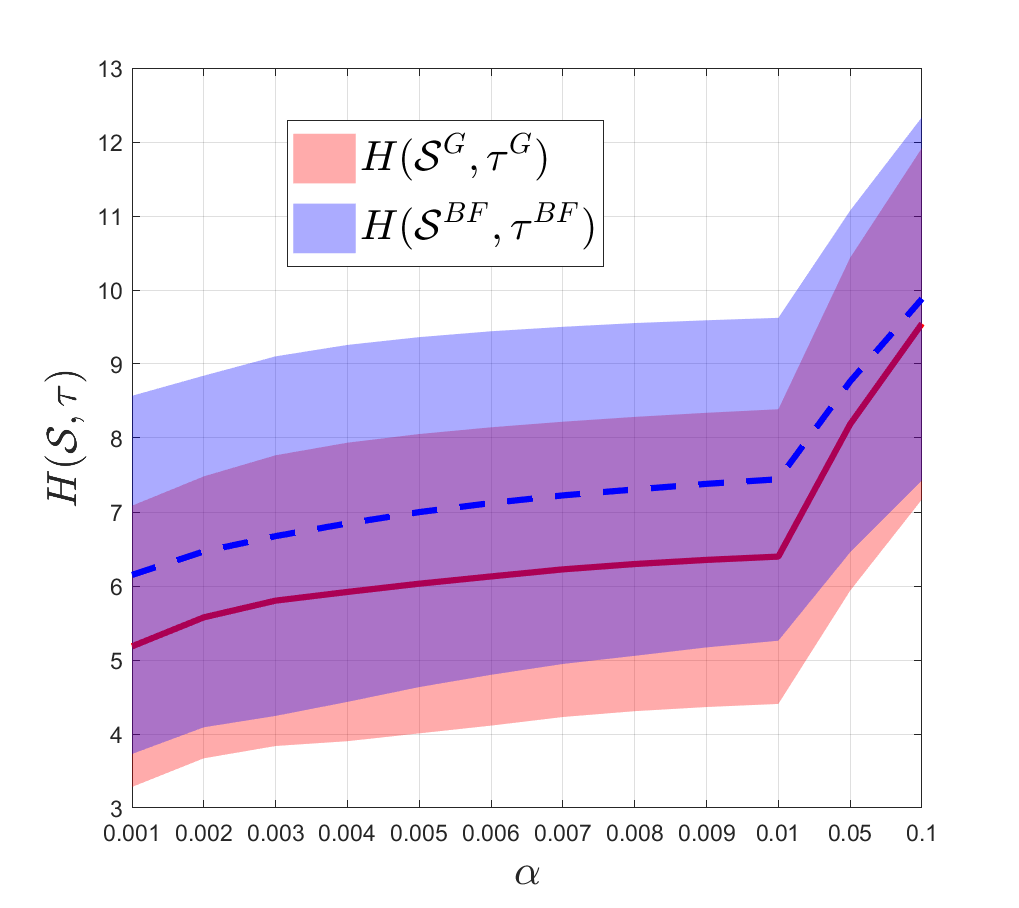}
  \caption{Comparison of the values of $H(\mathcal{S}, \tau)$ obtained by SGA (i.e., $H(\mathcal{S}^{G}, \tau^{G})$) and obtained by the sequential brute-force algorithm with respect to small risk levels $\alpha \in [0.001, 0.1]$ in 30 trials.}
  \label{fig:gre_bf_compare_assign}
\endminipage
\end{figure*}

We plot the distribution of assignment utility, $$f(\mathcal{S}^{G}, y) =  \sum_{i\in N} \text{max}_{j\in \mathcal{S}_{i}^{G}}e_{ij}$$ in Figure~\ref{fig:dis_assign_utility} by sampling $n_s=1000$ times from the underlying distribution of  $y$.
$\mathcal{S}_i^{G}$ is a set of vehicles assigned to demand location $i$ by SGA. $\mathcal{S}^{G} = \bigcup_{i=1}^{N} \mathcal{S}_{i}^{G}$. When the risk  level $\alpha$ is small, vehicle-demand pairs with low \rev{utilities} (equivalently, low uncertainties) are selected. This is because a small risk level indicates the assignment is conservative and only willing to take a little risk. Thus, lower \rev{utility} with lower uncertainty is assigned to avoid the risk induced by the uncertainty. In contrast, when $\alpha$ is large, the assignment is allowed to take more risk to gain more assignment utility. Vehicle-demand pairs with high \rev{utilities} (equivalently, high uncertainties) are selected in such a case. 

Figure~\ref{fig:add_err_assign} depicts the additive approximation error (Eq.~\ref{eqn:add_error}) of SGA with respect to the risk level $\alpha$. It shows that when $\alpha$ is close to zero, SGA has a large approximation error and the error vanishes quickly when $\alpha$ increases (especially when $\alpha \geq 0.1$). Since SGA may not give a good solution with a large additive approximation error, it is worth evaluating how well SGA does when the risk level $\alpha$ is very small. 

To this end, for $\alpha \in[0.001, 0.1]$, we compare SGA with a \textit{sequential brute-force algorithm} that uses the brute-force search as a subroutine instead of the greedy algorithm (lines~\ref{line:gre_while_start}-\ref{line:gre_while_end}) in Algorithm~\ref{alg:sga}, as shown in Figure~\ref{fig:gre_bf_compare_assign}. That is, for each specific $\tau_i$, we use the brute-force search to enumerate all possible sets $\mathcal{S}$ to find the optimal set $\mathcal{S}_i^{BF}$ that maximizes $H(\mathcal{S}, \tau_i)$. For example, in this mobility-on-demand case with 6 vehicles assigned to 4 demand locations, the total number of feasible assignments can be computed as the total number of ways to partition 6 vehicles into 4 disjoint sets (i.e., 65, see \cite{Countsubsets2020}) multiplying by the total number of permutations of 4 demand locations (i.e., 24), that is $65 \times 24 = 1560$. Over all these 1560 possible assignments, we select out the assignment set $\mathcal{S}_i^{BF}$ that maximizes $H(\mathcal{S}, \tau_i)$ at each $\tau_i$. Then we choose the best pair $(\mathcal{S}^{BF}, \tau^{BF})$ that gives the maximum value of $H(\mathcal{S}, \tau)$, denoted as $H(\mathcal{S}^{BF}, \tau^{BF})$, over all $(\mathcal{S}_i^{BF}, \tau_i)$ pairs (collected when searching for all possible values of $\tau$). By referring to the proof of the approximation ratio of SGA (proof of Theorem~\ref{thm:appro_bound_compu} in the appendix), it is easy to verify that the sequential brute-force algorithm gives the following guarantee for maximizing  $H(\mathcal{S}, \tau)$, 
\begin{equation}
{H}(\mathcal{S}^{BF},{\tau}^{BF}) \geq ~H(\mathcal{S}^{\star},\tau^{\star}) - \Delta - \epsilon,
\label{eqn:SBFA}
\end{equation}
with probability at least $1-\delta$, when the number of samples, $n_s = O(\frac{\Gamma^2}{\epsilon^2}\log \frac{1}{\delta}),~ \delta, \epsilon \in (0,1)$. Different from SGA, the sequential brute-force algorithm does not have the additive approximation error. 

We compare the performance of SGA and the sequential brute-force algorithm using the same positions of 6 vehicles and 4 demand locations randomly generated in 30 trials. Figure~\ref{fig:gre_bf_compare_assign} shows that SGA works comparatively with the sequential brute-force algorithm for maximizing $H(\mathcal{S}, \tau)$ on average when $\alpha$ is close to zero. This suggests SGA can still perform relatively well when it has a large additive approximation error due to very small risk levels. 

\begin{figure*}
\centering{
% \hspace*{-5.5mm}
\subfigure[Assignment when $\alpha = 0.01$.]{\includegraphics[width=0.65\columnwidth]{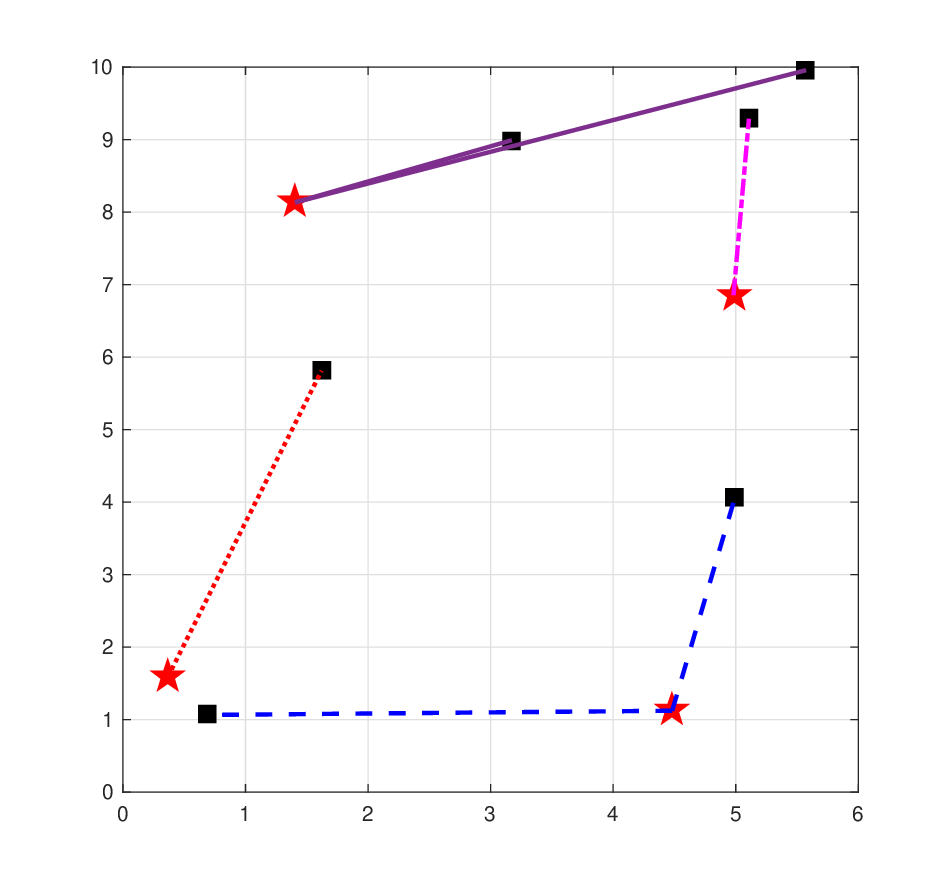}}
\subfigure[Assignment when $\alpha = 1$ ($\text{Risk-neutral})$). ]{\includegraphics[width=0.65\columnwidth]{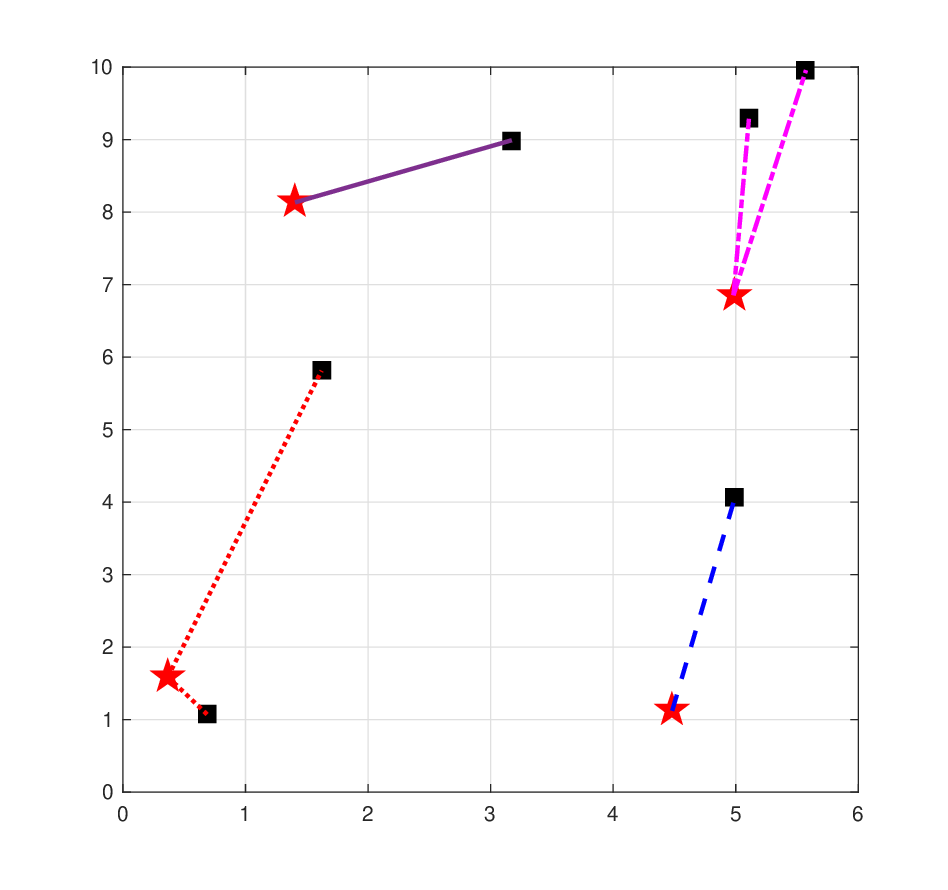}}
\subfigure[Assignment utility distributions at $\alpha=0.01$ and $\alpha =1$.]{\includegraphics[width=0.65\columnwidth]{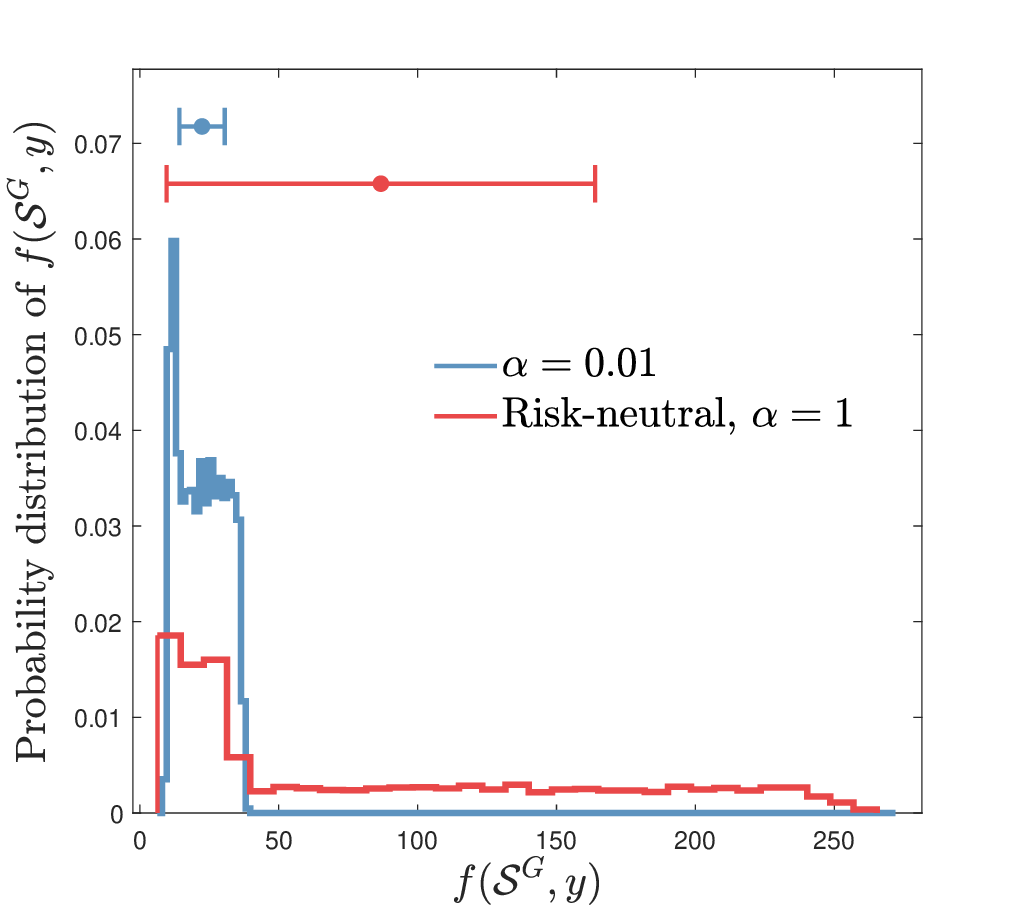}}
\caption{Assignments and utility distributions by SGA with two extreme risk level values. The red solid star represents the demand location. The black solid square represents the vehicle position. The line between the vehicle and the demand represents an assignment. 
\label{fig:two_exteme_assign}}}
\end{figure*}

We also compare SGA (using CVaR measure) with the greedy algorithm (using the expectation, i.e., risk-neutral measure~\cite{prorok2019redundant}) in Figure~\ref{fig:two_exteme_assign}. Note that risk-neutral measure is a special case of $\text{CVaR}_{\alpha}$ measure when $\alpha =1$. We give an illustrative example of the assignment by SGA for two extreme risk levels, $\alpha = 0.01$ and $\alpha = 1$. When $\alpha$ is small ($\alpha = 0.01$), the assignment is  conservative and thus farther vehicles (with lower \rev{utility} and lower uncertainty) are assigned to each demand (Fig.~\ref{fig:two_exteme_assign}-(a)). In contrast, when $\alpha =1$, nearby vehicles (with higher \rev{utility} and higher uncertainty) are selected for the demands (Fig.~\ref{fig:two_exteme_assign}-(b)). Even though the mean value of the assignment utility distribution is larger at $\alpha =1$ than at $\alpha =0.01$, it is exposed to the risk of receiving lower utility since the mean-std bar at $\alpha =1$ has smaller left endpoint than the mean-std bar at $\alpha =0.01$ (Fig.~\ref{fig:two_exteme_assign}-(c)).

\subsection{Robust Environment Monitoring}~\label{subsec:sim_envirnment}
% where the random variable $y$ models a multi-independent Bernoulli distribution from the working of a set of sensors $\mathcal{S}$. For any realization of $y$, $\tilde{y}$, the joint covering area $f(\mathcal{S}, \tilde{y})$ is a submodular function in $\mathcal{S}$ by considering the area overlapping redundancy. $f(\mathcal{S}, \tilde{y})$ is also normalized since $f(\emptyset, \tilde{y}) = 0$. 
We consider selecting $M=4$ locations from $N=8$ candidate locations to place sensors in the environment (Fig.~\ref{fig:environment_monitor}). Denote the area of the free space as $v^{\text{free}}$. The positions of $N$ candidate locations are randomly generated within the free space $v^{\text{free}}$. We calculate the visibility region for each sensor $v_i$ by using the VisiLibity library~\cite{VisiLibity1:2008}.  Based on the sensor model discussed in Section~\ref{subsec:case_study}, we set the probability of success for each sensor $i$ as 
$$p_i = 1 - v_i/v^{\text{free}},$$
and model the working of each sensor as a Bernoulli distribution with $p_i$ probability of success and $1-p_i$ probability of failure. Thus the random polygon monitored by each sensor $A_i$, follows the distribution 
\begin{equation}
    \begin{cases}
      \text{Pr}[A_i = v_i] = p_i,\\      
      \text{Pr}[A_i = 0] = 1- p_i.
    \end{cases} 
    \label{eqn:cover_dis}
\end{equation}

From the selection utility function (Eq.~\ref{eqn:fsy_covering}),  for any realization of $y$, say $\tilde{y}$, $$f(\mathcal{S}, y)  = \text{area}(\bigcup_{i=1:M} \tilde{A}_i),$$ where $\tilde{A}_{i}$ indicates one realization of $A_{i}$ by sampling $y$. If all sensors are independent of each other, we can model the working of a set of sensors as a multi-independent Bernoulli distribution. We sample $n_s = 1000$ times from the underlying multi-independent Bernoulli distribution of $y$ and  approximate the auxiliary function $H(\mathcal{S}, \tau)$ 
% \begin{equation}
% H(\mathcal{S}, \tau) = \tau - \frac{1}{\alpha}\mathbb{E}[(\tau-f(\mathcal{S},y))_{+}] = \tau - \frac{1}{\alpha}\mathbb{E}[(\tau-\bigcup_{i=1:M} A_i)_{+}],
% \label{eqn:hfun_area}
% \end{equation}
as $$\hat{H}(\mathcal{S}, \tau) = \tau - \frac{1}{n_s\alpha}\sum_{\tilde{y}} [(\tau-\bigcup_{i=1:M} \tilde{A}_i)_{+}].$$ We set the upper bound for the parameter $\tau$ as the area of all the free space $v^{\text{free}}$ and set the searching separation for $\tau$ as $\Delta =1$. 

\begin{figure}[t]
\centering{
\subfigure[The value of $H(\mathcal{S}^{G}, \tau^{G})$ with respect to different risk levels $\alpha$.]
{\includegraphics[width=0.65\columnwidth]{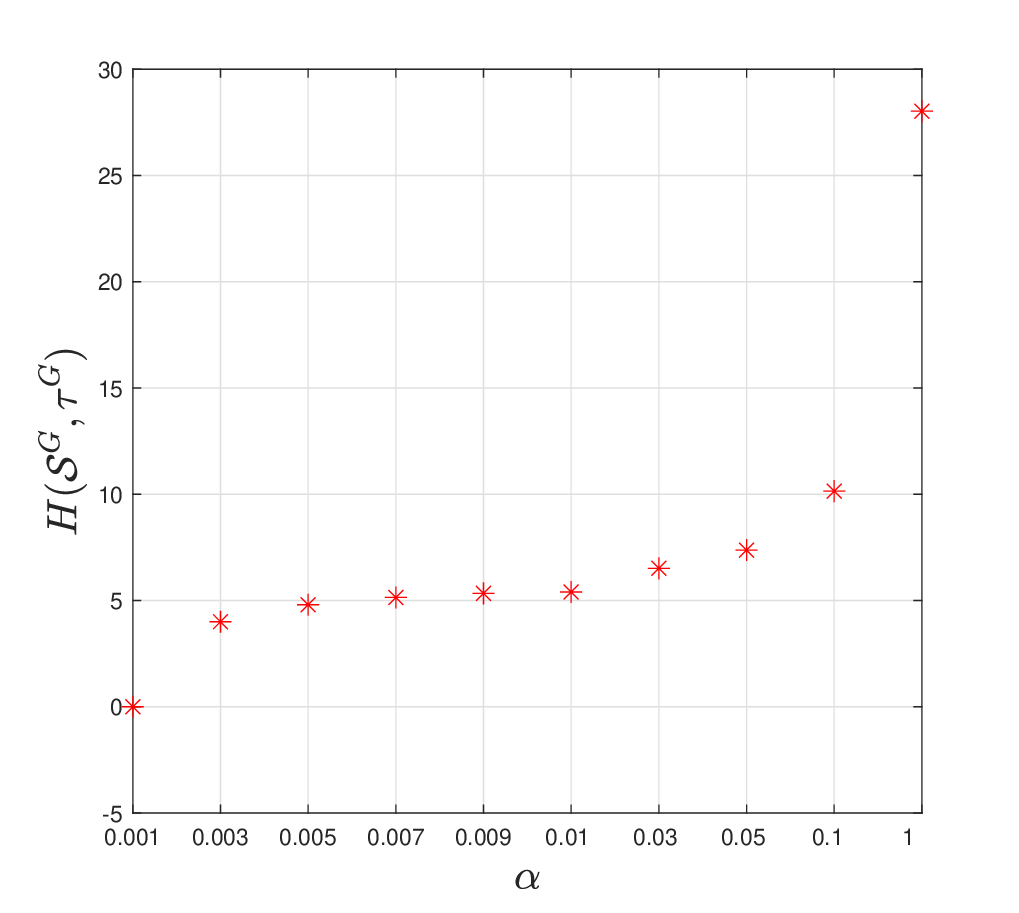}}
\subfigure[Function $H(\mathcal{S}^{G}, \tau)$ with respect to several risk levels $\alpha$.]
{\includegraphics[width=0.65\columnwidth]{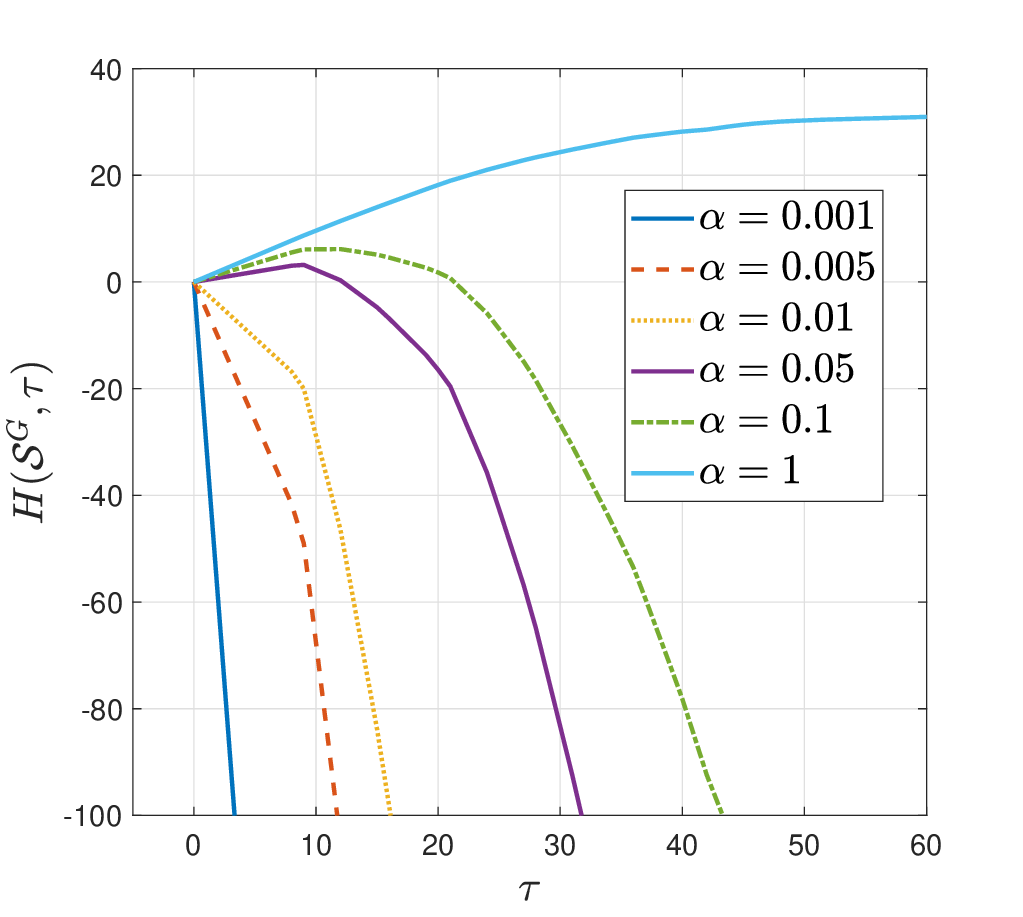}}
\caption{The value of $H(\mathcal{S}, \tau)$ by SGA with respect to different risk levels.  }
\label{fig:hsg_alpha_tau_vis}
}
\end{figure}

We use SGA to find the pair $(\mathcal{S}^{G}, \tau^{G})$ with respect to several risk levels $\alpha$. We plot the value of $H(\mathcal{S}^{G}, \tau^{G})$ for several risk levels in Figure~\ref{fig:hsg_alpha_tau_vis}-(a).  A larger risk level gives a larger $H(\mathcal{S}^{G}, \tau^{G})$, which means the pair $(\mathcal{S}^{G}, \tau^{G})$ found by SGA correctly maximizes
$H(\mathcal{S}, \tau)$ with respect to the risk level $\alpha$. Moreover, we plot functions $H(\mathcal{S}^{G}, \tau)$ for several risk levels $\alpha$ in Figure~\ref{fig:hsg_alpha_tau_vis}-(b). Note that $\mathcal{S}^{G}$ is computed by SGA at each $\tau$. For each $\alpha$, $H(\mathcal{S}^{G}, \tau)$ shows the concavity or piecewise concavity of function $H(\mathcal{S}, \tau)$. 

% \begin{figure}
%   \centering
%   \includegraphics[width=0.65\columnwidth]{figs/pdf_fs_area_2.eps}
%   \caption{Distribution of the selection utility $f(\mathcal{S}^{G}, y)$ by SGA.}
%   \label{fig:dis_select_utility}
% \end{figure}

% \begin{figure}
%   \centering
%   \includegraphics[width=0.65\columnwidth]{figs/H_add_vis_1.eps}
%   \caption{Additive term (Eq.~\ref{eqn:add_error}) in the approximation ratio with respect to risk level $\alpha$.}
%   \label{fig:add_err_vis}
% \end{figure}

% \begin{figure}
%   \centering
%   \includegraphics[width=0.68\columnwidth]{figs/gre_bf_comp_sensor.eps}
%   \caption{Comparison of the values of $H(\mathcal{S}, \tau)$ obtained by SGA (i.e., $H(\mathcal{S}^{G}, \tau^{G})$) and obtained by the sequential brute-force algorithm with respect to small risk levels $\alpha \in [0.001, 0.1]$ in 30 trials.}
%   \label{fig:gre_bf_compare_sensor}
% \end{figure}

\begin{figure*}[t]
\minipage{0.316\textwidth}
\centering
  \includegraphics[width=\linewidth]{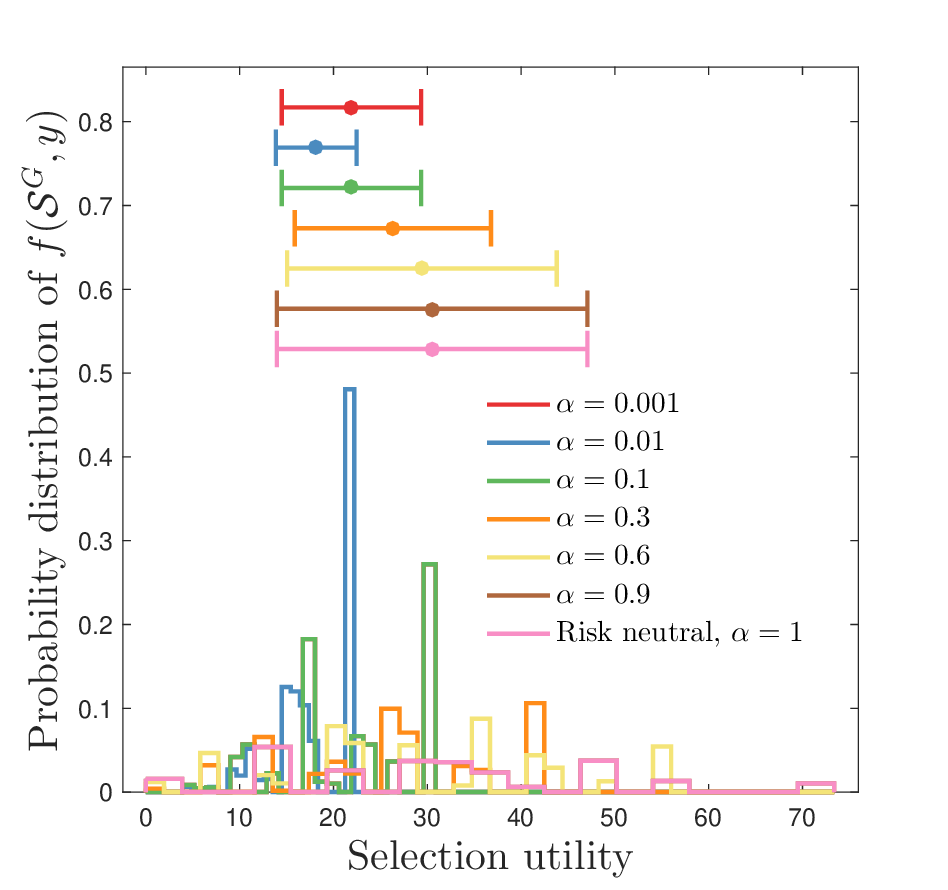}
      \caption{Distribution of the selection utility $f(\mathcal{S}^{G}, y)$ by SGA.}
  \label{fig:dis_select_utility}
\endminipage ~~~
\minipage{0.32\textwidth}
\centering
  \includegraphics[width=\linewidth]{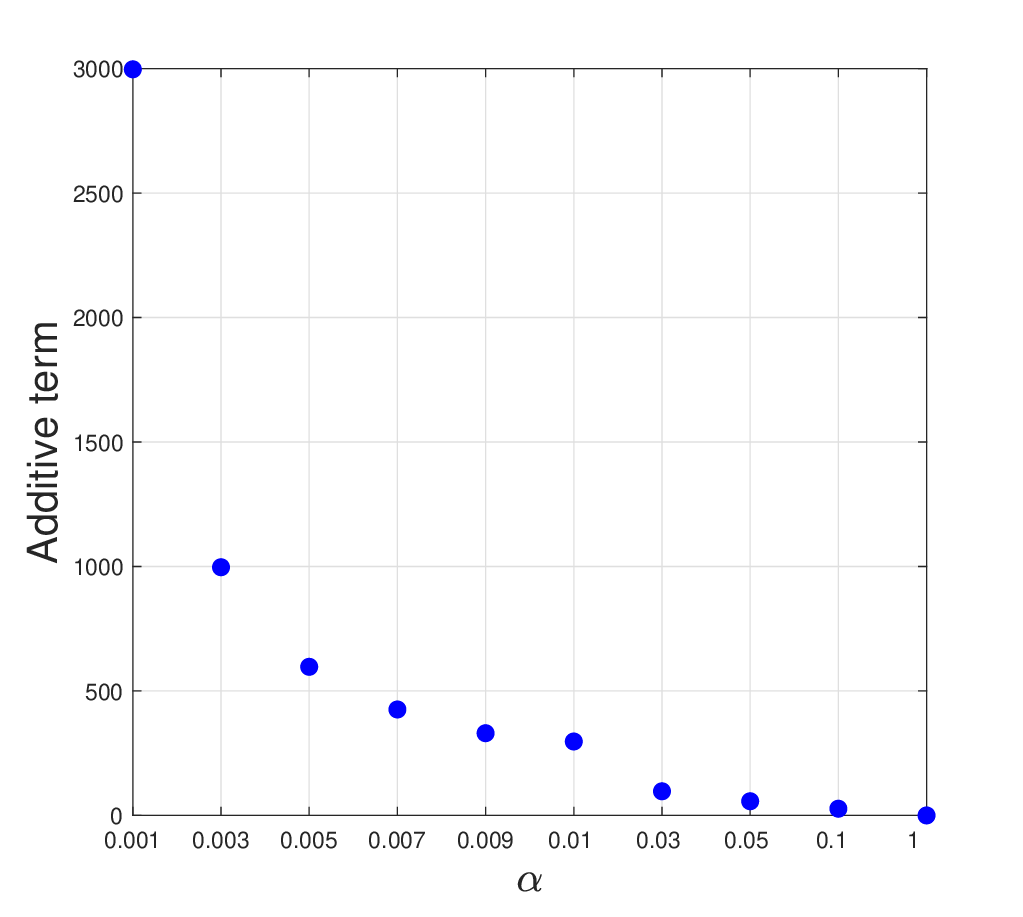}
  \caption{Additive term (Eq.~\ref{eqn:add_error}) in the approximation ratio with respect to risk level $\alpha$.}
  \label{fig:add_err_vis}
\endminipage ~~~
\minipage{0.28\textwidth}%
\centering
  \includegraphics[width=\linewidth]{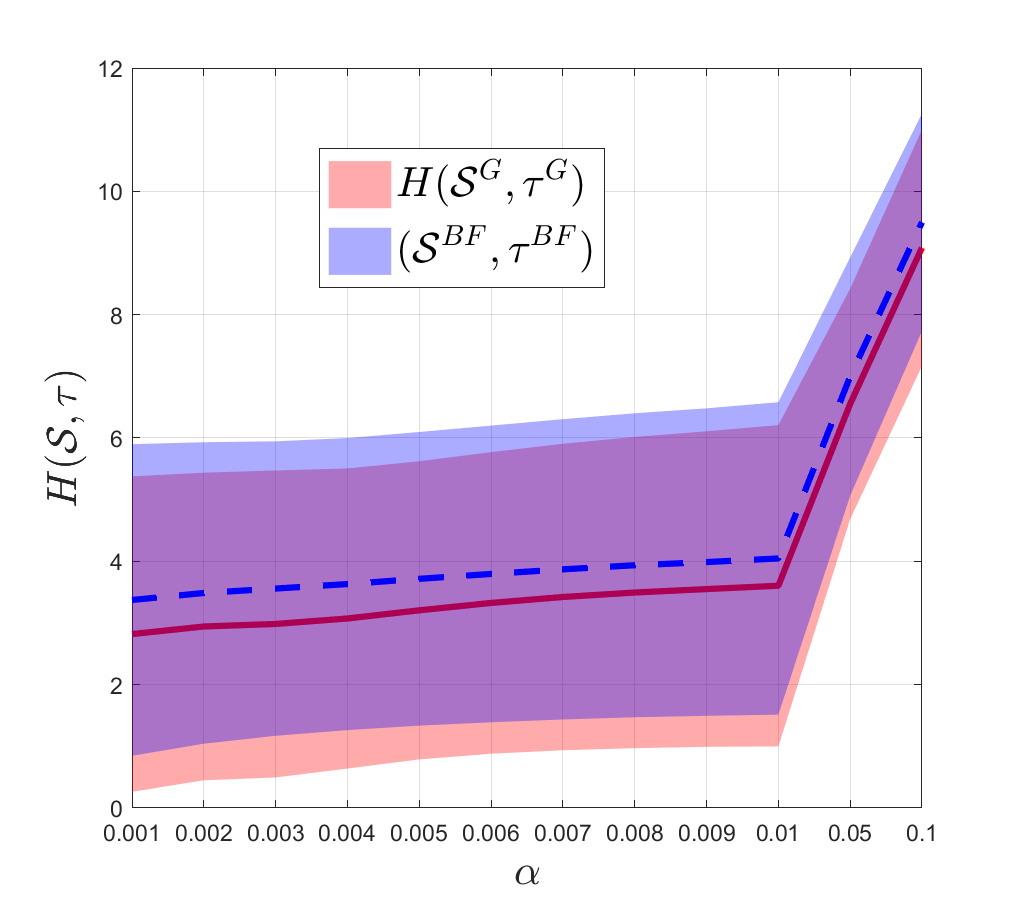}
  \caption{Comparison of the values of $H(\mathcal{S}, \tau)$ obtained by SGA (i.e., $H(\mathcal{S}^{G}, \tau^{G})$) and obtained by the sequential brute-force algorithm with respect to small risk levels $\alpha \in [0.001, 0.1]$ in 30 trials.}
  \label{fig:gre_bf_compare_sensor}
\endminipage
\end{figure*}

Based on the $\mathcal{S}^{G}$ calculated by SGA, we sample $n_s = 1000$ times from the underlying distribution of  $y$ and plot the distribution of the selection utility, $$f(\mathcal{S}^{G}, y)  = \bigcup_{i=1:M} A_i, i\in\mathcal{S}^{G}$$ in Figure~\ref{fig:dis_select_utility}. Note that, when the risk level $\alpha$ is small, the sensors with smaller visibility region and a higher probability of success should be selected. Lower risk level suggests a  conservative selection.  Sensors with a higher probability of success are selected to avoid the risk induced by the sensor failure. In contrast, when $\alpha$ is large, the selection would like to take more risk to gain more monitoring utility. The sensors with a larger visibility region and a lower probability of success should be selected.  Figure~\ref{fig:dis_select_utility} demonstrates this behavior except when  $\alpha = 0.001$. This is because when $\alpha$ is very small, the approximation error (Eq.~\ref{eqn:add_error}) is very large as shown in Figure~\ref{fig:add_err_vis}, and thus SGA may not give a good solution. 

Therefore, it is necessary to see how well SGA performs for small risk levels. Similar to the mobility-on-demand case, we compare SGA with a sequential brute-force algorithm that uses the brute-force search as a subroutine to enumerate all possible feasible placements for finding the optimal sensor placement $\mathcal{S}^{BF}$ for each specific $\tau_i$. For instance, in this environment monitoring scenario with selecting 4 placement locations from 8 candidate locations, there will be $\binom{8}{4} = 70$ placements in total. From these 70 possible placements, we choose the placement $\mathcal{S}_i^{BF}$ that maximizes $H(\mathcal{S}, \tau_i)$ at each $\tau_i$. Then we pick out the best pair $(\mathcal{S}^{BF}, \tau^{BF})$ that has the maximum value of $H(\mathcal{S}, \tau)$, denoted by $H(\mathcal{S}^{BF}, \tau^{BF})$, from all $(\mathcal{S}_i^{BF}, \tau_i)$ pairs (collected when searching for all possible values of $\tau$). Note that the sequential brute-force algorithm has no additive error for maximizing  $H(\mathcal{S}, \tau)$ (Eq.~\ref{eqn:SBFA}). 

With the same set of 8 candidate locations randomly generated in 30 trials, we compare SGA with the sequential brute-force algorithm for small risk levels $\alpha \in [0.001, 0.1]$ in Figure~\ref{fig:gre_bf_compare_sensor}. The result shows that on average SGA works comparatively with the sequential brute-force algorithm (especially when $\alpha \geq 0.01$) for maximizing $H(\mathcal{S}, \tau)$. This implies SGA can still work relatively well on average when the risk level is close to zero.

\begin{figure*}
\centering{
% \hspace*{-5.5mm}
\subfigure[Selection when $\alpha = 0.01$.]{\includegraphics[width=0.57\columnwidth]{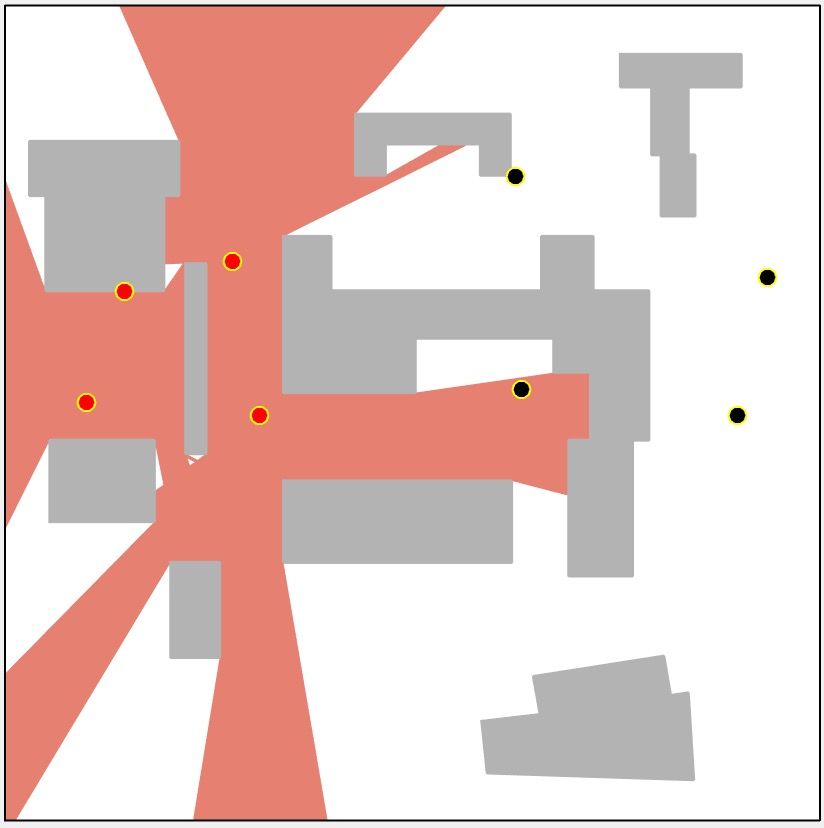}} ~~~~~~
\subfigure[Selection when $\alpha = 1$ ($\text{Risk-neutral}$). ]{\includegraphics[width=0.57\columnwidth]{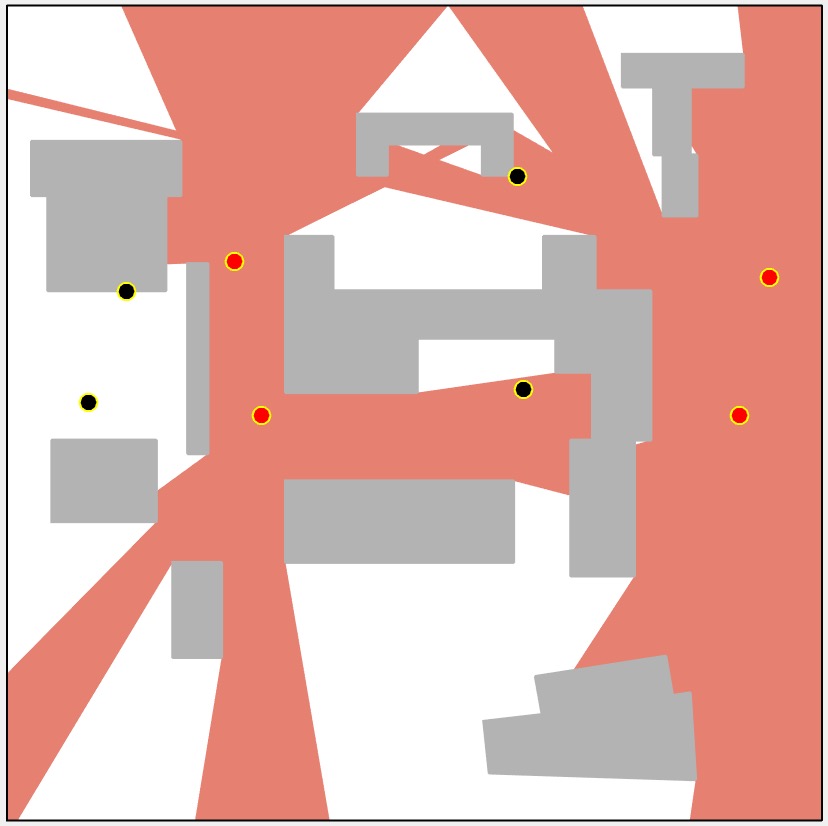}}~~~~~~
\subfigure[Selection utility distributions at $\alpha=0.01$ and $\alpha =1$.]{\includegraphics[width=0.65\columnwidth]{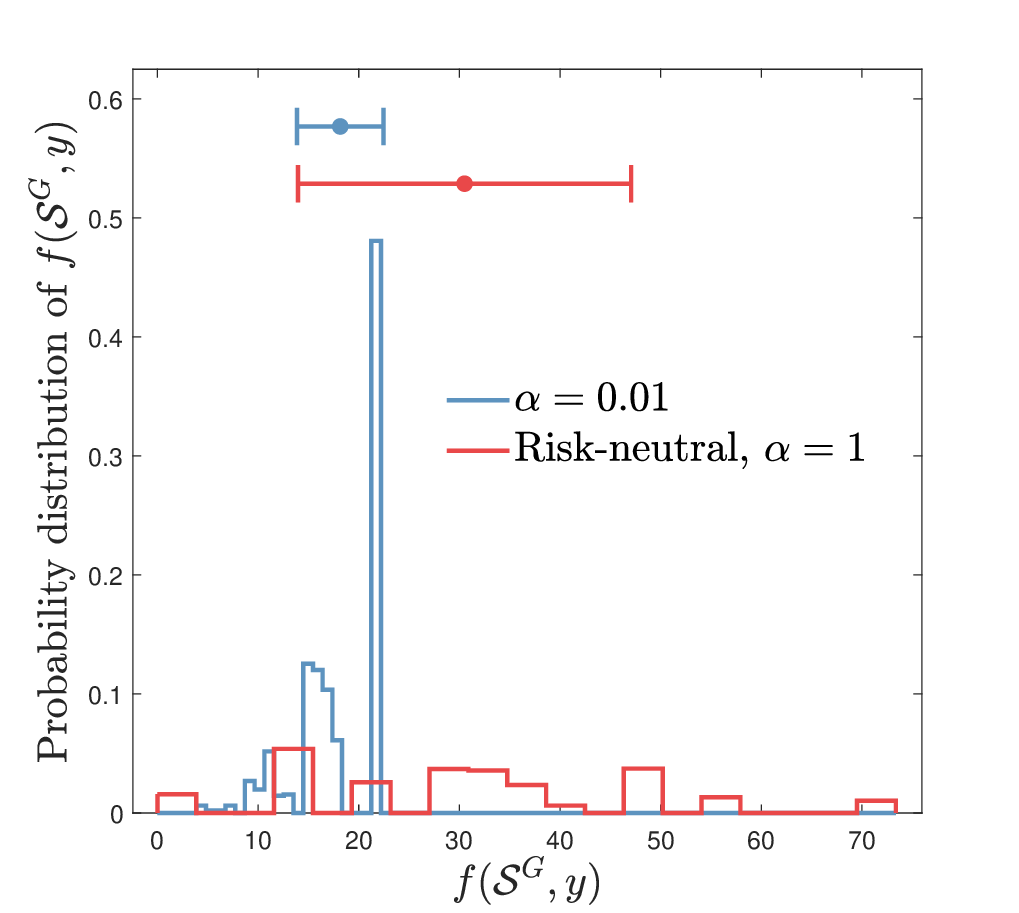}}
\caption{Sensor selection and utility distributions by  SGA with two extreme risk level values. The red solid circle represents the sensor selected by  SGA. 
\label{fig:two_exteme_visi}}}
\end{figure*}

We also compare SGA by using CVaR measure with the greedy algorithm by using the expectation, i.e., risk-neutral measure (mentioned in~\cite[Section 6.1]{krause2008near}) in Figure~\ref{fig:two_exteme_visi}. In fact, the risk-neutral measure is equivalent to the case of $\text{CVaR}_{\alpha}$ when $\alpha =1$. We give an illustrative example of the sensor selection by SGA for two extreme risk levels, $\alpha = 0.01$ and $\alpha = 1$. When risk level $\alpha$ is small ($\alpha = 0.01$), the selection is conservative and thus the sensors with small visibility region are selected (Fig.~\ref{fig:two_exteme_visi}-(a)). In contrast, when $\alpha =1$, the risk is neutral and the selection is more adventurous, and thus sensors with large visibility region are selected (Fig.~\ref{fig:two_exteme_visi}-(b)). The mean-std bars of the selection utility distributions in Figure~\ref{fig:two_exteme_visi}-(c) show that the selection utility at the expectation ($\alpha =1$) has larger mean value than the selection at $\alpha =0.01$. However, the selection at $\alpha =1$ has the risk of gaining lower utility since the left endpoint of mean-std bar at $\alpha =1$ is smaller than the left endpoint of mean-std bar at $\alpha =0.01$.

\subsection{Online Resilient Mobility-on-Demand}~\label{subsubsec:tri_assign_trig}
In this section, we show the effectiveness of ATA (Algs.~\ref{alg:online_tri_assign_gen},~\ref{alg:online_tri_assign}) by using the online mobility-on-demand with street networks. A detailed description of the settings is provided in the appendix. We compare the performance of ATA with the other three assignment strategies, \textit{offline}, \textit{one-step}, and \textit{all-step} assignments, in terms of the arrival time, the number of assignments, and the running time. The offline assignment is a clairvoyant strategy that knows the vehicles' arrival times (i.e., knows how the stochastic scenario will play out) exactly and uses the standard greedy algorithm~\cite{fisher1978analysis} to compute the assignment.\footnote{In the simulation, this is done by first storing the realistic arrival times of the vehicles at the final step and then using the greedy algorithm~\cite{fisher1978analysis} to compute an offline assignment.} Thus, it can be set as a baseline. The difference of the three online assignment strategies, ATA, \textit{one-step assignment}, and \textit{all-step assignment} comes from how often the assignment is scheduled--- In the \textit{one-step} assignment, the assignment is calculated at the initial time step and is fixed thereafter; In the \textit{all-step} assignment, the assignment is computed at each time step; In ATA, the assignment is triggered only at specific time steps.  In both \textit{one-step} and \textit{all-step} assignments, all vehicles apply the ``per-step online travel'' rule (Alg.~\ref{alg:online_tri_assign}, lines~\ref{line:for_online_travel}-\ref{line:for_online_end}) to go towards the assigned demand locations, as in the case of ATA.  Particularly, in these three online assignment strategies, we compute the arrival time as the summation of all per-step time intervals ($T^{\texttt{step}}$ in Eq.~\ref{eqn:one_step}) from time of the first step to the final step when all demand locations are reached.
% A video showing ATA in action is available online.\footnote{\url{}}

\begin{figure}
  \centering
  \includegraphics[width=0.52\columnwidth]{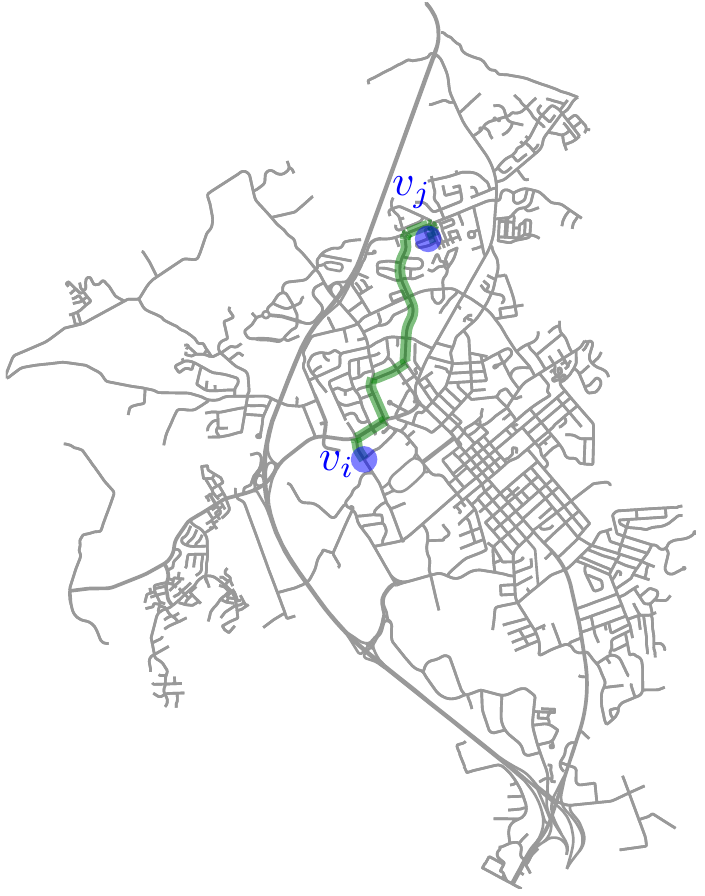}
  \caption{Street network created by OSMnx in Blacksburg, VA, USA with a shortest path (green path) from node $v_i$ to node $v_j$ (lighter blue dots).}
  \label{fig:bb_short}
\end{figure}

\begin{figure}
  \centering
  \includegraphics[width=0.65\columnwidth]{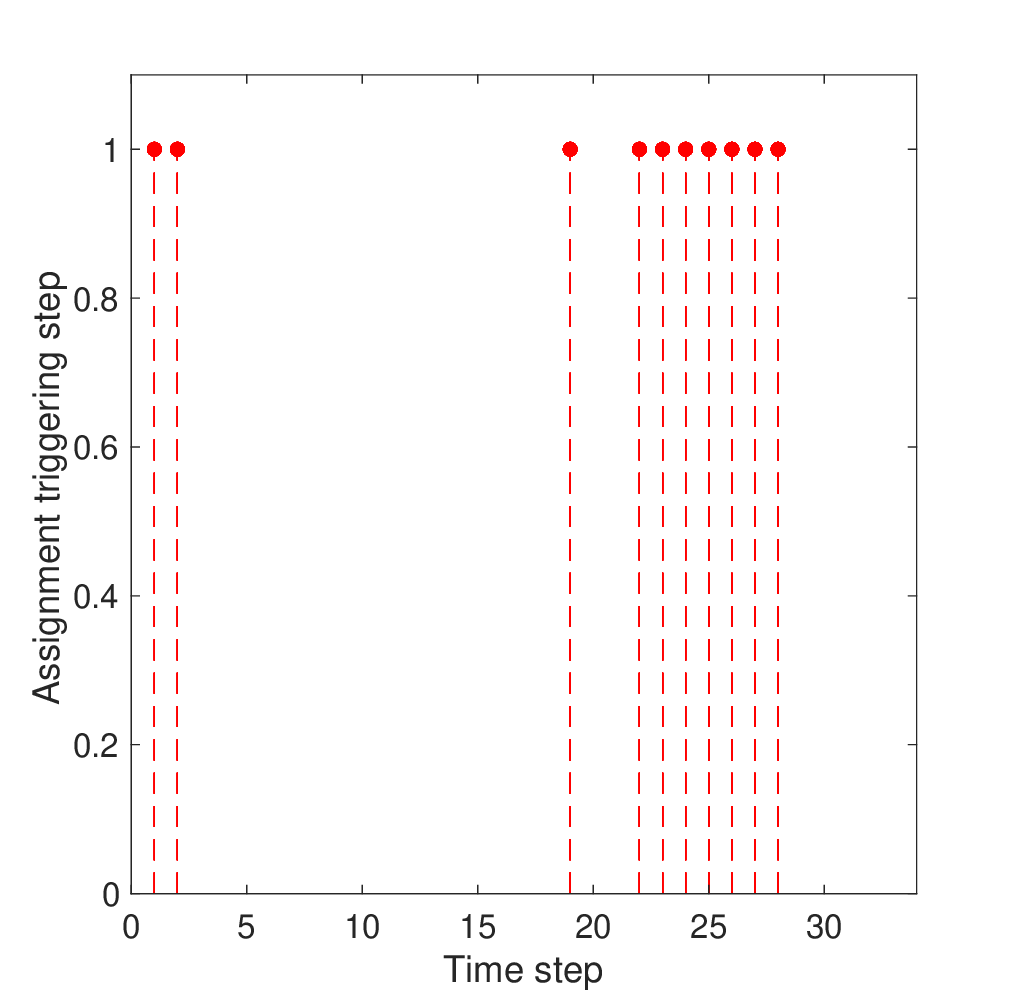}
  \caption{The triggering time steps in one trial of ATA (Alg.~\ref{alg:online_tri_assign}) in action. }
  \label{fig:trigger_inx}
\end{figure}

%%%%%%%%%%%%%%%%%%%%%%%%%%%%%%%%%%
\begin{figure}
\centering{
\subfigure[6-vehicle 4-demand system]
{\includegraphics[width=0.49\columnwidth]{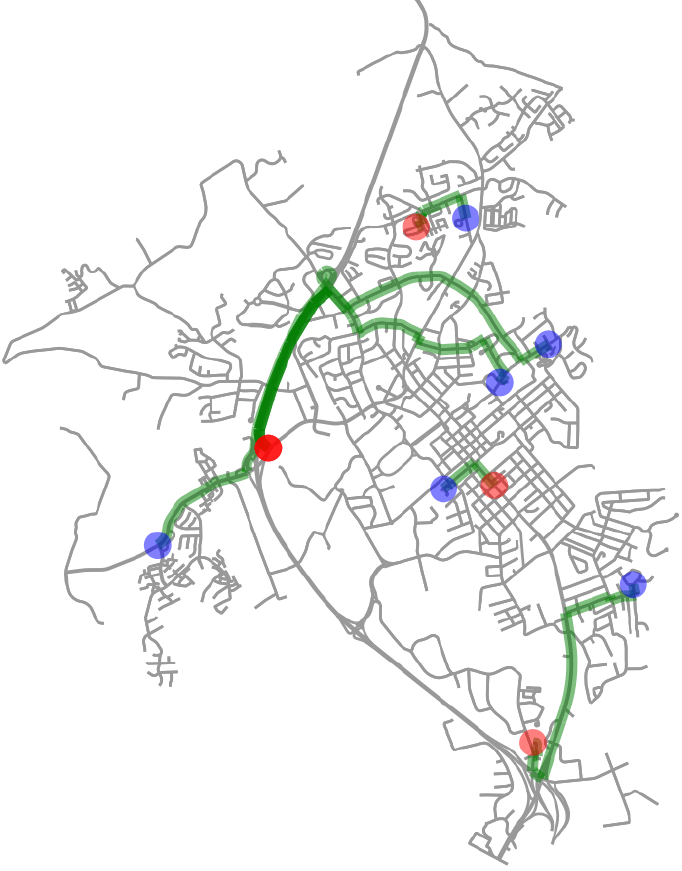}}~
\subfigure[12-vehicle 5-demand system]
{\includegraphics[width=0.49\columnwidth]{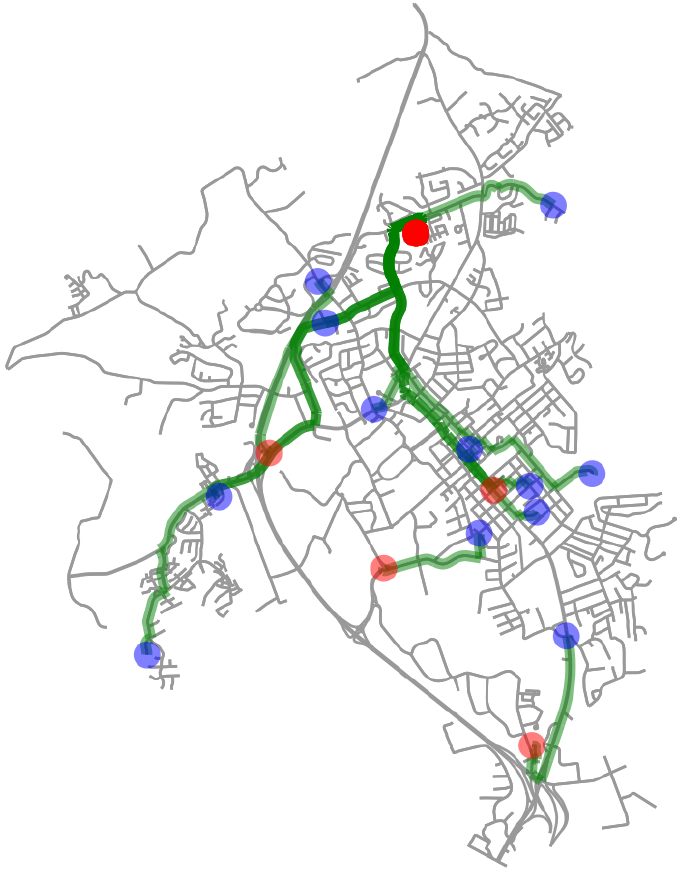}}
\caption{The initial assignment in (a), 6-vehicle 4-demand system and (b), 12-vehicle 5-demand system. 
\label{fig:other_two_scale}}
}
\end{figure}
%%%%%%%%%%%%%%%%%%%%%%%%%%%%%%%%%%

We consider assigning $R$ supply vehicles to $N$ demand locations on a street network, $\mathcal{G}$, created using OSMnx~\cite{boeing2017osmnx}. OSMnx is a python package that allows for easily constructing, projecting, visualizing, and analyzing complex street networks~\cite{boeing2017osmnx}. We give an example showing a network of drivable public streets created by OSMnx in Blacksburg, VA, USA in Figure~\ref{fig:bb_short}. OSMnx also allows for finding a shortest path between two nodes (intersections) of the networks by Dijkstra's algorithm if there exists one (Fig.~\ref{fig:bb_short}). Thus, we use OSMnx to compute a shortest path from each vehicle's initial position to each demand location, if there exists one. We can also extract the intersection degree and edge length of the street network directly by using OSMnx. 

We model the waiting time at intersection $v_i$ following a truncated normal distribution, 
\begin{equation}
    t(v_i) \sim \mathcal{N}^{\texttt{Truc}}(0, \beta_1 \texttt{deg}(v_i)), ~~t(v_i) \in [0, T(v_i)] %$\overline{T}(v_i)$
    \label{eqn:node_wait_time}
\end{equation} where $T(v_i)$ denotes the maximum waiting time at the intersection $v_i$. We set the variance of the waiting time to be proportional to the degree of the intersection, i.e., $\texttt{var}(t(v_i)) = \beta_1 \texttt{deg}(v_i)$ with $\beta_1 \in (0, +\infty)$. A larger degree means more incoming and outgoing edges, and thus the traffic at this intersection is more likely to be congested, which increases the probability of higher waiting time. Notably, the truncated Gaussian distribution is just one possible way to capture the randomness of the waiting time. In practice, instead of assuming a particular distribution, one would use historic traffic data to model the distribution at each intersection and along every edge. Our algorithm works for any model that can be sampled. 

% To compute the truncated Gaussian distribution of the waiting time at each intersection $v_i$ (Eq.~\ref{eqn:node_wait_time}), 
We set $\beta_1=1$ and model the maximum waiting time as $T(v_i)=5\texttt{deg}(v_i)$. The model assumes that a larger degree of an intersection implies a larger maximum waiting time there. We model the real-time waiting time at an intersection $v_i$ for each vehicle $i$ as a sample from the truncated normal distribution (Eq.~\ref{eqn:node_wait_time}). In practice, the real-time waiting time can be acquired when the vehicle reaches the intersection.

We set $\beta_2 = 1$ and use the speed limits of the street network in Blacksburg from Google Maps to compute the edge travel time (see Eq.~\ref{eqn:edge_travel_time}). If a demand location is not reachable from a vehicle's position, we set the path travel time (Eq.~\ref{eqn:path_travel_time}) from the vehicle to the demand as infinity.
We set the risk level $\alpha = 0.1$ in SGA (Alg.~\ref{alg:sga}). 

\subsubsection{Qualitative Results} 
\begin{figure*}
\centering{
% \hspace*{-5.5mm}
\subfigure[Step 1]{\includegraphics[width=0.472\columnwidth]{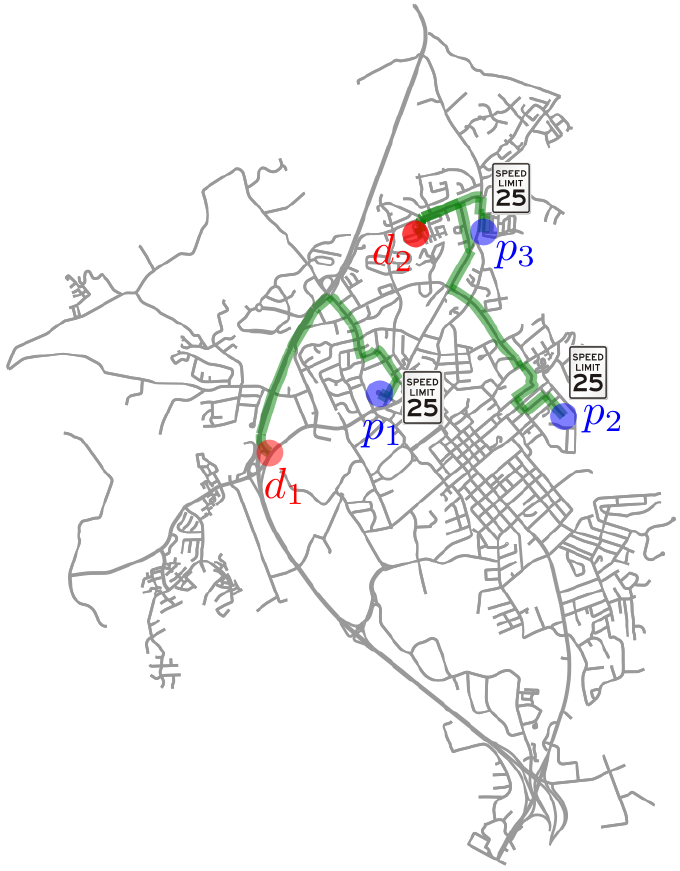}}
\subfigure[Step 2 ]{\includegraphics[width=0.472\columnwidth]{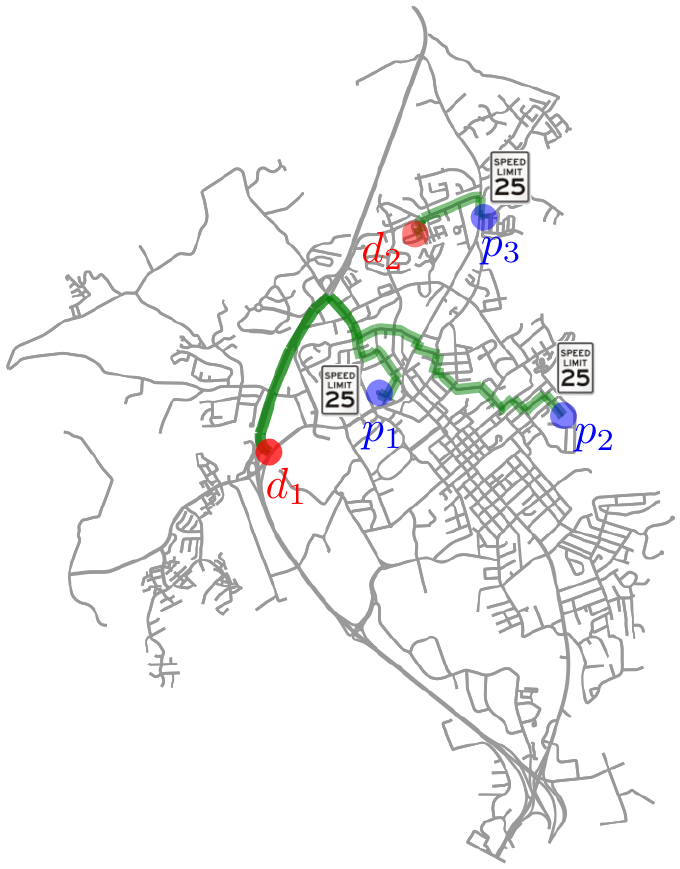}}
\subfigure[Step 19]{\includegraphics[width=0.472\columnwidth]{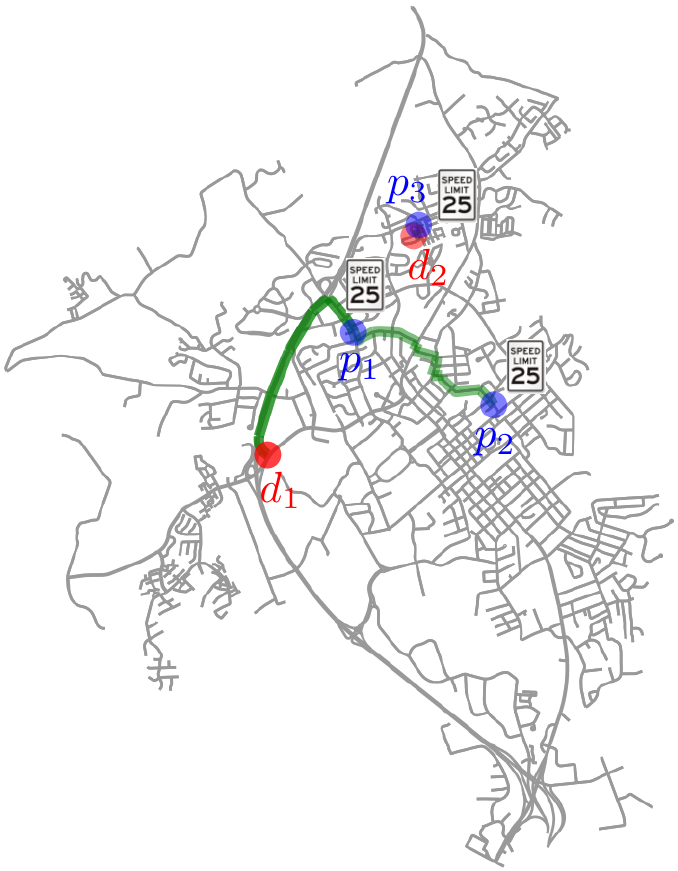}}
\subfigure[Step 20]{\includegraphics[width=0.472\columnwidth]{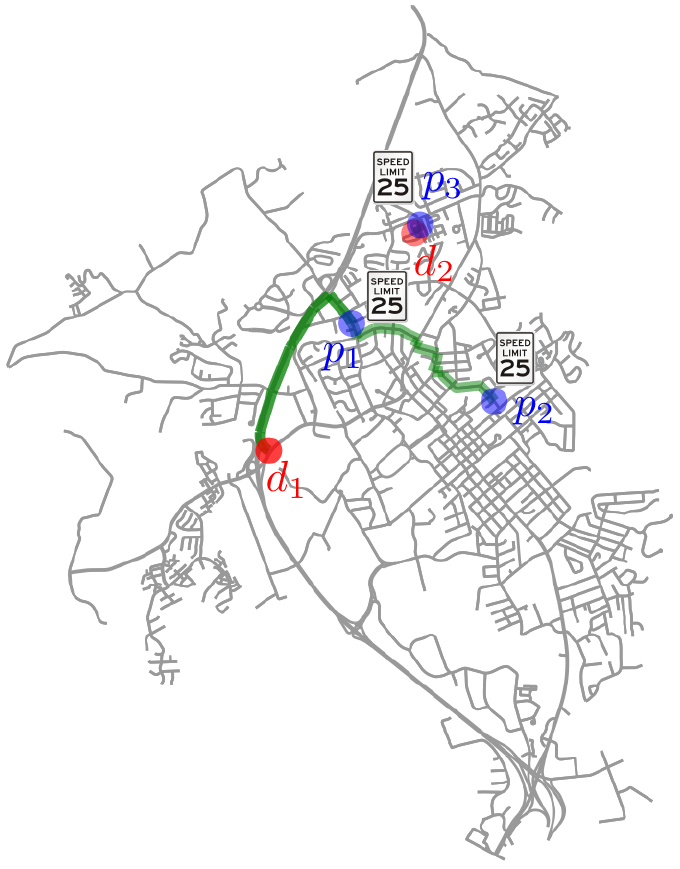}}
\subfigure[Step 22]{\includegraphics[width=0.472\columnwidth]{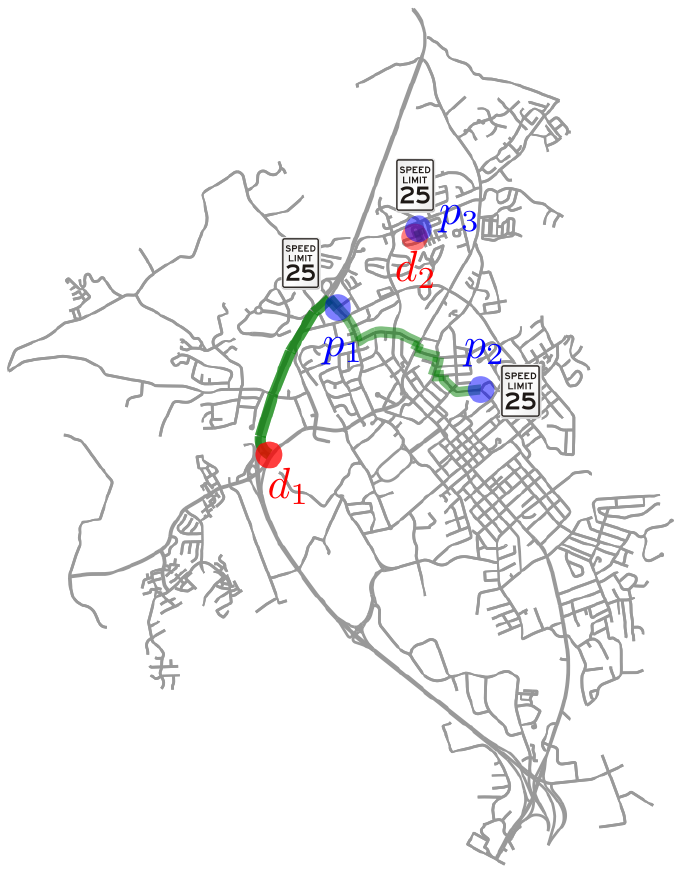}}
\subfigure[Step 23 ]{\includegraphics[width=0.472\columnwidth]{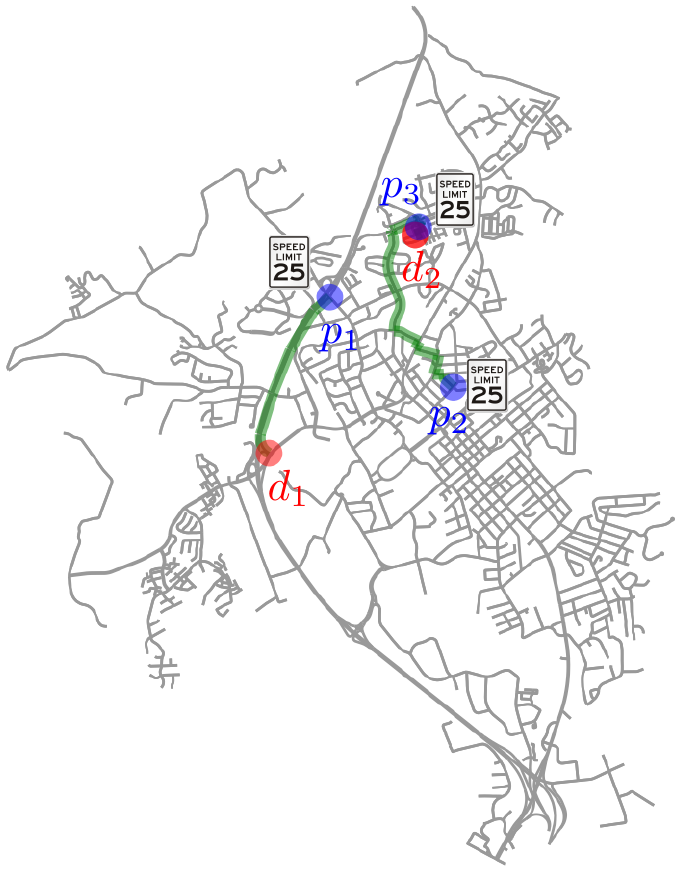}}
\subfigure[Step 28]{\includegraphics[width=0.472\columnwidth]{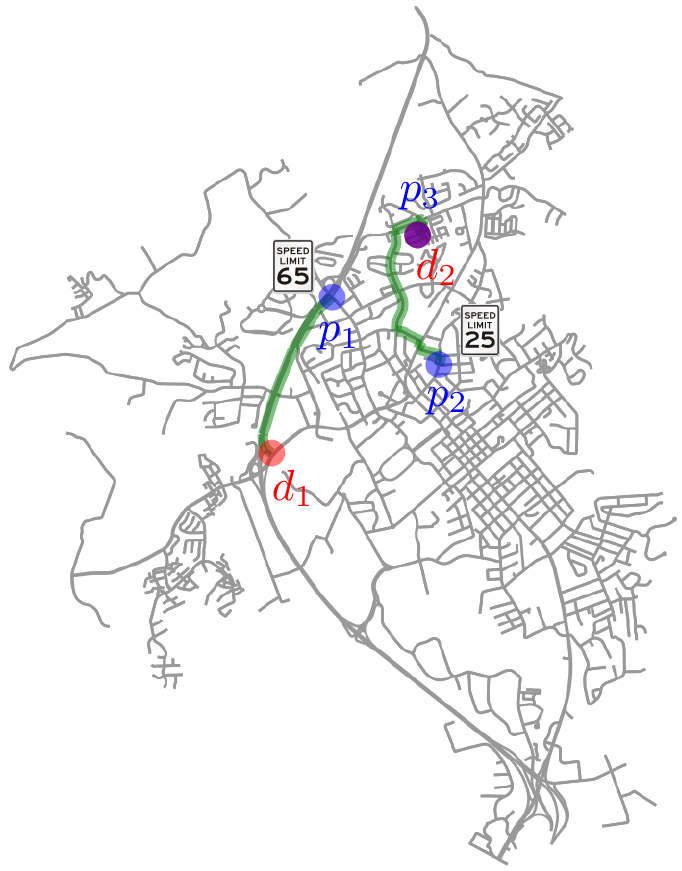}}
\subfigure[Step 29]{\includegraphics[width=0.472\columnwidth]{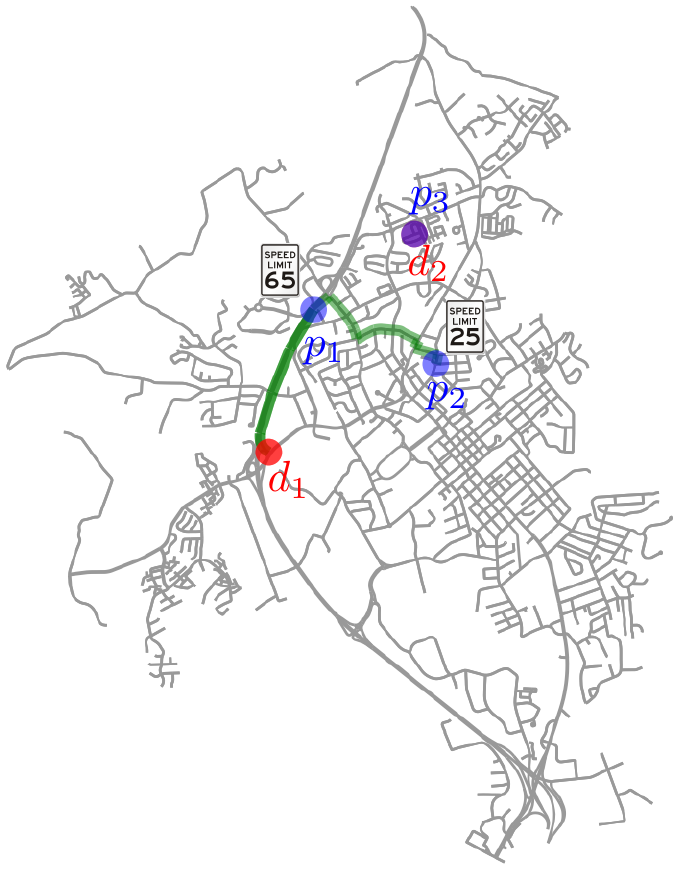}}
\subfigure[Step 30]{\includegraphics[width=0.472\columnwidth]{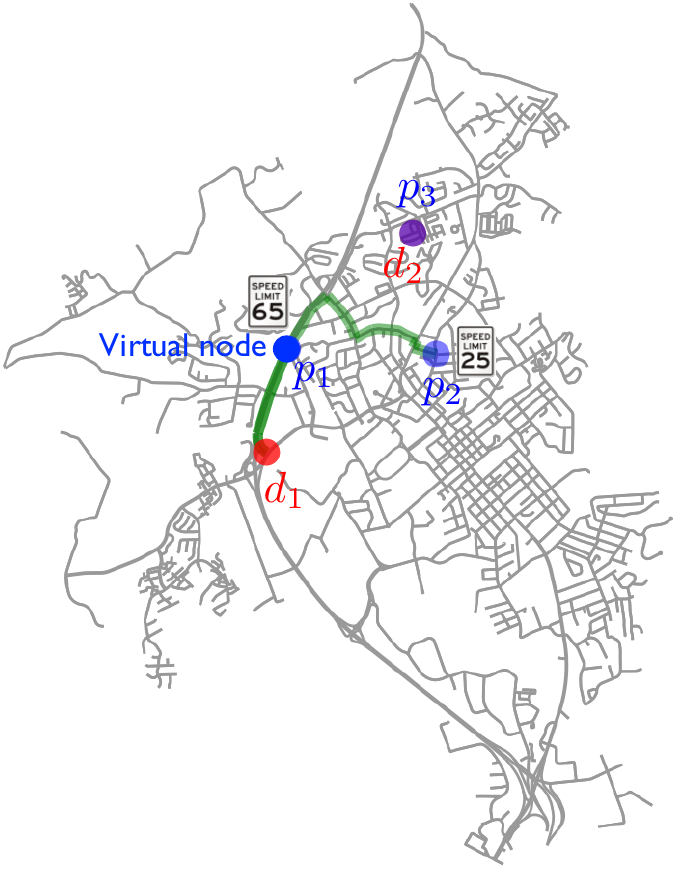}}
\subfigure[Step 31 ]{\includegraphics[width=0.472\columnwidth]{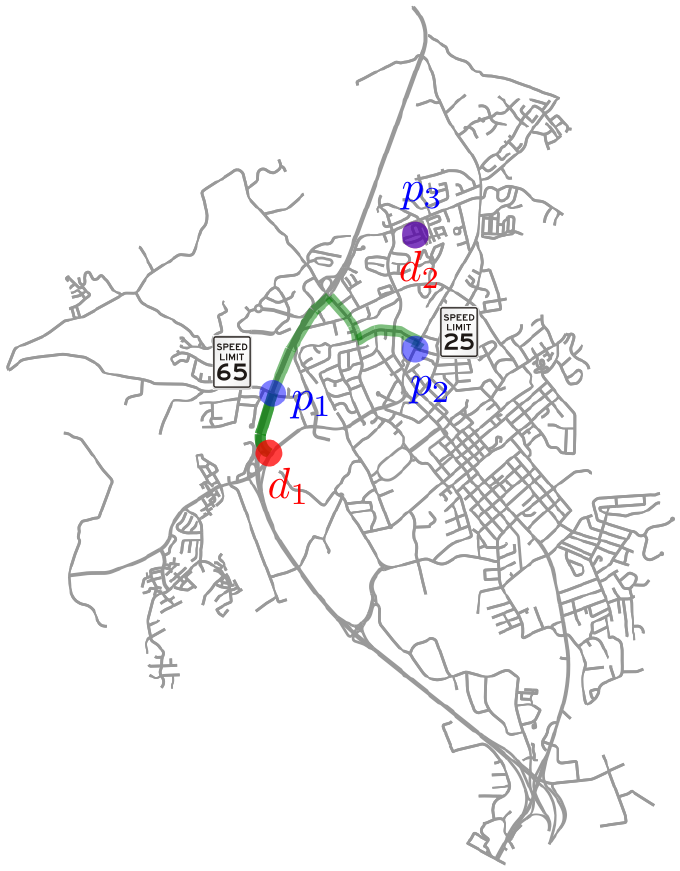}}
\subfigure[Step 32]{\includegraphics[width=0.472\columnwidth]{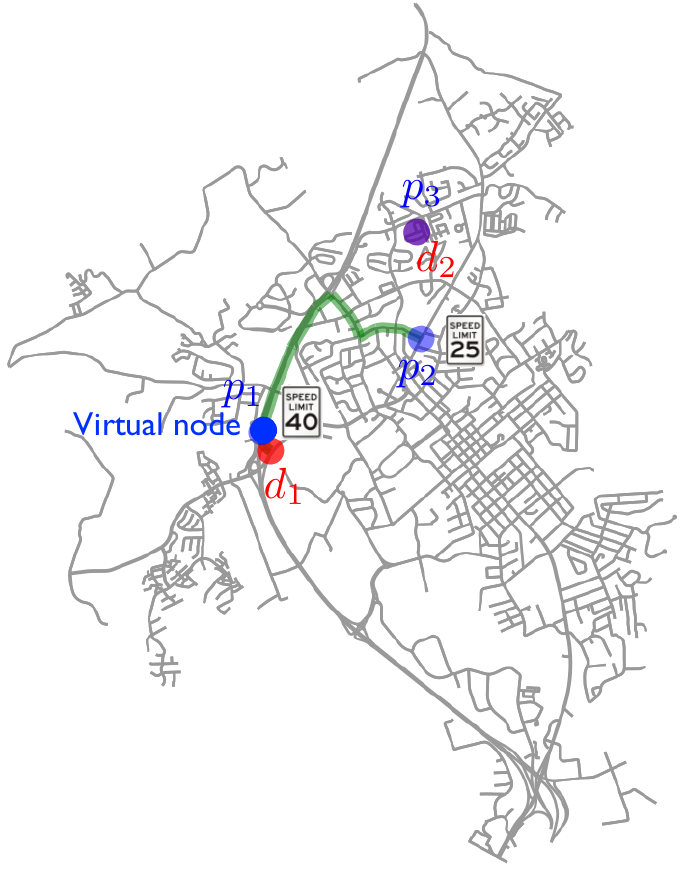}}
\subfigure[Step 34]{\includegraphics[width=0.472\columnwidth]{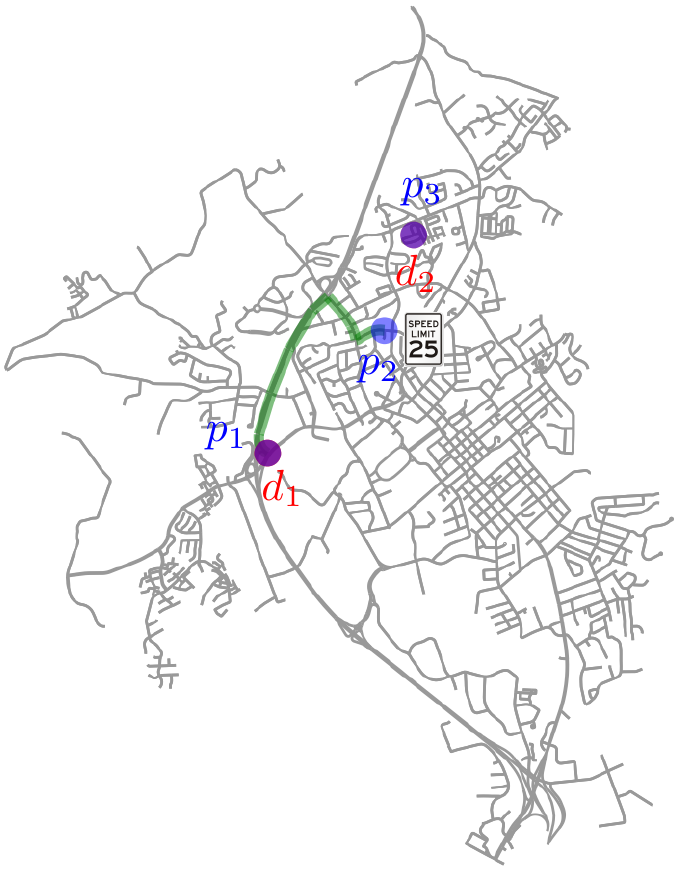}}
\caption{One trial of ATA (Alg.~\ref{alg:online_tri_assign}) in action with 3 vehicles and 2 demand locations.  The blue dots show the positions of the vehicles ($p_1, p_2, p_3$). The red dots show the demand locations ($d_1, d_2$). The speed sign shows the maximum speed on the street.
\label{fig:assign_in_action}}}
\end{figure*}

We first show one trial of ATA in action in Figure~\ref{fig:assign_in_action} where we assign $R=3$ vehicles to $N=2$ demand locations with triggering ratio $\gamma = 0.5$. Since we model the real-time waiting time at each intersection as a random sample from its distribution, ATA can end up with different results in different trials, even with the same vehicles' initial positions and demand locations. 

In Figure~\ref{fig:assign_in_action}, ATA starts at time step 1 (Fig.~\ref{fig:assign_in_action}-(a)) and ends at time step 34 (Fig.~\ref{fig:assign_in_action}-(l)) when the two demand locations, $d_1$ and $d_2$, are reached. Within these 34 time steps, the assignment is only triggered at 10 time steps (steps 1, 2, 19, 22, 23, 24, 25, 26, 27, 28) as shown in Figure~\ref{fig:trigger_inx}. At these 10 steps, the triggering conditions with $\gamma =0.5$ (Eqs.~\ref{eqn:tri_1} and \ref{eqn:tri_2}) are satisfied. For example, at step 1 (Fig.~\ref{fig:assign_in_action}-(a)), the path length from $p_3$ to $d_2$ is less than half of that from $p_2$ to $d_2$ (Eq.~\ref{eqn:tri_1}) and the path degree from $p_3$ to $d_2$ is less than that from $p_2$ to $d_2$ (Eq.~\ref{eqn:tri_2}). In this case, there is no need to continue assigning vehicle 2 to $d_2$ since it is likely that vehicle 3 will reach $d_2$ earlier than vehicle 2. Therefore, triggering the reassignment to assign vehicle 2 to $d_1$ could be helpful in reducing the arrival time at $d_1$ (step 2, Fig.~\ref{fig:assign_in_action}-(b)). Similar assignment flip happens from step 22 (Fig.~\ref{fig:assign_in_action}-(e)) to step 23 (Fig.~\ref{fig:assign_in_action}-(f)) where vehicle 2 is switched to be assigned to $d_2$. However, the assignment does not flip at all triggering time steps. For example, even though the assignment is triggered at step 19 (Fig.~\ref{fig:trigger_inx}), the assignment remains the same at step 20 (Fig.~\ref{fig:assign_in_action}-(c) \& (d)). One possible reason could be that assigning vehicle $2$ to $d_1$ is still more helpful, since vehicle $3$ is already very close to $d_2$ (Fig.~\ref{fig:assign_in_action}-(c) \& (d)). ATA also guarantees that once a demand is reached, e.g., in step 28, $d_2$ is reached by vehicle 3 (Fig.~\ref{fig:assign_in_action}-(g)), no vehicle will be assigned to it later (Fig.~\ref{fig:assign_in_action}-(h), (i), (j), (k), (l)). Particularly, in this 3-vehicle 2-demand system, there is not even reassignment triggered afterwards (Fig.~\ref{fig:trigger_inx}). 
In Figure~\ref{fig:assign_in_action}-(i) \& (k), we create virtual nodes (darker blue dots) to denote the intermediate positions of the vehicle on the highway. 

% \PRT{I think its better to reorganize this with some $\backslash$paragraph\{\} for qualitative and quantitative results}

\subsubsection{Quantitative Results}

% %%%%%%%%%%%%%%%%%%%%%%%%%%%%%%%%%%
% \begin{figure}[h]
% \centering{
% \subfigure[6-vehicle 4-demand system]
% {\includegraphics[width=0.5\columnwidth]{figs/6v4d_step1.pdf}}~
% \subfigure[12-vehicle 5-demand system]
% {\includegraphics[width=0.5\columnwidth]{figs/12v5d_step1.pdf}}
% \caption{The initial assignment in (a), 6-vehicle 4-demand system and (b), 12-vehicle 5-demand system. 
% \label{fig:other_two_scale}}
% }
% \end{figure}
% %%%%%%%%%%%%%%%%%%%%%%%%%%%%%%%%%%

\begin{figure*}
    \centering
    \begin{tabular}{c|c|c}
	\centering	
	\hspace*{-0.6mm}
	\subfigure[Arrival time, 3 vehicles \& 2 demands]{\includegraphics[width=0.619\columnwidth]{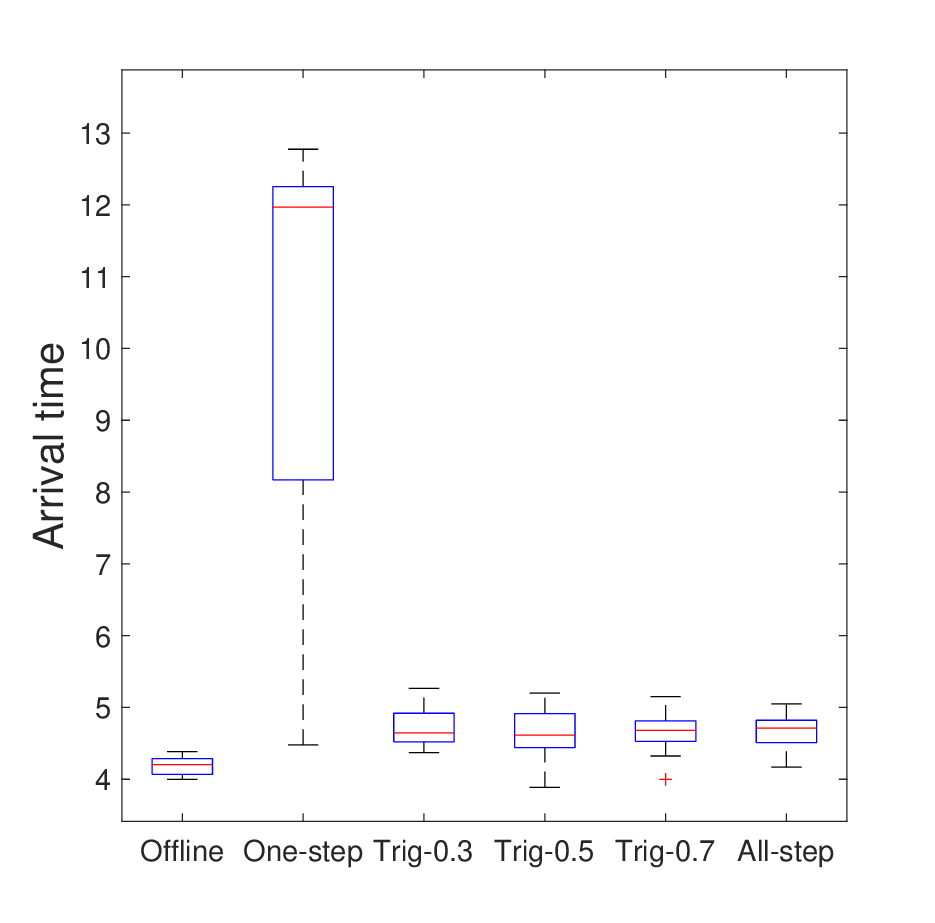}}
	&
	\subfigure[Arrival time, 6 vehicles \& 4 demands]{\includegraphics[width=0.619\columnwidth]{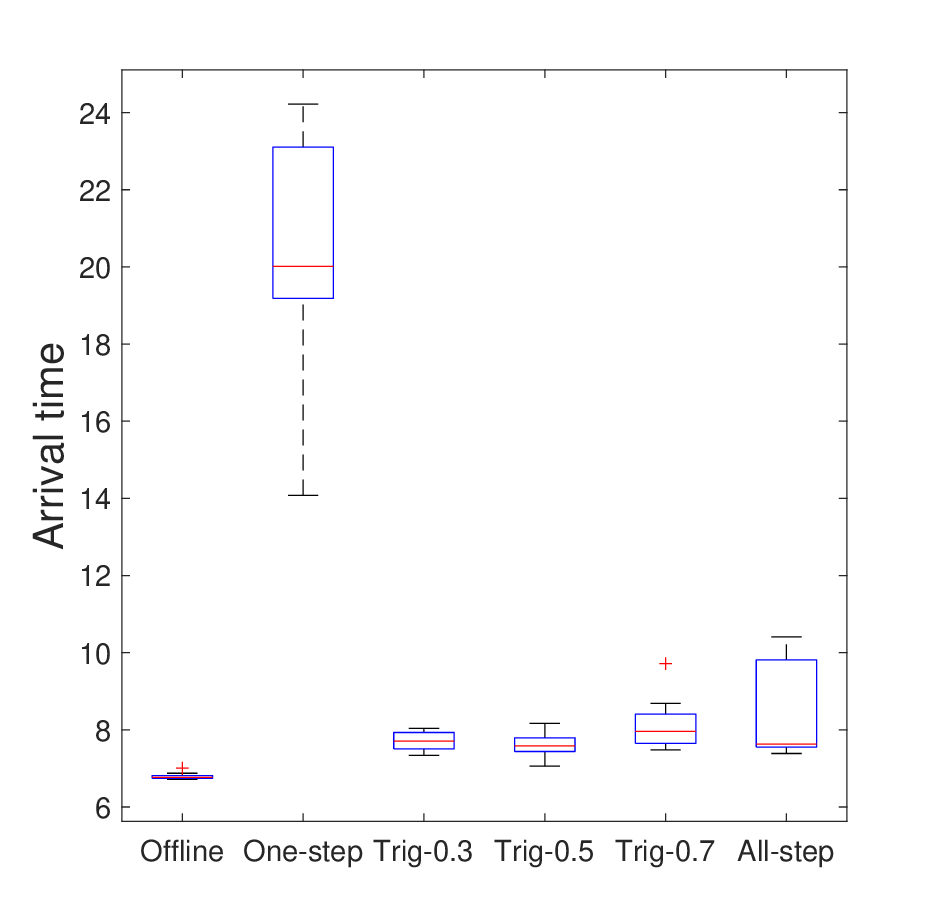}}
	&		
	\subfigure[Arrival time, 12 vehicles \& 5 demands] {\includegraphics[width=0.619\columnwidth]{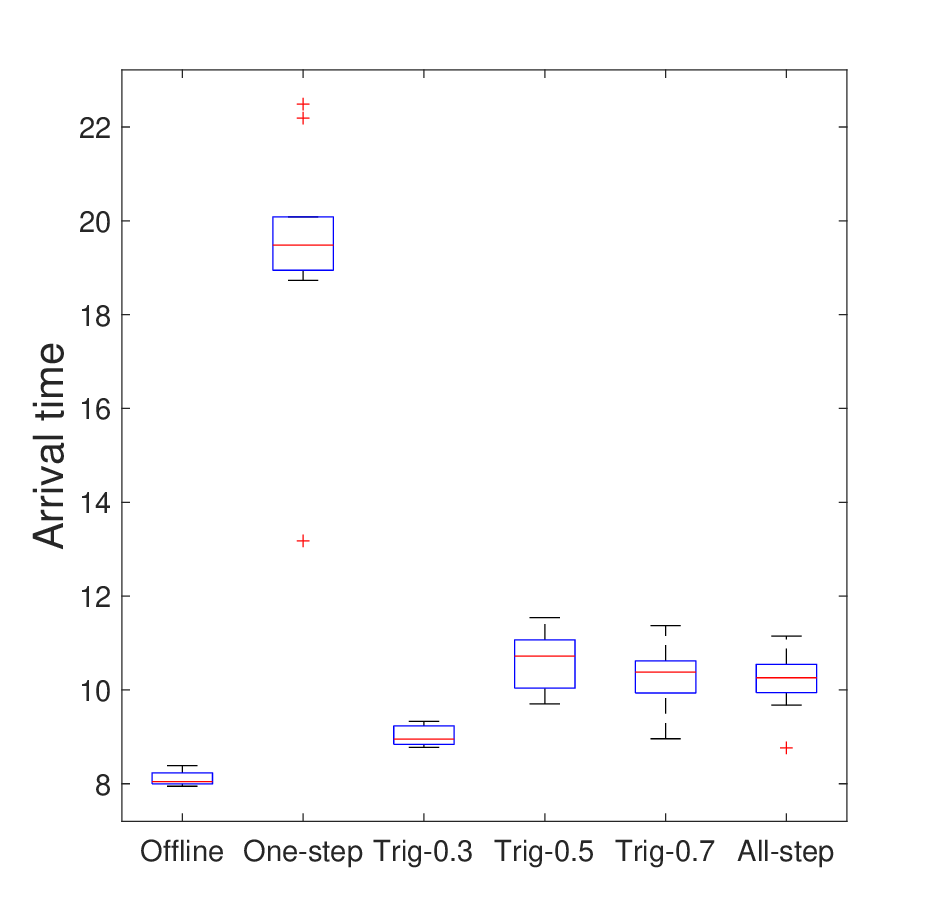}}
	\\
	\hspace*{-0.6mm}
	\subfigure[Number of assignments, 3 vehicles \& 2 demands]{\includegraphics[width=0.619\columnwidth]{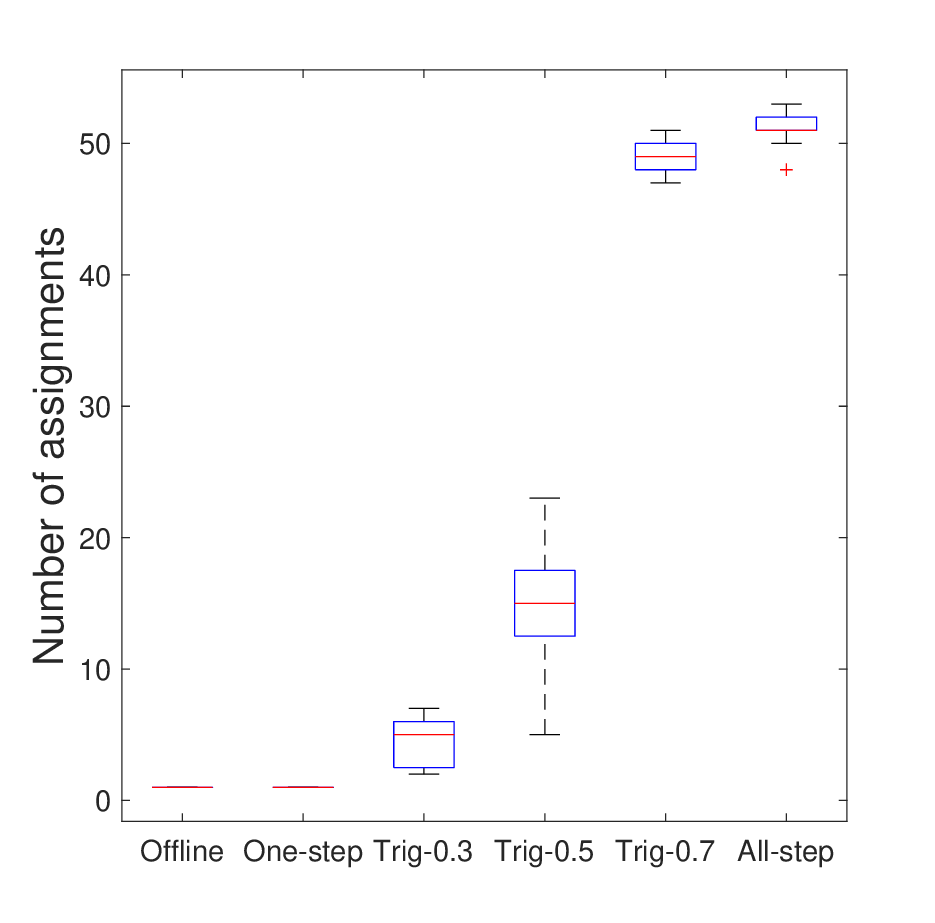}}
	&
	\subfigure[Number of assignments, 6 vehicles \& 4 demands]{\includegraphics[width=0.619\columnwidth]{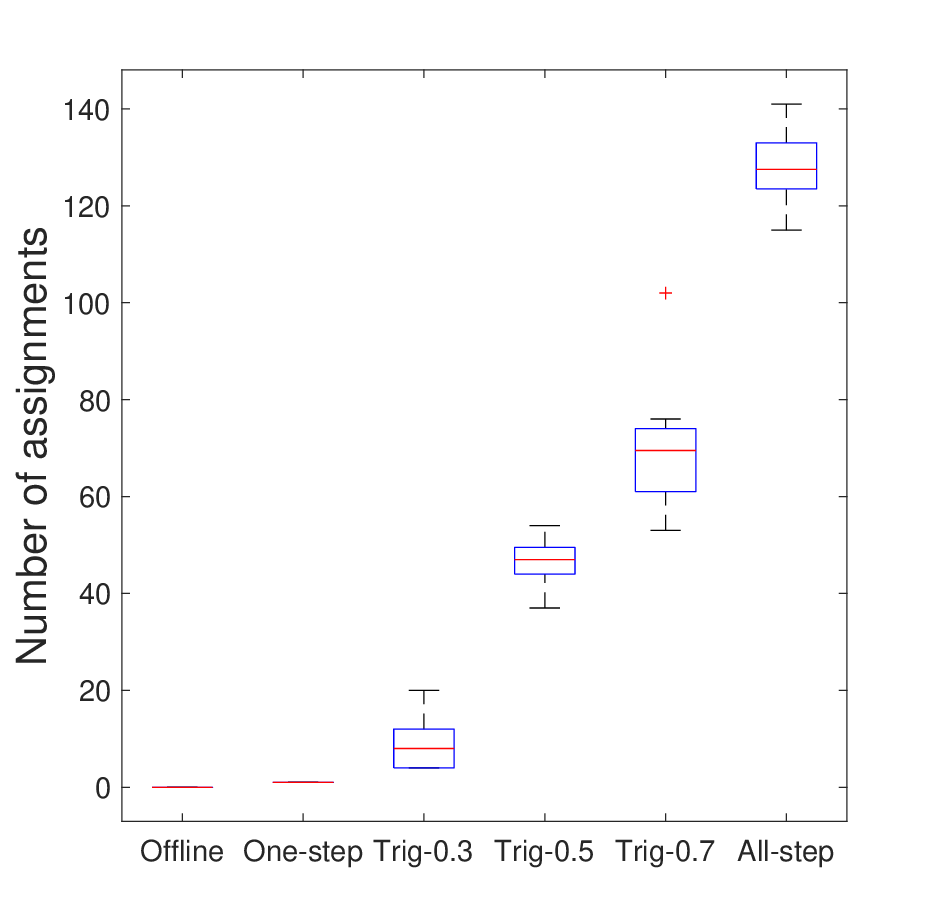}}
	&		
	\subfigure[Number of assignments, 12 vehicles \& 5 demands]{\includegraphics[width=0.619\columnwidth]{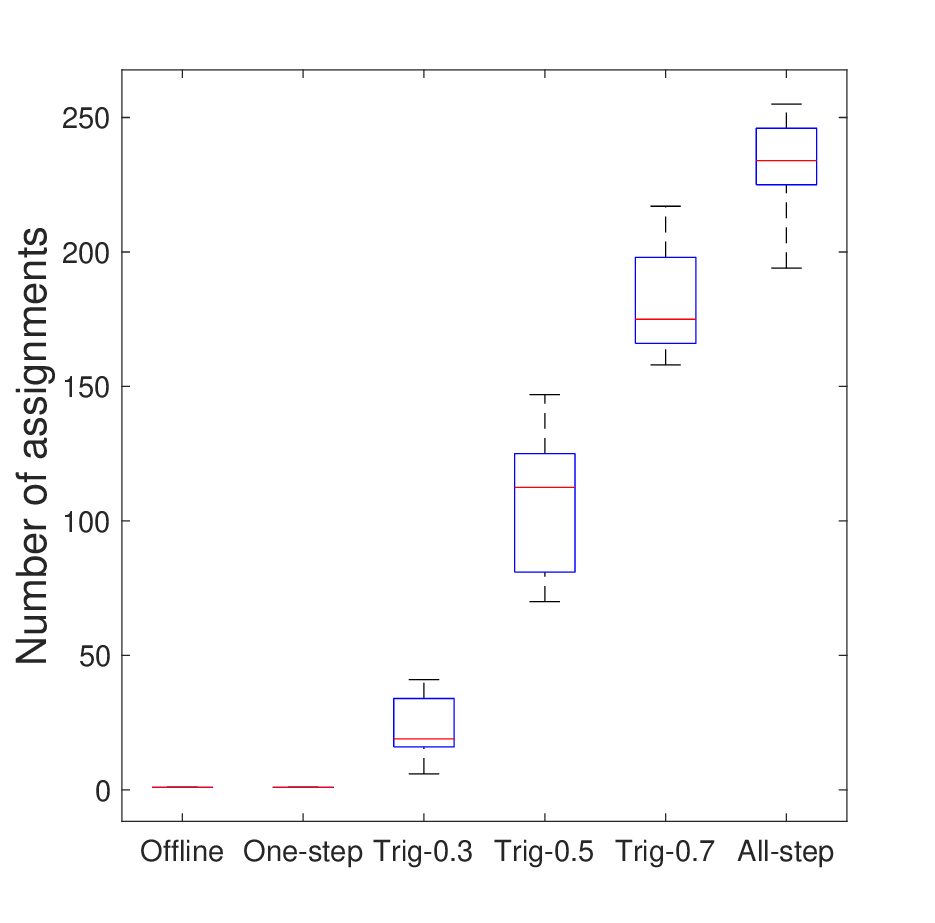}}
    \\
	\hspace*{-0.6mm}
	\subfigure[Running time, 3 vehicles \& 2 demands]{\includegraphics[width=0.619\columnwidth]{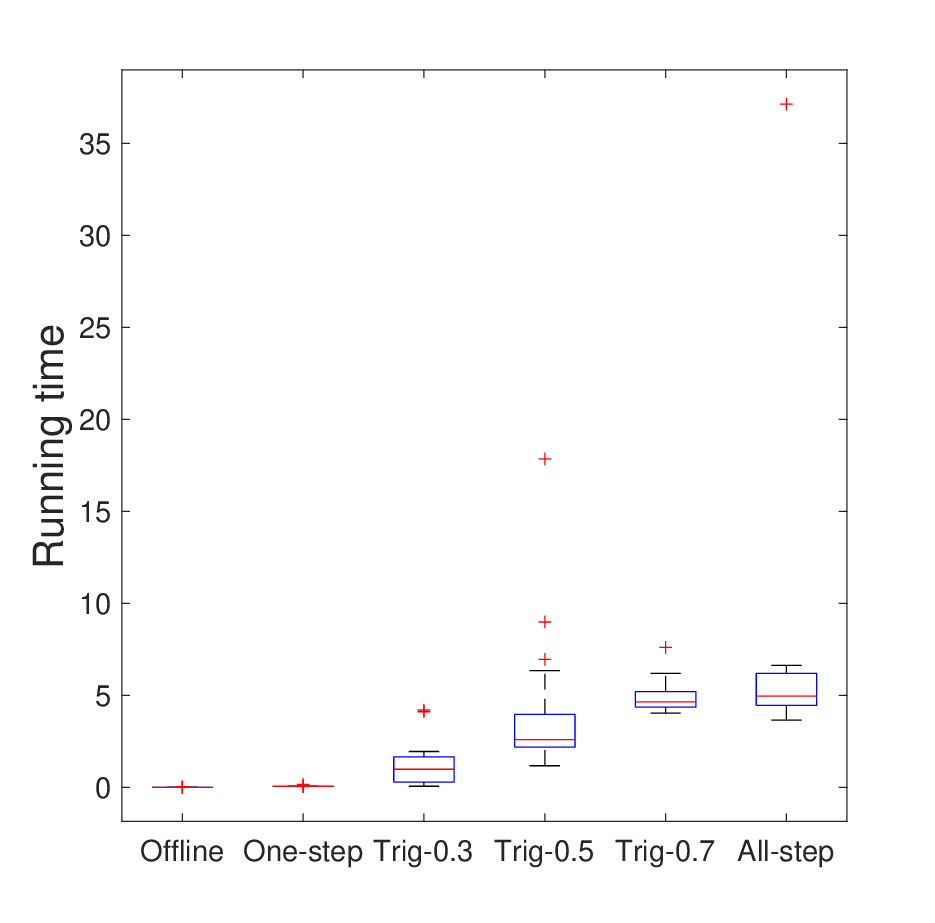}}
	&
	\subfigure[Running time, 6 vehicles \& 4 demands]{\includegraphics[width=0.619\columnwidth]{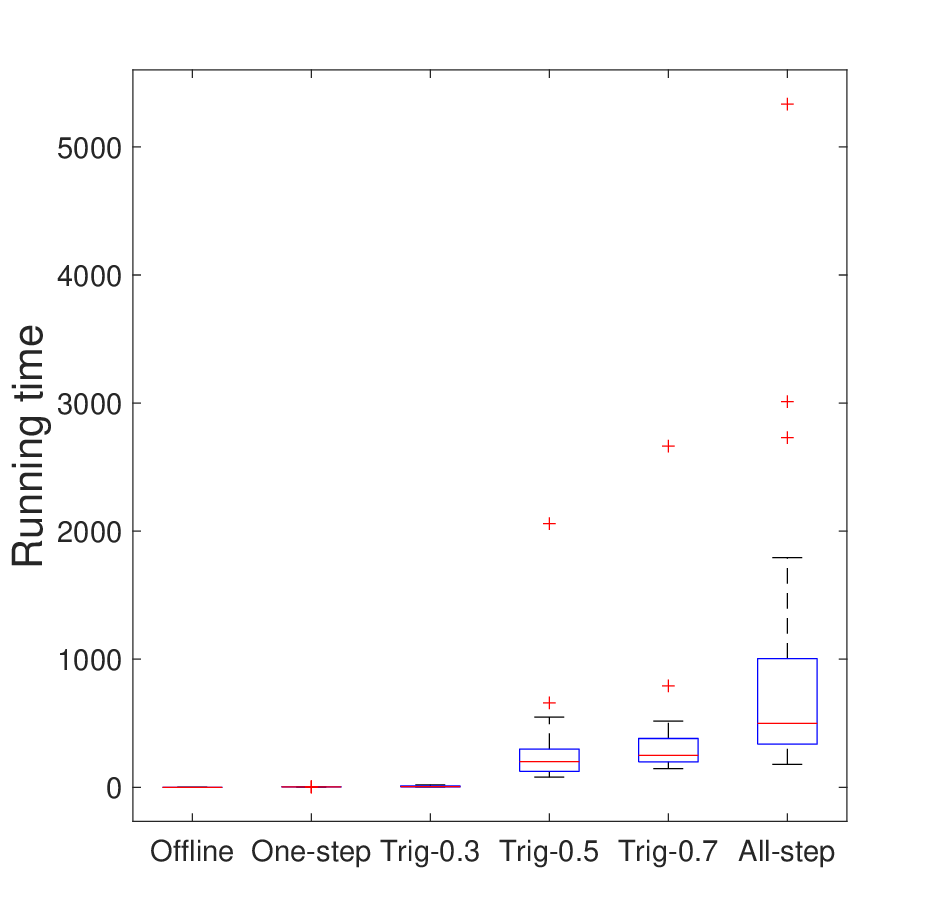}}
	&		
	\subfigure[Running time, 12 vehicles \& 5 demands]{\includegraphics[width=0.619\columnwidth]{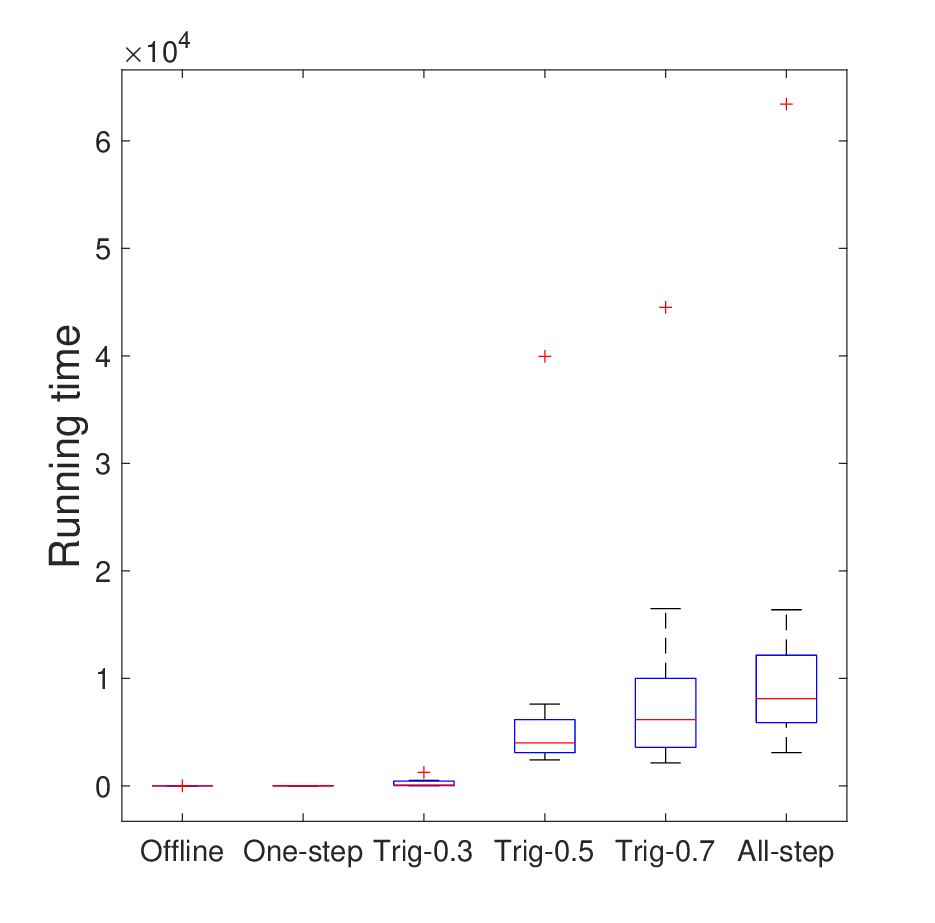}}
\end{tabular}
\caption{Comparison of the number of assignments, arrival time, and running time by the \textit{offline} assignment, \textit{one-step} assignment,  ATA (Alg.~\ref{alg:online_tri_assign}) with three triggering ratios, $\gamma= 0.3, 0.5, 0.7$, and \textit{all-step} assignment.}
\label{fig:compare_one_tri_per}
\end{figure*}

We then compare the arrival time, number of assignments, and running time of these three assignment strategies in Figure~\ref{fig:compare_one_tri_per}. We execute all three assignment strategies in 10 trials with the same initial vehicles' positions and demand locations. In particular, we consider three scales of vehicle-demand system, e.g., $R=3, N=2$ (Fig.~\ref{fig:assign_in_action}), $R=6, N=4$ (Fig.~\ref{fig:other_two_scale}-(a)) and $R=12, N=5$ (Fig.~\ref{fig:other_two_scale}-(b)). For each scale of the vehicle-demand system, we evaluate three different triggering ratios, $\gamma = \{0.3, 0.5, 0.7\}$, in each trial. 

%%%%%%%%%%%%%%%%%%%%%%%%%%%%%%%%%%%%%%%%%%

Figure~\ref{fig:compare_one_tri_per}-(a), (b) \& (c) show that the \textit{offline assignment} performs the best in terms of the arrival time, which is intuitive given that it knows the arrival times exactly, while the other three online assignments have the uncertain knowledge of the arrival times only. We observe that ATA with some suitable triggering ratios (e.g., $\gamma =0.3, 0.5, 0.7$ in Fig.~\ref{fig:compare_one_tri_per}-(a), $\gamma =0.3, 0.5$ in Fig.~\ref{fig:compare_one_tri_per}-(b), and $\gamma =0.3$ in Fig.~\ref{fig:compare_one_tri_per}-(c)) performs close to the offline strategy, which shows the effectiveness of ATA. Also, note that ATA and the \textit{all-step} assignment reach all demand locations earlier than the \textit{one-step} assignment. This is expected since in ATA and \textit{all-step} assignment, the vehicles are reassigned to reduce arrival time using the real-time information, while \textit{one-step} assignment only assigns the vehicles at the initial time step.  For the same reason, \textit{one-step} assignment has less running time and fewer assignments, as shown in Figure~\ref{fig:compare_one_tri_per}-(d) to (i).

Figures~\ref{fig:compare_one_tri_per}-(d) to (i) show that ATA schedules fewer assignments and use less running time than that of the \textit{all-step} assignment. In particular, Figures~\ref{fig:compare_one_tri_per}-(d), (e), \& (f) show that ATA with $\gamma=0.5$ avoids at least half of the assignments compared to \textit{all-step} assignment. 
% Also, Figures~\ref{fig:compare_one_tri_per}-(g), (h) \& (i) show that the all-time assignment has longer running time when the problem becomes larger.  
Even with fewer assignments, ATA performs comparably with or in some cases even better than \textit{all-step} assignment in terms of the arrival time as shown in Figure~\ref{fig:compare_one_tri_per}-(a), (b) \& (c). Particularly, with respect to the average of arrival time, ATA with triggering ratio $\gamma = 0.5$, $\gamma = 0.5$, and $\gamma = 0.3$ performs better than \textit{all-step} assignment in 3-vehicle 2-demand (Fig.~\ref{fig:compare_one_tri_per}-(a)), 6-vehicle 4-demand (Fig.~\ref{fig:compare_one_tri_per}-(b)), and 12-vehicle 5-demand (Fig.~\ref{fig:compare_one_tri_per}-(c)) systems, respectively. This result echoes the analysis at the end of Section~\ref{sec:adaptive_submodular} that more assignments may not always help because the assignment using the traffic conditions revealed so far is nearsighted. 

We also observe from Figure~\ref{fig:compare_one_tri_per}-(d) to (i) that a larger triggering ratio $\gamma$ leads to more assignments and more running time of ATA. This is because, with a larger triggering ratio, the assignment will be triggered more frequently as shown in Algorithm~\ref{alg:online_tri_assign_gen}, line~\ref{line:tri_cond_gen}.   Particularly,  
for three triggering ratios $\gamma = 0.3, 0.5, 0.7$, ATA with $\gamma = 0.3$ schedules fewer assignments (Fig.~\ref{fig:compare_one_tri_per}-(d), (e) \& (f)) and spends less running time (Fig.~\ref{fig:compare_one_tri_per}-(g), (h) \& (i)), while achieving a comparable (Fig.~\ref{fig:compare_one_tri_per}-(a) \& (b)) or even better (Fig.~\ref{fig:compare_one_tri_per}-(c)) performance than that of ATA with $\gamma = 0.5, 0.7$. This again verifies the fact that more assignments during the online mobility-on-demand process may be unhelpful and even misleading.

\section{Conclusion and Discussion}\label{sec:conclue}

We studied a risk-aware discrete submodular maximization problem. We provided the first positive results for discrete CVaR submodular maximization for selecting a set under matroidal constraints. In particular, we proposed the Sequential Greedy Algorithm and analyzed its approximation ratio and the running time. We also studied the problem of the adaptive risk-aware submodular problem for the online version of the mobility-on-demand and environmental monitoring cases. We presented a triggering replanning strategy for this problem. In particular, for online mobility-on-demand, we designed an adaptive triggering assignment strategy  (Algorithm~\ref{alg:online_tri_assign_gen}) to reduce unnecessary computation. We demonstrated the two practical use-cases of the CVaR submodular maximization problem and verified the effectiveness of adaptive triggering assignment approach (Algorithm~\ref{alg:online_tri_assign_gen}). 

Notably, our Sequential Greedy Algorithm works for any matroid constraint. In particular, the multiplicative approximation ratio can be improved to $1/c_f(1-e^{-c_f})$ if we know that the constraint is a uniform matroid~\cite[Theorem 5.4]{conforti1984submodular}. 

The additive term in our analysis depends on $\alpha$. This term can be large when the risk level $\alpha$ is very small. Even though, in practice, SGA works relatively well on average with small risk levels (Figs.~\ref{fig:gre_bf_compare_assign} \& \ref{fig:gre_bf_compare_sensor}), our ongoing work is to remove this dependence on $\alpha$, perhaps by designing another algorithm specifically for low risk levels. We note that if we use an optimal algorithm instead of the greedy algorithm as a subroutine, then the additive term disappears from the approximation guarantee (e.g., the sequential brute-force algorithm). The algorithm also requires knowing $\Gamma$. We showed how to find $\Gamma$ (or an upper bound for it) for the two case studies considered in this paper. Devising a general strategy for finding $\Gamma$ is part of our ongoing work. 

Note that, in the adaptive triggering assignment (ATA), increasing the triggering ratio $\gamma$ results in more assignments but may not always improve the performance (i.e., achieving a shorter arrival time) as shown in Figure~\ref{fig:compare_one_tri_per}. Thus, our second future research direction is to design a generic approach for deciding the best triggering ratio $\gamma$.

Our third line of ongoing work focuses on applying the risk-aware strategy to multi-vehicle routing, patrolling, and informative path planning in dangerous environments~\cite{jorgensen2018team} and mobility on demand with real-world data sets (2014 NYC Taxicab Factbook).\footnote{\url{http://www.nyc.gov/html/tlc/downloads/pdf/2014_taxicab_fact_book.pdf}}

\rev{In the mobility-on-demand application, we define the objective function as the sum of arrival utilities at all demand locations (Eq.~\ref{eqn:fsy_assign}). The ``sum'' metric captures total assignment utility and is commonly used in vehicle assignment problems (see~\cite{prorok2019redundant,prorok2020robust}). However, in the mobility-on-demand scenario, one typically expects all demand locations to be reached by vehicles within the least time. Hence, a more desirable objective would be to maximize the \textit{minimum arrival utility} (or equivalently, to minimize the \textit{maximum arrival time}) to each demand location~\cite{malencia2021fair}. However, the minimum arrival utility (or maximum arrival time) is not submodular (or supermodular) in the decision set $\mathcal{S}$, so that Algorithm~\ref{alg:sga} may not be directly utilized or have a bounded approximation guarantee to optimize it. Therefore, our fourth future research avenue is to revisit the mobility-on-demand application with the objective of maximizing minimum arrival utility (or minimizing maximum arrival time) and, building on Algorithm~\ref{alg:sga}, design corresponding risk-aware algorithms with hard guarantees.}

As a final comment, we discuss the reason for choosing CVaR over other risk-aware measures such as mean-variance~\cite{markowitz1952portfolio} and VaR~\cite{yang2017algorithm}. In general, this is a design choice and CVaR may not be the best choice in some cases. Generally speaking, for the utility with Gaussian distribution, it is more convenient to use the mean-variance measure given that it has a closed-form expression~\cite{chung2019risk} and is equivalent to CVaR with appropriate risk parameters selected. However, since it needs to know the mean and variance of the distribution, it does not work for arbitrary distributions (without a closed-form expression). VaR measures how much a low utility we might get with a given probability. If the distribution of the utility is smooth (or stable), VaR is sufficient for measuring the risks. However, the more fluctuant the utility, the greater the chance that VaR will not quantify the risks well, as it only cares about a single cutoff point. CVaR is a more suitable measure for this scenario by considering all the bad tailed cases. CVaR is also more desirable for the cases where the bad low probability (tailed) events have large value or impact and can cause huge losses since it is more conservative than VaR (Fig.~\ref{fig:var_cvar}). Further, CVaR has nicer properties than mean-variance and VaR for capturing risks in robotics~\cite{majumdar2020should} and is easier for optimization when it is computed by drawing samples~\cite{rockafellar2000optimization}.

% % use section* for acknowledgment
% \section*{Acknowledgement}\label{sec:Acknow}
% This material is based upon work supported by the NSF under grant number IIS-1637915 and ONR under grant number N00014- 18-1-2829.

\appendix
\section{Appendix}\label{sec:appendix} 
We start by describing the settings for the online mobility-on-demand with street networks. We then introduce the corresponding OTA in Algorithm~\ref{alg:online_tri_assign} based on these settings. After that, we provide proofs for the lemmas and the theorem in this paper. 

\subsection{Online Mobility-on-Demand with Street Networks}
\label{subsec:online_assign_setting} 
We consider the vehicles travel on a street network $\mathcal{G}=\{\mathcal{V}, \mathcal{W}\}$ with the intersections as nodes $\mathcal{V}=\{v_1, v_2, \cdots, v_V\}$ and the drive-way connecting two adjacent intersections, $v_i$ and $v_j$, as edges $w_{v_i, v_j} \in \mathcal{W}\subseteq \mathcal{V}\times \mathcal{V}$. Notably, the street network $\mathcal{G}$ is a directed graph because of the driving direction on the street. 
We use $\texttt{deg}(v_i)$ to denote the degree of a node, which is the number of the incoming and outgoing edges of the node. 

For an edge $w_{v_i, v_j}$, we denote its the length by $\texttt{len}(w_{v_i, v_j})$. Denote a path $P_{v_i, v_k}$ from node $v_i$ to node $v_k$ as a series of orderly connected edges on the graph with the following form $\{(v_i, v_j), (v_j, v_m), \cdots, (v_n, v_k)\}$. We compute the length $\texttt{len}(P_{v_i, v_k})$ and the degree $\texttt{deg}(P_{v_i, v_k})$ of path $P_{v_i, v_k}$ by the summation of the lengths of all edges and the summation of the degrees of all nodes on the path, respectively. Denote the set of all nodes on path $P_{v_i, v_k}$ as $\mathcal{V}(P_{v_i, v_k})$.

% and follows a truncated normal distribution, 
% \begin{equation}
%     t(v_i) \sim \mathcal{N}^{\texttt{Truc}}(0, \beta_1 \texttt{deg}(v_i)), ~~t(v_i) \in [0, T(v_i)] %$\overline{T}(v_i)$
%     \label{eqn:node_wait_time}
% \end{equation} where $T(v_i)$ denotes the maximum waiting time at the intersection $v_i$. We set the variance of the waiting time to be proportional to the degree of the intersection, i.e., $\beta_1 \texttt{deg}(v_i)$ with $\beta_1 \in (0, +\infty)$. Because, for an intersection, a larger degree means more incoming and outgoing edges, and thus the traffic at this intersection is more likely to be congested, which increases the probability of higher waiting time. Notably, the truncated Gaussian distribution is just one possible way to capture the randomness of the waiting time. Our algorithm works for any other reasonable distribution model.

% To model this real-time scenario, we set the real-time waiting time at an intersection $v_i$ for each vehicle $i$ as a sample from the truncated normal distribution distribution (Eq.~\ref{eqn:node_wait_time}).

Due to the unpredictable traffic condition at the intersection (node), we assume that the waiting time at an unseen intersection $v_i$ is a random variable, denoted as $t(v_i)$. We assume the variance of the waiting time, $\texttt{var}(t(v_i))$ is proportional to the degree of the intersection. Because, for an intersection, a larger degree means more incoming and outgoing edges, and thus the traffic condition at this intersection can have larger uncertainty. 
% The distribution of $t(v_i)$ can be obtained by sampling from the historical traffic data. 
However, when vehicle $j$ arrives at an intersection, it can acquire the real-time traffic condition and the waiting time $t_j^{\texttt{wait}}$ at the intersection. 

We compute the time to traverse an edge $w_{v_i, v_j}$ by 
\begin{align}
    t(w_{v_i, v_j}) = \beta_2 \texttt{len}(w_{v_i, v_j})/\texttt{maxv}(w_{v_i, v_j})
    \label{eqn:edge_travel_time}
\end{align} where $\texttt{maxv}(w_{v_i, v_j})$  denotes the limited speed on edge $w_{v_i, v_j}$ and $\beta_2 \in [1, +\infty)$. Here,  we do not consider the uncertain traffic condition when the vehicle is travelling on the edges. Thus, as shown in Equation~\ref{eqn:edge_travel_time}, the edge travel time can be obtained beforehand. 
% Notably, the edge travel time is a deterministic variable, since we do not consider the uncertain traffic condition when the vehicle is travelling on the edge.

Then, the time to traverse a path $P_{v_i, v_j}$ combines the waiting time at all the intersections and the time to travel through all the edges on the path. That is, 
\begin{align}
    t(P_{v_i, v_j}) = \sum_{v \in \mathcal{V}(P_{v_i, v_j})}t(v) + \beta_2 \sum_{w \in P_{v_i, v_j}} t(w)
    \label{eqn:path_travel_time}
\end{align} Notably, the path travel time $t(P_{v_i, v_j})$ is a random variable with its randomness induced by the waiting time. Similarly, we compute the path travel time by using a sampling method---taking an equal number of samples of the waiting time at each intersection and then using  Equation~\ref{eqn:path_travel_time} with appropriate dimension manipulation. Clearly, the arrival \rev{utility} to $v_j$ from $v_i$ through traveling along path $P_{v_i, v_j}$ can be computed as
\begin{align}
e(P_{v_i, v_j}) = 1/t(P_{v_i, v_j}).
    \label{eqn:path_travel_effi}
\end{align}
 
We denote the time that each vehicle $j$ needs to arrive at the next intersection as $t_j^{\texttt{next}}$. Then the per-step time interval is computed as 
\begin{equation}
    T^{\texttt{step}} = \min_{j\in\{1,\cdots, R\}} ~ t_j^{\texttt{next}}
    \label{eqn:one_step}
\end{equation} Notably, the per-step time interval $T^{\texttt{step}}$  depends on the $t_j^{\texttt{next}}$  from all the vehicles and can be changing at each time step. Also, when computing $T^{\texttt{step}}$ at each step, some vehicles may wait at the intersections whereas the others may be travelling on the edges. 

% We denote, for each vehicle $i$, its assigned demand location by $d_{j(i)}$. 
% Considering the speed limit on the road, we use $\bar{l}$ to denote the maximum length that each vehicle can travel at each time-step at the current edge. 
For simplicity, we assume  the vehicles' initial positions and the demand locations are some nodes (intersections) of graph $\mathcal{G}$.~\footnote{Indeed, if the positions of some vehicles are not at the intersections, we can easily create corresponding virtual intersections for them and add the virtual intersections on the graph. } 
\subsection{Online Triggering Assignment on Street networks}
\label{subsec:trigger_assign}
In this section, we explicitly introduce our online triggering assignment strategy on street networks in Algorithm~\ref{alg:online_tri_assign}. 

% *** Online Triggering Assignment ALG *** 
\begin{algorithm}[t]
\caption{Triggering Assignment on Street Networks}
\begin{algorithmic}[1]
\REQUIRE 
\begin{itemize}
\item Graph $\mathcal{G}$ with intersections $\mathcal{V}$ and edges $\mathcal{W}$

\item $N$ demand locations $d_i$ with corresponding intersections, $v(d_i)$, $i\in\{1, \cdots, N\}$. $R$ vehicles' initial positions $p_j$ with corresponding intersections, $v(p_j)$, $j\in \{1, \cdots, R\}$ 
\end{itemize}

\STATE Find a shortest path $P^{\star}_{v(p_j), v(d_i)}$ from each $v(p_j)$ to each $v(d_i)$. Collect these shortest paths as $\{P^{\star}_{v(p_j), v(d_i)}\}$
\label{line:ini_short_path}

\STATE Compute the arrival \rev{utility} of these shortest paths as $e(\{P^{\star}_{v(p_j), v(d_i)}\})$ by Equations~\ref{eqn:path_travel_time},~\ref{eqn:path_travel_effi}
\label{line:compu_path_effi}

\STATE  $\mathcal{S} =\text{SGA}(e(\{P^{\star}_{v(p_j), v(d_i)}\}), \alpha)$
\label{line:ini_assign}

\STATE Set $v_j^{\texttt{prev}} = \emptyset, ~t_j^{\texttt{next}} =0, ~j \in \{1, \cdots, R\}$ and $T^{\texttt{step}} = 0$
\label{line:ini_pre_intersect_min_next}

\WHILE{not all $N$ demand locations are reached}
\label{line:termin_con}

    \FOR{each vehicle $j \in \{1, \cdots, R\}$}
    \label{line:for_step_time_update}

        \IF{$v(p_j) \neq v_j^{\texttt{prev}}$}
        \label{line:if_new_intersection}
        
            \STATE updates $v_j^{\texttt{prev}} \leftarrow v(p_j)$
            \label{line:update_prev_intersection}

            \STATE acquires real-time $t_i^{\texttt{wait}}$ at intersection $v(p_j)$
            \label{line:sample_wait_time}

            \STATE computes $t_j^{\texttt{next}} = t_j^{\texttt{wait}} + t(e_{v(p_j), v(p_j)+1})$ (Eq.~\ref{eqn:edge_travel_time})
            \label{line:full_next_time}

        \ELSE
        \label{line:else_old_intersection}
        
            \STATE $t_j^{\texttt{next}} = t_j^{\texttt{next}} - T^{\texttt{step}}$
            \label{line:reduced_next_time}

        \ENDIF
        \label{line:end_if_intersection}
    \ENDFOR
    \label{line:for_step_time_update_end}
    
    \STATE update $ T^{\texttt{step}} = \min_{j\in\{1,\cdots, R\}} ~ t_j^{\texttt{next}}$ (Eq.~\ref{eqn:one_step})
    \label{line:one_step_interval}

    \FOR{each vehicle $j \in \{1, \cdots, R\}$}
    \label{line:for_online_travel}

        \IF{$t_j^{\texttt{next}} == T^{\texttt{step}}$}
         \label{line:equal_to_min_next_time}
         
            \STATE updates $v(p_j) \leftarrow v(p_j)+1$
            \label{line:goto_next_intersection}

        \ELSIF{$t_j^{\texttt{wait}}>= T^{\texttt{step}}$}
        \label{line:wait_larger_min_next}
        
            \STATE updates $t_j^{\texttt{wait}} = t_j^{\texttt{wait}} - T^{\texttt{step}}$
            \label{line:wait_time_reducing}

        \ELSIF{$t_j^{\texttt{wait}}< T^{\texttt{step}}$}
        \label{line:wait_less_min_next}
        
            \STATE updates $t_j^{\texttt{wait}} = 0$
            \label{line:wait_time_zero}
            
            \STATE travels $\texttt{maxv}(e_{v(p_j), v(p_j)+1})(t_j^{\texttt{next}} - T^{\texttt{step}})/\beta_2$ distance on the current edge $e_{v(p_j), v(p_j)+1}$
            \label{line:travel_some_distance}
            
        \ENDIF
        \label{line:if_online_travel}
        
    \ENDFOR
    \label{line:for_online_end}

    \STATE Check for all demand locations $d_i, ~i\in \{1,\cdots, N\}$
    \label{line:check_demand}

    \IF{$ \exists~ \texttt{len}(P^{\star}_{v(p_j), v(d_{i})}) \leq \gamma \texttt{len}(P^{\star}_{v(p_j'), v(d_{i})})$  \text{and} 	 $~\texttt{deg}(P^{\star}_{v(p_j), v(d_{i})}) \leq \texttt{deg}(P^{\star}_{v(p_j'), v(d_{i})})$}
    \label{line:if_trigger}
    
         \STATE Find a shortest path $P^{\star}_{v(p_j), v(d_i)}$ from each $v(p_j)$ to each $v(d_i)$. Collect these shortest paths as $\{P^{\star}_{v(p_j), v(d_i)}\}$
        \label{line:reassign_short_path}

        \STATE Compute the arrival \rev{utility} of these shortest paths as $E(\{P^{\star}_{v(p_j), v(d_i)}\})$ by Equations~\ref{eqn:path_travel_time},~\ref{eqn:path_travel_effi}
        \label{line:reassign_path_effi}

        \STATE  $\mathcal{S} =\texttt{SGA}(e(\{P^{\star}_{v(p_j), v(d_i)}\}), \alpha)$, 
        \label{line:re_assign}  
        
    \ENDIF
    \label{line:if_assign_end}
    
\ENDWHILE
\label{line:end_while_loop}

\end{algorithmic}
\label{alg:online_tri_assign}
\end{algorithm}
% *** Online Triggering Assignment ALG END *** 

\textbf{Initial assignment (Alg.~\ref{alg:online_tri_assign}, lines~\ref{line:initiliaze}-\ref{line:ini_assign}).} We first find a shortest path from each vehicle's initial position to each demand location by using Dijkstra's algorithm (Alg.~\ref{alg:online_tri_assign}, line~\ref{line:initiliaze}). Then, we use Equations~\ref{eqn:path_travel_time},~\ref{eqn:path_travel_effi} to compute the arrival \rev{utility} of these shortest paths (Alg.~\ref{alg:online_tri_assign}, line~\ref{line:compu_path_effi}). Finally, we select a risk level $\alpha$ and use SGA (Alg.~\ref{alg:sga}) to get an initial vehicle-to-demand assignment $\mathcal{S}$ with $\mathcal{S}= \bigcup_{i=1}^{N} \mathcal{S}_i$ (Alg.~\ref{alg:online_tri_assign}, line~\ref{line:ini_assign}).  

After the initial assignment, each vehicle $j$ knows its assigned demand location $d_{i}$ with the corresponding shortest path, $P^{\star}_{v(p_j), v(d_{i})}$ (computed in Alg.~\ref{alg:online_tri_assign}, line~\ref{line:initiliaze}). Then each vehicle travels on its shortest path towards its assigned demand location. Notably, the demand location and the corresponding (shortest) path assigned to each vehicle can be changing because of the online reassignment later on. 

Our terminated condition is to guarantee that all demand locations are reached (Alg.~\ref{alg:online_tri_assign}, line~\ref{line:termin_con}), which needs an underlying assumption that the number of vehicles is larger than the number of demand locations, i.e., $R \geq N$.  

\textbf{Per-step time interval update (Alg.~\ref{alg:online_tri_assign}, lines~\ref{line:ini_pre_intersect_min_next},~\ref{line:for_step_time_update}-\ref{line:one_step_interval}).} 
% We assume each vehicle $i$ is at the intersection or in the middle of some edge $e_{v(p_i), v(p_i)+1}$. 
To update the per-step time interval $T^{\texttt{step}}$ at each time step, each vehicle $j$ stores the previous intersection it was in, $v_j^{\texttt{prev}}$, which is initialized by an empty set (Alg.~\ref{alg:online_tri_assign}, line~\ref{line:ini_pre_intersect_min_next}). 

\begin{itemize}
\item If vehicle $j$ arrives at a new intersection $v(p_j)$ (Alg.~\ref{alg:online_tri_assign}, line~\ref{line:if_new_intersection}), it updates its previous intersection by the current new intersection (Alg.~\ref{alg:online_tri_assign}, line~\ref{line:update_prev_intersection}) and acquires the real-time waiting time at the new intersection $v(p_j)$ (Alg.~\ref{alg:online_tri_assign}, line~\ref{line:sample_wait_time}). Also, it computes the time to spend for arriving at the next intersection $v(p_j) + 1$ as $t_j^{\texttt{next}}$ (Alg.~\ref{alg:online_tri_assign}, line~\ref{line:full_next_time}). 

\item While, if vehicle $j$ has not reached a new intersection (Alg.~\ref{alg:online_tri_assign}, line~\ref{line:else_old_intersection}), it reduces the time to reach the new intersection by the per-step time interval (at the last time step) $T^{\texttt{step}}$ (Alg.~\ref{alg:online_tri_assign}, line~\ref{line:reduced_next_time}). 

\item After all vehicles computing  $t_j^{\texttt{next}}$, they select the minimum $t_j^{\texttt{next}}$ as the current per-step time interval (Alg.~\ref{alg:online_tri_assign}, line~\ref{line:one_step_interval}).
\end{itemize}

\textbf{Per-step online travel (Alg.~\ref{alg:online_tri_assign}, lines~\ref{line:for_online_travel}-\ref{line:for_online_end}).} Based on the current per-step time interval $T^{\texttt{step}}$, each vehicle $j$ can decide its motion at the current time step. 

\begin{itemize}
    \item if vehicle $j$ can reach the next intersection $v(p_j)+1$ within per-step time $T^{\texttt{step}}$ (Alg.~\ref{alg:online_tri_assign}, line~\ref{line:equal_to_min_next_time}), it moves to $v(p_j)+1$ and updates its current intersection $v(p_j)$ by the next intersection $v(p_j)+1$ (Alg.~\ref{alg:online_tri_assign}, line~\ref{line:goto_next_intersection}). 
   
    \item if vehicle $j$'s waiting time $t_j^{\texttt{wait}}$ at the current intersection $v(p_j)$ is larger than per-step time $T^{\texttt{step}}$ (Alg.~\ref{alg:online_tri_assign}, line~\ref{line:wait_larger_min_next}), it still needs to wait at $v(p_j)$ after per-step time $T^{\texttt{step}}$, but its waiting time is reduced by $T^{\texttt{step}}$ (Alg.~\ref{alg:online_tri_assign}, line~\ref{line:wait_time_reducing}). 
    
    \item if vehicle $j$'s waiting time $t_j^{\texttt{wait}}$ at the current intersection $v(p_i)$ is less than per-step time $T^{\texttt{step}}$ (Alg.~\ref{alg:online_tri_assign}, line~\ref{line:wait_less_min_next}), it will leave the current intersection $v(p_j)$ or it is already travelling on the current edge $e_{v(p_j), v(p_j)+1}$, i.e., $t_j^{\texttt{wait}} = 0$. In both cases, even though vehicle $j$ cannot reach the next intersection $v(p_j)+1$ (since $t_j^{\texttt{next}}>T^{\texttt{step}}$), it will traverse some distance on the current edge (Alg.~\ref{alg:online_tri_assign}, line~\ref{line:travel_some_distance}), which is computed by leveraging on Equation~\ref{eqn:edge_travel_time} as, 
    \begin{align*}
       & \texttt{len}(e_{v(p_j), v(p_j)+1}) \frac{t_j^{\texttt{next}}-T^{\texttt{step}}}{t(e_{v(p_j), v(p_j)+1})} \\
       & = \texttt{len}(e_{v(p_j), v(p_j)+1}) \frac{t_j^{\texttt{next}}-T^{\texttt{step}}}{\beta_2 \frac{\texttt{len}(e_{v(p_j), v(p_j)+1})}{\texttt{maxv}(e_{v(p_j), v(p_j)+1})}} \\
       & = \texttt{maxv}(e_{v(p_j), v(p_j)+1})(t_j^{\texttt{next}} - T^{\texttt{step}})/\beta_2
        % \label{eqn:dis_travel_edge}
    \end{align*}
\end{itemize}

\textbf{Triggering reassignment (Alg.~\ref{alg:online_tri_assign}, lines~\ref{line:check_demand}-\ref{line:if_assign_end}).} We trigger the reassignment when some condition happens. Recall that, for each demand location $d_i$, it has a set of vehicles $\mathcal{S}_i$ assigned to it. Thus, in total, there are $|\mathcal{S}_i|$ shortest paths to demand $d_i$. 
% Denote the shortest paths with the maximum length and the minimum length to demand location $d_j$ as $\bar{P}^{\star}_{j}$ and $\underaccent{\bar}{P}^{\star}_{j}$. 
We check for all demand locations (Alg.~\ref{alg:online_tri_assign}, line~\ref{line:check_demand}) and trigger the reassignment as long as there exists a demand location $i$ with, 
\begin{align}
    & \texttt{len}(P^{\star}_{v(p_j), v(d_{i})}) \leq \gamma \texttt{len}(P^{\star}_{v(p_j'), v(d_{i})}), ~\text{and} \label{eqn:tri_1}\\
    & \texttt{deg}(P^{\star}_{v(p_j), v(d_{i})}) \leq \texttt{deg}(P^{\star}_{v(p_j'), v(d_{i})}), \label{eqn:tri_2}
\end{align}
where $\gamma \in (0,1)$ denotes the triggering ratio (Alg.~\ref{alg:online_tri_assign},  line~\ref{line:if_trigger}). Because, in this case, there exists a path $P^{\star}_{v(p_j), v(d_{i})}$ that has a  shorter travel length and smaller travel time uncertainty (remember we use the degree to capture the waiting time uncertainty), than that of the other path $P^{\star}_{v(p_j'), v(d_{i})}$. Thus, there is no need to continue assigning the vehicle with path $P^{\star}_{v(p_j'), v(d_{i})}$ to the  demand $i$. We therefore trigger the reassignment and hopefully assign this vehicle to other demand location to reduce the total travel time (Alg.~\ref{alg:online_tri_assign}, lines~\ref{line:reassign_short_path}-\ref{line:re_assign}). Notably, the triggering ratio regulates the frequency of reassignment, i.e., a larger $\gamma$ triggers more reassignments.

\subsection{Proofs}
\noindent \textbf{Proof of Lemma~\ref{lem:auxiliary_function}:}
\begin{proof}
Since $f(\mathcal{S},y)$ is monotone increasing and submodular in $\mathcal{S}$, $\text{max}\{\tau-f(\mathcal{S},y), 0\}$ is monotone decreasing and supermodular in $\mathcal{S}$, and its expectation is also monotone decreasing and supermodular in $\mathcal{S}$. Then $H(\mathcal{S},\tau)$ is monotone increasing and submodular in $\mathcal{S}$. 

$H(\emptyset,\tau) = \tau(1-\frac{1}{\alpha})$ given $f(\mathcal{S},y)$ is normalized ($f(\emptyset, y) = 0$). Thus, $H(\mathcal{S},\tau)$ is not necessarily normalized since $\tau$ is not necessarily zero. See a similar proof in~\cite{rockafellar2000optimization,maehara2015risk}. 
\label{proof:sub_}
\end{proof}

\vspace{5pt} \noindent \textbf{Proof of Lemma~\ref{lem:auxi_concave}:}
\begin{proof} Since $\text{max}(\tau-f(\mathcal{S},y), 0)$ is convex in $\tau$, its expectation is also convex in $\tau$. Then $- \frac{1}{\alpha}\mathbb{E}[\text{max}(\tau-f(\mathcal{S},y), 0)]$ is concave in $\tau$ and $H(\mathcal{S},\tau)$ is concave in $\tau$. 
\label{proof:aux_concave}
\end{proof}

\vspace{5pt} \noindent \textbf{Proof of Lemma~\ref{lem: gradient_auxiliary_function}:}
\begin{proof}
By using the result in ~\cite[Lemma 1 and Proof of Theorem 1]{rockafellar2000optimization}, we know that $H(\mathcal{S},\tau)$ is concave and continuously differentiable with derivative given by
$$\frac{\partial H(\mathcal{S},\tau)}{\partial \tau} = 1 - \frac{1}{\alpha} (1 - \Phi(f(\mathcal{S}, y)))$$ where $\Phi(f(\mathcal{S}, y))$ is the cumulative distribution function of $f(\mathcal{S}, y)$. Thus, $0 \leq \Phi(f(\mathcal{S}, y))) \leq 1$, which proves the lemma. 
\label{proof:gradient_auxiliary_fun}
\end{proof}

\vspace{5pt} \noindent \textbf{Proof of the number of samples $n_s$:}
\begin{proof}
We reach the statement by using the proof by Ohsaka and Yoshida~\cite[Lemmas 4.1-4.5]{ohsaka2017portfolio}. We have via taking $c = \Gamma$ (the upper bound of the search separation $\tau$), in their Lemma 4.4. Then, in their Lemma 4.5, to make 
\begin{equation}
      |\text{CVaR}_{\alpha}(X) - \text{CVaR}_{\alpha}(\hat{X})|\leq\epsilon,
      \label{eqn:sample_error}
\end{equation}
we need to set 
\begin{equation}
  \text{sup}_{x \in \mathbb{R}} |F_X(x) - F_{\hat{X}}(x)| \leq \frac{\epsilon}{\Gamma}. 
  \label{eqn:DKWinq}
\end{equation}
By Dvoretzky-Kiefer-Wolfowitz inequality, Inequality~\ref{eqn:DKWinq} fulfills with probability at least $1-2\exp(-2n_s\frac{\epsilon^2}{\Gamma^2})$. Thus, by setting $n_s = O(\frac{\Gamma^2}{\epsilon^2}\log \frac{1}{\delta}), ~\delta, \epsilon \in (0,1)$, we have $1-2\exp(-2n_s\frac{\epsilon^2}{\Gamma^2}) = 1 - \delta$.   
\end{proof}

\begin{figure}
\centering{
\subfigure[$H(\mathcal{S}, \tau)$ is under the red dotted line.]
{\includegraphics[width=0.48\columnwidth]{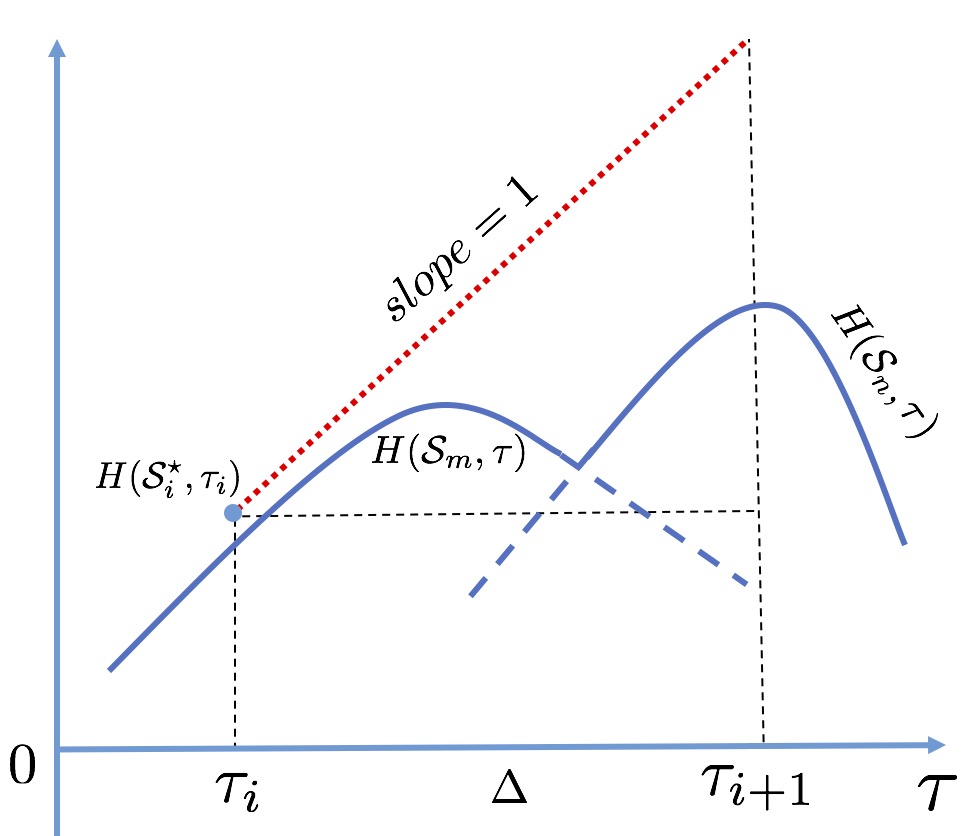}}~~
\subfigure[$H(\mathcal{S}, \tau)$ is under the red dotted line and green dotted line.]
{\includegraphics[width=0.48\columnwidth]{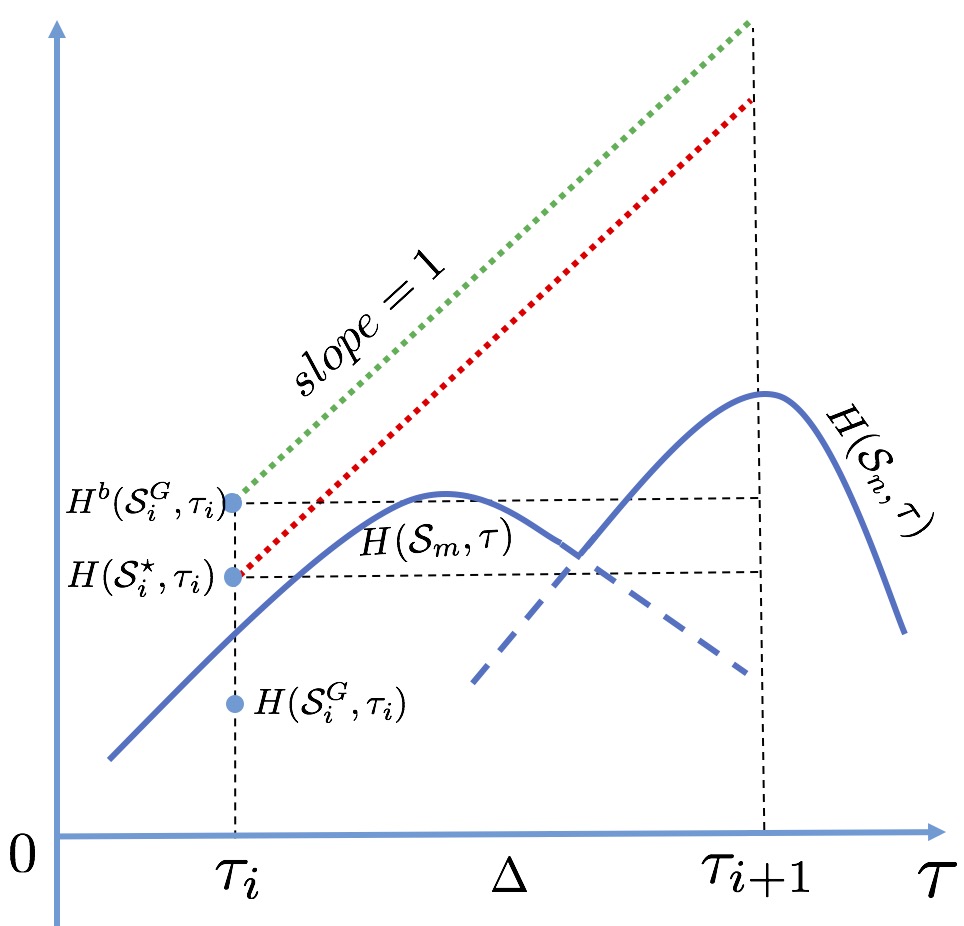}}
\caption{Illustration of $H(\mathcal{S}, \tau)$ within $\tau \in [\tau_i, \tau_{i+1}]$. 
\label{fig:h_tau_ij}}
}
\end{figure}

\vspace{5pt} \noindent \textbf{Proof of Lemma~\ref{lem:tau_sstar}:}
\begin{proof}
Denote $H_{i}^{\star} = {\text{max}}~H(\mathcal{S}, \tau)$ with ${\tau\in [i\Delta, (i+1)\Delta), ~\mathcal{S} \in \mathcal{I}}$. From Lemmas~\ref{lem:auxi_concave} and \ref{lem: gradient_auxiliary_function}, we know $H(\mathcal{S},\tau)$ is concave in $\tau$ and $\frac{\partial H(\mathcal{S},\tau)}{\partial \tau} <1$. The properties of concavity and bound on the gradient give 
$$H_{i}^{\star} - H(\mathcal{S}_{i}^{\star}, \tau_i) \leq \Delta.$$ We illustrate this claim by using Figure~\ref{fig:h_tau_ij}-(a). Since $\mathcal{S}_{i}^{\star}$ is the optimal set at $\tau_i$ for maximizing $H(\mathcal{S},\tau)$, the value of $H(\mathcal{S},\tau)$ with any other set $\mathcal{S} \in \mathcal{I}$ at $\tau_i$ is at most $H(\mathcal{S}_{i}^{\star}, \tau_i)$. That is, $H(\mathcal{S}, \tau_i)\leq H(\mathcal{S}_{i}^{\star}, \tau_i)$. Since $H(\mathcal{S},\tau)$ is a concave function of $\tau$ for any specific $\mathcal{S}$, $H(\mathcal{S},\tau)$ can be a single concave function, i.e., $H(\mathcal{S}_m, \tau)$ or $H(\mathcal{S}_n, \tau)$ or a piecewise concave function by a combination of several concave functions, i.e., a combination of $H(\mathcal{S}_m, \tau)$ and $H(\mathcal{S}_n, \tau)$ during $\tau \in [\tau_i, \tau_{i+1}]$ (Figure~\ref{fig:h_tau_ij}-(a)). In either case, $H(\mathcal{S}, \tau)$ is below the line starting at $H(\mathcal{S}_{i}^{\star}, \tau_i)$ with $slope = 1$ during $\tau \in [\tau_i, \tau_{i+1}]$ (the red dotted line in Fig.~\ref{fig:h_tau_ij}-(a)). Since $H(\mathcal{S}, \tau_i)\leq H(\mathcal{S}_{i}^{\star}, \tau_i)$ and  $H(\mathcal{S},\tau)$ has a bounded gradient $\frac{\partial H(\mathcal{S},\tau)}{\partial \tau} \leq 1$. Thus, $H_{i}^{\star} - H(\mathcal{S}_{i}^{\star}, \tau_i) \leq \frac{\partial H(\mathcal{S},\tau)}{\partial \tau}\Delta = \Delta, ~ \forall i\in \{0,1,\cdots, \ceil{\frac{\Gamma}{\Delta}}\}$. 

Then we have $ H_{i}^{\star} - \text{max}_{i} H(\mathcal{S}_{i}^{\star}, \tau_i)  \leq \Delta, ~ \forall i\in \{0,1,\cdots, \ceil{\frac{\Gamma}{\Delta}}\}$. Note that $H_{i}^{\star}$ is the maximum value of $H(\mathcal{S},\tau)$ at each interval $\tau\in [i\Delta, (i+1)\Delta)$. The maximum value of $H(\mathcal{S},\tau)$, $H(\mathcal{S}^{\star}, \tau^{\star})$ is equal to one of $H_{i}^{\star}, i\in \{0,1,\cdots, \ceil{\frac{\Gamma}{\Delta}}\}$. Thus, we reach the  claim in Lemma~\ref{lem:tau_sstar}.
\label{proof:tau_sstar_delta} 
\end{proof}

\vspace{5pt}
\noindent \textbf{Proof of Lemma~\ref{lem:rela_gre_opt_tau}:}
\begin{proof}
We use a the previous result~\cite[Theorem 2.3 ]{conforti1984submodular} for the proof of this claim. We know that for any given $\tau$, $H(\mathcal{S}, \tau)$ is a non-normalized monotone submdoular function in $\mathcal{S}$ (Lemma~\ref{lem:auxiliary_function}). For maximizing normalized monotone submodular set functions, the greedy approach can give a  $1+1/c_f$ approximation of the optimal performance with any matroid constraint~\cite[Theorem 2.3]{conforti1984submodular}. After normalizing $H(\mathcal{S}, \tau)$ by $H(\mathcal{S}, \tau) - H(\emptyset, \tau)$, we have 
% \begin{equation}
% \frac{H(\mathcal{S}_{i}^{G}, \tau_i) - H(\emptyset, \tau_i)}{H(\mathcal{S}_{i}^{\star}, \tau_i) - H(\emptyset, \tau_i)} \geq 1-1/e, 
% \label{eqn:cardinality_constraint}
% \end{equation}
% when the selected set satisfies a cardinality constraint (or a uniform matroid constraint), i.e., $|\mathcal{S}_{i}^{G}|\leq \kappa$ and $|\mathcal{S}_{i}^{\star}|\leq \kappa$. When the selected set satisfies a matroid constraint, $\mathcal{S} \in \mathcal{I}$, the greedy algorithm gives a $1/2$ approximation~\cite{fisher1978analysis}, which means
% \begin{equation}
% \frac{H(\mathcal{S}_{i}^{G}, \tau_i) - H(\emptyset, \tau_i)}{H(\mathcal{S}_{i}^{\star}, \tau_i) - H(\emptyset, \tau_i)} \geq 1/2, 
% \label{eqn:matroid_constraint}
% \end{equation}
% where $|\mathcal{S}_{i}^{G}|\in \mathcal{I}$ and $|\mathcal{S}_{i}^{\star}|\in \mathcal{I}$. More specifically, if we know the curvature of the ground set $\mathcal{X}$, $c_f$, the greedy algorithm can guarantee 
\begin{equation}
\frac{H(\mathcal{S}_{i}^{G}, \tau_i) - H(\emptyset, \tau_i)}{H(\mathcal{S}_{i}^{\star}, \tau_i) - H(\emptyset, \tau_i)} \geq \frac{1}{1+c_f}, 
\label{eqn:curvature_matroid_constraint}
\end{equation}
with any matroid constraint. Given $0 \leq c_f \leq 1$ and $H(\emptyset, \tau) = -\tau(\frac{1}{\alpha}-1)$, we transform Equation~\ref{eqn:curvature_matroid_constraint} into,
\begin{align}
H(\mathcal{S}_{i}^{G},{\tau}_i) &\geq  \frac{1}{1+c_f} H(\mathcal{S}_{i}^{\star},\tau_i) - \frac{c_f}{1+c_f} \tau_i(\frac{1}{\alpha} -1).
% \nonumber\\
% &\geq\frac{1}{1+c_f} H(\mathcal{S}_{i}^{\star},\tau_i) - \frac{c_f}{1+c_f} \Gamma(\frac{1}{\alpha} -1),
\label{eqn:sgstar_taub}
\end{align}
% where Equation~\ref{eqn:sgstar_taub} holds since $\Gamma$ is the upper bound of $\tau$. 
Thus, we prove the Lemma~\ref{lem:rela_gre_opt_tau}. 
\label{pro:lem_gre_opt_relative}
\end{proof}

%%%%%%%%%%%%%%%%%%%%%%%%%%%%%%%%%%
%%%%%%%%%%%%%%%%%%%%%%%%%%%%%%%%%%%%

\vspace{5pt} \noindent \textbf {Proof of Theorem~\ref{thm:appro_bound_compu}:}

\textbf{Approximation bound.} From Equation~\ref{eqn:appro_bound} in Lemma~\ref{lem:rela_gre_opt_tau}, we have $H(\mathcal{S}_{i}^{\star},\tau_i)$ is bounded by
\begin{equation}
H(\mathcal{S}_{i}^{\star},\tau_i) \leq (1+c_f)H(\mathcal{S}_{i}^{G},\tau_i) + c_f \tau_i(\frac{1}{\alpha} -1). 
\label{eqn:upper_hstar_tau}
\end{equation}
Denote this upper bound as $$H^{b}(\mathcal{S}_{i}^{G},\tau_i) : = (1+c_f)H(\mathcal{S}_{i}^{G},\tau_i) + c_f \tau_i(\frac{1}{\alpha} -1).$$
We know $H(\mathcal{S}, \tau)$ is below the line starting at $H(\mathcal{S}_{i}^{\star}, \tau_i)$ with $slope = 1$ during $\tau \in [\tau_i, \tau_{i+1}]$ (the red dotted line in Fig.~\ref{fig:h_tau_ij}-(a)/(b)) (Lemma~\ref{lem:tau_sstar}). $H(\mathcal{S}, \tau)$ must be also below the line starting at $H^{b}(\mathcal{S}_{i}^{G},\tau_i)$ with $slope = 1$ during $\tau \in [\tau_i, \tau_{i+1}]$ (the green dotted line in Fig.~\ref{fig:h_tau_ij}-(b)). Similar to the proof in Lemma~\ref{lem:tau_sstar}, we have $H_{i}^{\star} - H^{b}(\mathcal{S}_{i}^{G}, \tau_i) \leq \Delta$  and 
\begin{equation}
{\text{max}}_{i\in \{0,1,\cdots, \ceil{\frac{\Gamma}{\Delta}}\}} H^{b}(\mathcal{S}_{i}^{G}, \tau_i) \geq H(\mathcal{S}^{\star}, \tau^{\star}) -\Delta. 
\label{eqn: max_bound_hstar}
\end{equation}
% Then $H^{b}(\mathcal{S}_{i}^{G},\tau_i) \geq H(\mathcal{S}_{i}^{\star},\tau_i) \geq H_{i}^{\star} - \Delta$ as shown in Figure.  $H(\mathcal{S}, \tau)$ is below the line starting at $H(\mathcal{S}_{i}^{\star}, \tau_i)$ with $slope = 1$ during $\tau \in [\tau_i, \tau_{i+1}]$ (the red dotted line in Figure).
% Since $H(\mathcal{S}_{i}^{\star},\tau_i) \leq H^{b}(\mathcal{S}_{i}^{G},\tau_i), ~ \forall i\in \{0,1,\cdots, \ceil{\frac{\Gamma}{\Delta}}\}$, we have 
% $\text{max}_i H(\mathcal{S}_{i}^{\star},\tau_i) \leq \text{max}_i H^{b}(\mathcal{S}_{i}^{G},\tau_i), i\in \{0,1,\cdots, \ceil{\frac{\Gamma}{\Delta}}\}$. By Lemma~\ref{lem:tau_sstar}, we have 
% \begin{equation}
% \text{max}_i H^{b}(\mathcal{S}_{i}^{G},\tau_i) \geq \text{max}_i H(\mathcal{S}_{i}^{\star},\tau_i) \geq H(\mathcal{S}^{\star}, \tau^{\star}) -\Delta. 
% \label{eqn: max_bound_hstar}
% \end{equation}
SGA selects the pair $(\mathcal{S}^{G},\tau^{G})$ as the pair  $(\mathcal{S}_{i}^{G},\tau_i)$ with $\text{max}_i ~H(\mathcal{S}_{i}^{G},\tau_i)$. Then by Inequalities~\ref{eqn:upper_hstar_tau} and \ref{eqn: max_bound_hstar}, we have
\begin{eqnarray}
(1+c_f)H(\mathcal{S}^{G},{\tau}^{G}) + c_f \tau^G(\frac{1}{\alpha} -1) \geq H(\mathcal{S}^{\star}, \tau^{\star}) -\Delta.
\label{eqn:approx_final_derive}
\end{eqnarray}
By rearranging the terms, we get 
\begin{align}
H(\mathcal{S}^{G},{\tau}^{G}) \geq & ~\frac{1}{1+c_f}(H(\mathcal{S}^{\star},\tau^{\star}) - \Delta) \nonumber\\
& ~-\frac{c_f}{1+c_f}\tau^G(\frac{1}{\alpha} -1).
\label{eqn:appro_bound_2}
\end{align}
Note that we use an oracle $\mathcal{O}$ to approximate
$H(\mathcal{S}, \tau)$ with error $\epsilon$ by the sampling method (Eq.~\ref{eqn:sample_error}). By considering the sampling error  $\epsilon$, we reach the statement of the approximation bound in Theorem~\ref{thm:appro_bound_compu}.

\textbf {Computational time.} Next, we give the proof of the computational time of SGA in Theorem~\ref{thm:appro_bound_compu}. 
We verify the computational time of SGA by following the stages of the pseudo code in Algorithm~\ref{alg:sga}.  First, from Algorithm~\ref{alg:sga}, lines~\ref{line:search_tau_forstart} to \ref{line:search_tau_forend}, we use a ``for'' loop for searching $\tau$ which takes $\ceil{\frac{\Gamma}{\Delta}}$ evaluations. Second, within the ``for'' loop, we use the greedy algorithm to solve the subproblem (Alg.~\ref{alg:sga},  lines~\ref{line:gre_empty}--\ref{line:gre_while_end}). In order to select a subset $\mathcal{S}$ with size $|\mathcal{S}|$  from a ground set $\mathcal{X}$ with size $|\mathcal{X}|$, the greedy algorithm takes  $|\mathcal{S}|$ rounds (Alg.~\ref{alg:sga}, line~\ref{line:gre_while_start}), and calculates the marginal gain of  the remaining elements in $\mathcal{X}$ at each round (Alg.~\ref{alg:sga}, line~\ref{line:gre_while_margin}). Thus, the greedy algorithm takes $\sum_{i=1}^{|\mathcal{S}|} |\mathcal{X}|-i$ evaluations. Third, by calculating the marginal gain for each element, the oracle $\mathcal{O}$ samples $n_s$ times for computing $H(\mathcal{S}, \tau)$. Thus, overall, the ``for'' loop containing the greedy algorithm with the oracle sampling takes $\ceil{\frac{\Gamma}{\Delta}}(\sum_{i=1}^{|\mathcal{S}|} |\mathcal{X}|-i)n_s$ evaluations. Last, finding the best pair from storage set $\mathcal{M}$ (Alg.~\ref{alg:sga}, line~\ref{line:find_best_pair}) takes $O(\ceil{\frac{\Gamma}{\Delta}})$ time. Therefore, the computational complexity for SGA is, 
$$\ceil{\frac{\Gamma}{\Delta}}(\sum_{i=1}^{|\mathcal{S}|} |\mathcal{X}|-i)n_s + O(\ceil{\frac{\Gamma}{\Delta}}) = O(\ceil{\frac{\Gamma}{\Delta}} |\mathcal{X}|^{2} n_s),$$
given $|\mathcal{S}|\leq |\mathcal{X}|$. 
\label{prf:compu_time}

\bibliographystyle{IEEEtran}
\bibliography{myrefs.bib}

\begin{IEEEbiography}[{\includegraphics[width=1in,height=1.25in,clip,keepaspectratio]{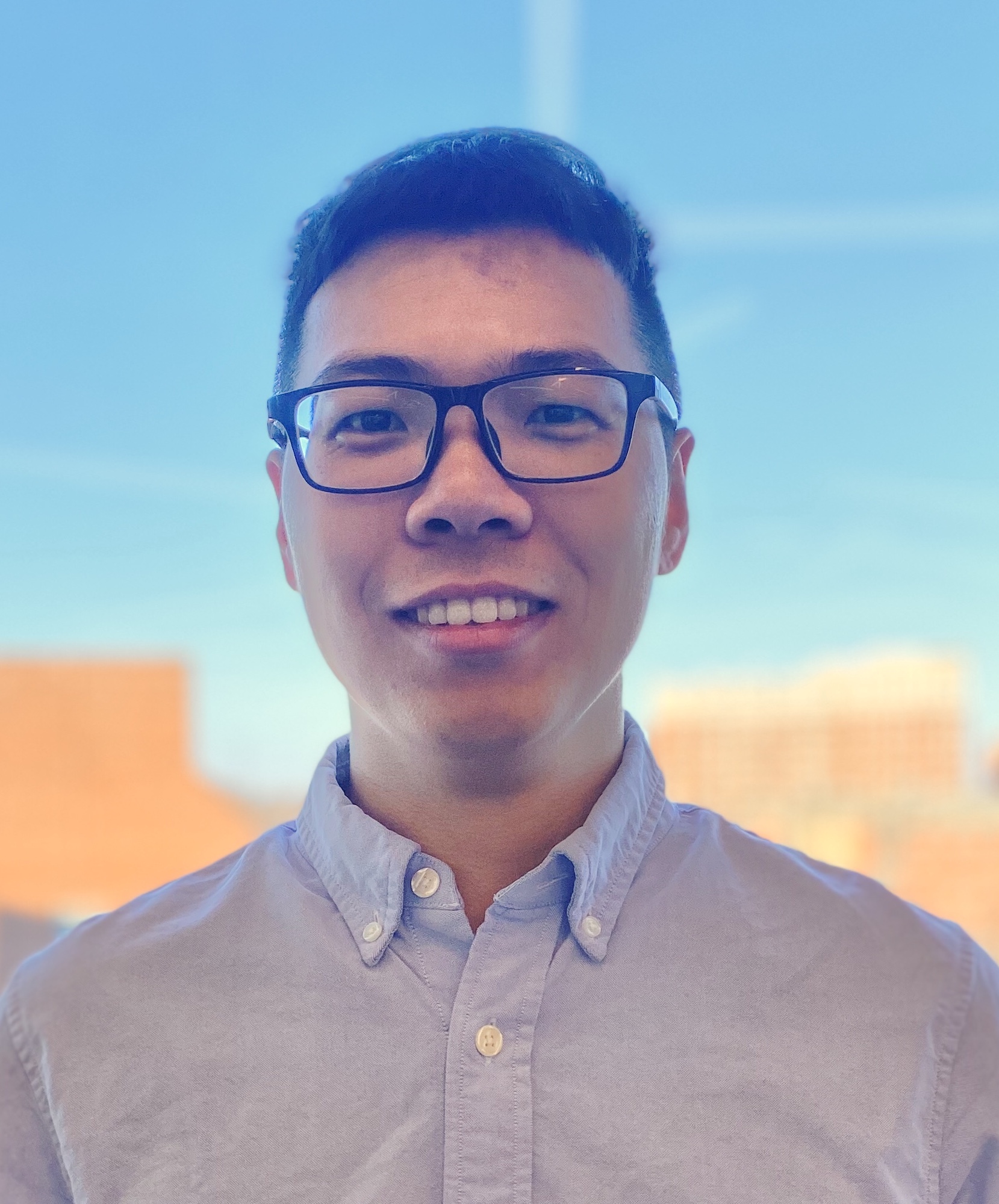}}]{Lifeng Zhou} is currently a Postdoctoral Researcher in the GRASP Lab at the University of Pennsylvania. He received his Ph.D. degree in Electrical \& Computer Engineering at Virginia Tech in 2020. He obtained his master’s degree in Automation from Shanghai Jiao Tong University, China in 2016, and his Bachelor's degree in Automation from Huazhong University of Science and Technology, China in 2013. 

His research interests include multi-robot coordination, approximation algorithms, combinatorial optimization, model predictive control, graph neural networks, and resilient, risk-aware decision making.
\end{IEEEbiography}

% % \vskip 0pt plus -1fil

% % if you will not have a photo at all:
\begin{IEEEbiography}[{\includegraphics[width=1in,height=1.25in,clip,keepaspectratio]{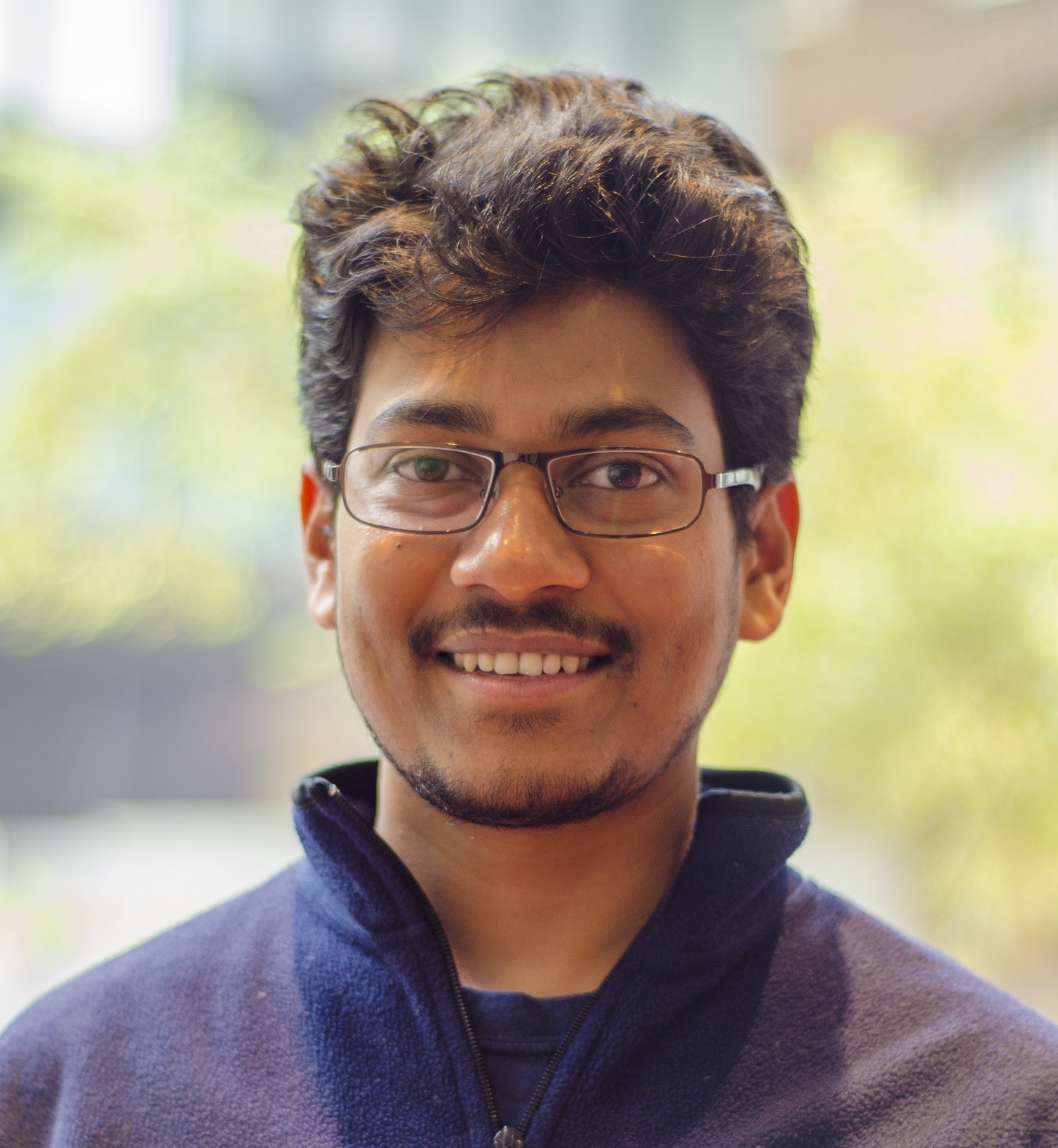}}]{Pratap Tokekar} is an Assistant Professor in the Department of Computer Science at the University of Maryland. Previously, he was a Postdoctoral
Researcher at the GRASP lab of University of
Pennsylvania. Between 2015 and 2019, he was an Assistant Professor at the Department of Electrical and Computer Engineering at Virginia Tech. He obtained his Ph.D. in Computer Science from the University of Minnesota in 2014 and Bachelor of Technology degree in Electronics and Telecommunication from College of Engineering Pune, India in
2008. He is a recipient of the NSF CISE Research Initiation Initiative award and an Associate Editor for the IEEE Robotics and Automation Letters and Transactions of Automation Science and Engineering.
\end{IEEEbiography}

\end{document}